\documentclass[mphil,icsa,twoside,logo]{infthesis}

\title{Deep Learning-driven Mobile Traffic Measurement Collection and Analysis}
\date{2021}
\author{Yini Fang}
\submityear{2021}

\usepackage{xcolor}
\usepackage{cite}
\usepackage{amsmath,amssymb,amsfonts}
\usepackage{graphicx}\usepackage{textcomp}
\usepackage{xcolor}
\usepackage{url}
\usepackage{flushend}
\usepackage{multirow}
\usepackage{mathtools,xparse}

\DeclarePairedDelimiter{\norm}{\lVert}{\rVert}
\usepackage{blindtext}
\usepackage[ruled,linesnumbered]{algorithm2e}
\usepackage{todonotes}
\usepackage{regexpatch}
\usepackage{xspace}
\usepackage{subfig}
\usepackage[normalem]{ulem}
\usepackage{enumitem}
\usepackage{pdfpages}

\newcommand{\name}{Spider\xspace}
\newcommand{\model}{SDGNet }
\usepackage{afterpage}
\usepackage[acronym,nomain,nonumberlist]{glossaries}
\makeglossaries

\newacronym{eb}{EB}{Exabytes}
\newacronym{pgws}{PGWs}{Packet Gateways}
\newacronym{RNCs}{RNCs}{Radio Network Controllers}
\newacronym{GPUs}{GPUs}{Graphics
Processing Units}
\newacronym{SDGNet}{SDGNet}{Spatiotemporal Dynamic Graph Network}
\newacronym{DGCN}{DGCN}{Dynamic Graph Convolution Network}
\newacronym{POMDP}{POMDP}{ Partially Observable Markov
Decision Process}
\newacronym{MAB}{MAB}{Multiarmed
Bandit}
\newacronym{KNN}{KNN}{K-Nearest Neighbors }
\newacronym{CS}{CS}{Compressive Sensing}
\newacronym{Co-STCS}{Co-STCS}{Collective Spatiotemporal Compressive Sensing}
\newacronym{GAN}{GAN}{Generative Adversarial neural Network}
\newacronym{LOO-SA}{LOO-SA}{Leave-One-Out Statistical Analysis}
\newacronym{MAE}{MAE}{Mean Absolute Error }
\newacronym{RL}{RL}{Reinforcement Learning}
\newacronym{MDP}{MDP}{ Markov Decision Process}
\newacronym{DRL}{DRL}{Deep Reinforcement
Learning }
\newacronym{DQN}{DQN}{Deep Q Network}
\newacronym{MCTS}{MCTS}{Monte Carlo Tree Search}
\newacronym{MCS}{MCS}{Mobile Crowdsensing}
\newacronym{SMCS}{SMCS}{Sparse Mobile Crowdsensing}
\newacronym{QBC}{QBC}{Query by Committee}
\newacronym{SRMF}{SRMF}{Sparsity Regularized Matrix Factorization}
\newacronym{MPNN}{MPNN}{Message passing Neural Network}
\newacronym{DCRNN}{DCRNN}{Diffusion Convolutional Recurrent Neural
Network}
\newacronym{GAT}{GAT}{Graph
Attention Networks }
\newacronym{GNN}{GNN}{Graph Neural Network}
\newacronym{GCRN}{GCRN}{ Graph Convolutional Recurrent Network}
\newacronym{STGCN}{STGCN}{Spatio-Temporal Graph Convolutional Network }
\newacronym{GLU}{GLU}{Gated Linear Unit}
\newacronym{DTW}{DTW}{Dynamic Time Warping }
\newacronym{PCC}{PCC}{Pearson Correlation
Coefficient}
\newacronym{MTRNet}{MTRNet)}{Mobile Traffic Reconstruction neural
Network }
\newacronym{BCE}{BCE}{Binary Cross-Entropy}
\newacronym{NMAE}{NMAE}{ Normalized Mean Absolute Error}
\newacronym{RMSE}{RMSE}{ Root Mean Square Error}
\newacronym{MSE}{MSE}{Mean Square Error}
\newacronym{STCS}{STCS}{Spatio-temporal Compressive
Sensing}
\newacronym{LSTM}{LSTM}{Long Short Term
Memory}
\newacronym{CNN}{CNN}{Convolutional Neural Network}
\newacronym{ST}{ST}{Spatiotemporal}
\newacronym{TCN}{TCN}{ Temporal Convolution Network}
\newacronym{GRU}{GRU}{Gated Recurrent Unit}
\newacronym{OTS}{OTS}{Ouroboros Training Scheme }
\newacronym{BNN}{BNN}{Bayesian Neural Network}

\newcommand{\Sign}{\includegraphics[scale=0.3, trim={0cm 0.85cm 0cm 1cm}]{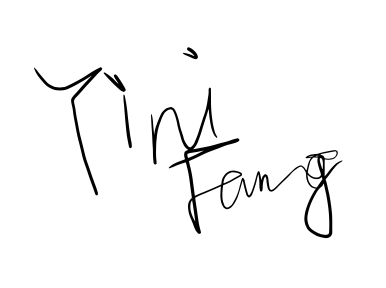}}

\makeatletter
\xpatchcmd{\@todo}{\setkeys{todonotes}{#1}}{\setkeys{todonotes}{inline,backgroundcolor=yellow,#1}}{}{}
\makeatother

\abstract{
The global mobile traffic consumption has been growing sharply, driven by the surge of smart IoT devices and the adoption of 5G technology. Further, the COVID-19 pandemic has accelerated the digital transformation of work and study, leading to tremendous traffic demands. Due to these reasons, mobile operators are facing pressing resource management problems related to storage and overhead of transferring large amounts of metadata that serve for analysing user demands and planning the infrastructure needed to accommodate these. In this context, intelligent measurement collection, precise geo-spatial traffic analysis, and forecasting are becoming essential to assuring high performance and cost-efficiency of cellular network deployments. 

Modelling dynamic traffic patterns and especially the continuously changing dependencies between different base stations, which previous studies overlook, is challenging. Traditional algorithms struggle to process large volumes of data and to extract deep insights that help elucidate mobile traffic demands with fine granularity, as well as how these demands will evolve in the future. Therefore, in this thesis we harness the powerful hierarchical feature learning abilities of Deep Learning (DL) techniques in both spatial and temporal domains and develop solutions for precise city-scale mobile traffic analysis and forecasting. Firstly, we design Spider, a mobile traffic measurement collection and reconstruction framework with a view to reducing the cost of measurement collection and inferring traffic consumption with high accuracy, despite working with sparse information. In particular, we train a reinforcement learning agent to selectively sample subsets of target mobile coverage areas and tackle the large action space problem specific to this setting. We then introduce a lightweight neural network model to reconstruct the traffic consumption based on historical sparse measurements. Our proposed framework outperforms existing solutions on a real-world mobile traffic dataset. Secondly, we design SDGNet, a handover-aware graph neural network model for long-term mobile traffic forecasting. We model the cellular network as a graph, and leverage handover frequency to capture the dependencies between base stations across time. Handover information reflects user mobility such as daily commute, which helps in increasing the accuracy of the forecasts made. We proposed dynamic graph convolution to extract features from both traffic consumption and handover data, showing that our model outperforms other benchmark graph models on a mobile traffic dataset collected by a major network operator.
}
\begin{document}

\maketitle
\begin{laysummary}
The COVID pandemic has accelerated digitalization as various events, lectures and meetings moved online. The total mobile traffic consumption continues to grow, and the cost resulting from computational overhead leads to expensive real-time traffic monitoring. It thus becomes urgent for mobile operators to apply intelligent solutions for precise traffic analysis with cost and latency reduction in mind. 

Deep learning techniques have gained increasing attention due to their excellent ability to make sense of complex patterns hidden in data, and due to recent advance in parallel computing that facilitates experimenting with such tools in an affordable manner. On the other hand, public mobile traffic high quality datasets are increasingly available, thanks to contributions by mobile operators. All of these combined opened up new opportunities for advancing research in the mobile networking area. 

In this thesis, we provide solutions to two of key challenges in this area, which build on deep learning.
Firstly, we propose a measurement collection and reconstruction framework, which enables mobile operators to only collect measurements from small sunsets of base stations and use these to `paint' a complete picture of the traffic consumption across the entire network. We apply deep reinforcement learning, a special branch of machine learning, and propose an algorithm to address the difficulty of developing a strategy for where to take such measurements, among the numerous choices available. Subsequently, we propose a lightweight model for reconstruction which accurately captures spatiotemporal relationships that are characteristic to mobile traffic. Our framework can significantly reduce the cost of data collection, transfer and storage, while offering high reconstruction quality.

Secondly, we propose a deep learning solution for forecasting future mobile traffic consumption, given historical traffic snapshots. Previous studies either assume static dependencies between base stations or
model such dependencies through location proximity,  which overlooks essential dynamics. In contrast, we leverage handover frequency information to model such dependencies. Our model provably effective in capturing the dynamic characteristics of mobile traffic. 

We conclude the thesis by pinpointing future research directions that are worth pursuing based on the work presented. 

\end{laysummary}

\begin{acknowledgements}
 First, I would like to express my sincere gratitude to my principal supervisor, Paul Patras, for giving me this precious opportunity to have a meaningful research life in Edinburgh and having offered me generous and patient guidance and support over the past two years. He has taught me the way of writing logical documents, giving clear presentation and improving my research skills. Not only in academic research, such skills also benefits me in the daily life. I will always keep what he has taught me in mind.
 
Secondly, I would also like to thank my second supervisor Rik Sarkar for providing me valuable feedback in the annual reviews. These comments have guided my research ideas and improved the outcomes. My thanks also go to my colleagues, Alec Diallo, Haoyu Liu, Luyang Xu, and Chaoyun Zhang. They have provided insightful feedback on my work and my presentations. I am particularly grateful to Chaoyun and Alec, as co-authors of my Globecom paper, their help has greatly accelerated my progress and improved the quality of the work. I would also like to thank Rui Li, although we have not worked together, she has offered me encouragement and inspired me.

Many thanks to Salih Ergut from Turkcell for providing me the dataset for my second project and offering useful feedback through meetings. Great appreciation to Cisco, for supporting me financially during my studies.

I am extremely grateful to my parents Weijie Fang and Caifen Zhang for their love, support and encouragement. Without them, I could have never finished this thesis.
\end{acknowledgements}

\begin{declaration}
I declare that this thesis was composed by myself, that the work contained herein is my own except where explicitly stated otherwise in the text, and that this work has not been submitted for any other degree or professional qualification except as specified.
Parts of this work have been published in or submitted to academic conferences and journals, including:

\begin{itemize}
    \item Y. Fang, A. Diallo, C. Zhang and P. Patras.``Spider: Deep Learning-driven Sparse Mobile Traffic Measurement Collection and Reconstruction", IEEE Globecom 2021, Madrid, Spain.
    \item Y. Fang, S. Ergut  and P. Patras. ``SDGNet: A Handover-aware Spatiotemporal Graph Neural Network for Mobile Traffic Forecasting", IEEE Communications Letters, 2021.
\end{itemize}
   
      \par 
  \vspace{0.2in}
   \begin{flushright}
   \Sign
   \end{flushright}
    \raggedleft({\em Yini Fang\/})
\end{declaration}

\tableofcontents

\glsaddall
\printglossary[type=\acronymtype,title=Acronyms]

\clearpage

\chapter*{Mathematical Notations}

\section*{Compressive Sensing}
\begin{itemize}
    \item $F$: Full sensing matrix
    \item $\hat{F}$: Inferred matrix
    \item $\mathcal{S}$: Binary selection matrix
    \item $\mathcal{C}$: Sparse matrix
    \item $\mathcal{L}$: Left part of singular value decomposition
    \item $\mathcal{R}$: Right part of singular value decomposition
    \item $\lambda$: A trade-off between rank minimization and accuracy fitness
\end{itemize}
\section*{Leave-One-Out Statistical-Analysis}
\begin{itemize}
    \item $e$: Inference error
    \item $\epsilon$: Predefined error threshold
    \item $\beta$: Predefined probability threshold
    \item $\mathbf{y}$: The collected values
    \item $\mathbf{\hat{y}}$:  The corresponding re-inferred values to $\mathbf{y}$
    \item $\mathbf{O}$: The observation, i.e.,  the absolute different of $\mathbf{y}$ and $\mathbf{\hat{y}}$
\end{itemize}
\section*{Reinforcement Learning}
\begin{itemize}
    \item $s$: State
    \item $a$: Action
    \item $p$: Transition probability
    \item $r$: Reward
    \item $\gamma$: Discount factor
    \item $\mathcal{R}$: Return
    \item $t$: Timestamp
\end{itemize}
\section*{Graph Convolution}
\begin{itemize}
    \item $G$: Graph
    \item $v$: Graph signal
    \item $* \mathcal{G}$: Spectral graph convolution operator
    \item $A$: Adjacency matrix
    \item $g$: Graph convolution kernel
    \item $L$: Normalized graph Laplacian
    \item $\Gamma$: Diagonal matrix of eigenvalues of $L$
    \item $U$: Graph Fourier basis
    \item $I$: Identity matrix
    \item $D$: Diagonal degree matrix
    \item $\theta$: A vector of polynomial coefficients
    \item $\mathcal{T}$: Chebyshev polynomial
    \item $\lambda_{max}$: Largest eigenvalue of $L$
    \item $* \mathcal{D}$: DCRNN operator
    \item $P$:  Power series of the transition matrix
    \item $\sigma$: the parameters of $g$
\end{itemize}
\section*{Chapter 3}
\begin{itemize}
    \item $X$: The number of cells in a row in the target area
    \item $Y$: The number of cells in a column in the target area
    \item $F$: The traffic consumption snapshot across all cells
    \item $d_{i, j}$: The volume of traffic at cell ($i, j$)
    \item $B$: Binary selection matrix
    \item $M$: Sparse measurement matrix
    \item $\mathbb{M}$: The sparse measurement matrices collected over recent timestamps
    \item $\hat{a}$: Pseudo action
\end{itemize}
\section*{Chapter 4}
\begin{itemize}
    \item $N$: The number of base stations
    \item $\mathbb{F}$: A sequence of mobile traffic consumption measurements
    \item $F$: Traffic snapshot across all base stations 
    \item $f_i$: Volume of traffic at the $i$-th base station
    \item $C$: The number of features
    \item $H$: The number of predicted steps
    \item $T$: The number of past observations
    \item $* \textbf{g}$: dilated causal convolution operator
    \item $\textbf{w}$: Weight of the graph
    \item $\mathbb{L}$: Loss
\end{itemize}

\chapter{Introduction}
The total global monthly mobile data traffic consumption exceeded 66 exabytes (EB) in Q1/2021 and is expected to surpass 300 EB across different mobile technologies combined by 2026 \cite{ericsson}. Additionally, the COVID-19 pandemic has forced people and organization to adjust to a new working paradigm: working-from-home. This leads to an acceleration in digitization and increased traffic demands. This continual rise in demand poses evermore pressing challenges on mobile networks and highlights the need for precise networking analysis and demand-aware management as a driver for assuring high performance of the modern 5G networking infrastructures. 

Network visibility plays an important role in the network monitoring and management. Data-driven network management hinges on accurate traffic measurements \cite{zhang2018long} collected by dedicated probes that are deployed, e.g., at packet gateways (PGWs) or Radio Network Controllers (RNCs)~\cite{naboulsi2015large}. Yet processing vast amounts of data in a scalable and timely fashion is ever more challenging, as it requires substantial local storage capabilities, it requires a heavy overhead in transferring detailed logs to central locations for analysis, it involves data filtering by scope (e.g., cell ID, session start/end times, traffic volume/type, etc.) to serve specific use cases \cite{cheng:2017}, and it relies on high-performance computing platforms to extract essential insights. On the other hand, modelling the complex cellular network precisely remains challenging, as previous studies overlook the dynamic dependencies between base stations and thus fail to capture the dynamic spatiotemporal correlations. In this context, developing an intelligent policy of measurement collection and performing accurate traffic reconstruction and forecasting become vital to establishing cost-efficient cellular networks with high performance and high user experience.


Recently, with the success of deep learning techniques, Convolutional Neural Networks, Recurrent Neural Networks  and Reinforcement Learning have revolutionized the way of solving a variety of machine learning tasks due to their powerful hierarchical feature learning ability in both spatial and temporal domains \cite{goodfellow2016deep}. They have produced outstanding results in research fields including computer vision, speech recognition, natural language processing, etc. \cite{zhang2019deep}. Deep learning heavily relies on massive data and computational resources. With the advanced parallel computing (e.g. Graphics Processing Units (GPUs), high-performance chips) and a number of available mobile traffic datasets, we can utilize the deep learning technique to the fullest. Subsequently, there remains huge potentials to address the aforementioned challenges with the power of deep learning. In this thesis, we will harness deep learning techniques to provide solutions for mobile traffic measurement collection and analysis, by means of algorithm design, implementation, optimization, experiments on real mobile network data sets, and comparison with the-state-of-art algorithms. 

\section{Research Challenges}

Monitoring large-scale mobile traffic across cities is a costly process that relies on dedicated measurement equipment, and it becomes urgent to develop elegant solutions to deal with the increasing cost driven by big data. Some of these probes have limited precision or coverage, others gather tens of gigabytes of logs daily, which independently offer limited insights. Extracting fine-grained patterns involves expensive spatial aggregation of measurements, storage, and post-processing. Processing such huge data relies on high-performance computing platforms to extract essential insights. Therefore, mobile operators urgently need more cost-effective  alternative  solutions  for  traffic  monitoring. 
Though recent progress in this domain appears promising, there still remain several research challenges to be addressed. Specifically:
\begin{itemize}
    \item One cost-effective  alternative to extensive traffic measurement collection is to gather data only from a subset of probes and reconstruct the traffic consumption from those unsampled cells. Recent work on adaptive sampling \cite{bash2004approximately}
\cite{willett2004backcasting}
\cite{unnikrishnan2013sampling}
\cite{salama2017adaptive}
\cite{wu2019dynamic} can only underpin suboptimal allocation of network resources and implicitly modest end-user quality of experience, because such methods randomly select the sampling locations without considering spatiotemporal correlations that are specific to mobile traffic. In practice, mobile operators need intelligent strategies to  instrument   dynamic   measurement   collection   campaigns using virtual probe modules that can be instantiated as needed. However, developing an intelligent policy for measurement collection is challenging: the cellular deployments usually comprise a large number of cells, so the number of possible sampling point selection options grows dramatically. How to quickly and accurately select a minimum number of optimal cells and reduce the collection cost needs to be further investigated.

    \item 
    Mobile traffic forecasting is another important driver for network resource allocation. Making highly-accurate mobile traffic predictions at city-scale is however challenging, as services continues to diversify and exhibit highly dynamic  patterns, while spatiotemporal correlations across different parts of a deployment are not straightforward to elucidate given unpredictable user mobility or terrain/coverage irregularities. A question to be answered is how can we meaningfully represent complex cellular network to be able to make high-quality forecasts? Prior work builds on distance-based Euclidean (grids) or invariant graph representations, which cannot capture dynamic spatiotemporal correlations with high fidelity. Especially, location proximity of base stations may not reflect strong dependency due to terrain constraints, while distant base stations may exhibit strong dependencies because of user daily commute patterns (e.g. along a motorway).
\end{itemize}

\section{Thesis Contributions}
In this thesis we utilize the power feature learning ability of deep learning techniques (e.g., reinforcement learning, graph neural networks) to capture spatiotemporal correlations in the non-linear and dynamic mobile traffic patterns, and develop state-of-the-art solutions for city-scale mobile traffic measurement collection and analysis.


\subsection{Deep Learning-driven Sparse Mobile Traffic Measurement Collection and Reconstruction}
Firstly, we design Spider, a mobile traffic measurement collection and reconstruction framework which significantly  reduces the cost of measurement collection and infers complete traffic consumption with high accuracy from sparse information. The framework outputs cell selection matrices  indicating at which locations to activate measurement collection given sparse historical traffic snapshot, to reduce the collection overhead while reconstructing the complete snapshot at all non-sampled cells. We utilize  a deep reinforcement learning  approach to find an optimal policy. We address the challenge of large action space at city scale by proposing an effective KNN-based algorithm to reduce the action space. Our agent learns in a tractable manner by sampling a subset of the action space including the most likely action and its nearest neighbours and selecting the action with the highest value, thus  circumventing the need to evaluate the entire action space.

We further introduce MTRNet, a lightweight and accurate neural model for mobile traffic reconstruction. MTRNet leverages the excellent spatiotemporal extraction ability of a 3D Convolutional Neural Network and outputs the complete snapshot from historical sparse measurements. MTRNet outperforms existing  algorithms with up to 67\% error reduction and up to 835$\times$ runtime reduction.

The experiments that we conducted on a real-world mobile traffic dataset show that our proposed framework outperforms existing solutions by effectively learning the complex characteristic of mobile traffic from sparse samples, even when applied to previously unseen traffic patterns such as those observed during holidays. Our \name samples up to 48\% fewer cells compared to a range of  benchmarks considered.

This work has been published in:
\begin{itemize}
    \item Y. Fang, A. Diallo, C. Zhang and P. Patras.``Spider: Deep Learning-driven Sparse Mobile Traffic Measurement Collection and Reconstruction", IEEE Globecom 2021, Madrid, Spain.
\end{itemize}

\subsection{A Handover-aware Spatiotemporal Graph Neural Network for Mobile Traffic Forecasting}
Secondly, we design SDGNet (Spatiotemporal Dynamic Graph Network), a handover-aware graph neural network model for mobile traffic forecasting. The cellular network is modelled as a directed and weighted dynamic graph. As  location proximity fails to reflect the dependencies between base stations, we leverage the handover frequency (i.e., the number of handovers), which reflects user mobility such  as  the daily  commute, to represent the weights of the graph.   SDGNet uses gated linear units to extract temporal features and we propose a DGCN (Dynamic Graph Convolution Network) that combines dynamic graph spectral convolution and diffusion graph convolution for spatial features extraction in both short- and long-term. Validated on a real-world mobile traffic dataset, our SDGNet excels at short-/mid-/long-term mobile traffic forecasting and achieves  up to 75.2\%, 59.4\% and 56.0\% higher short-, mid- and long-term accuracy than a range of benchmarks.

This work has been published in:
\begin{itemize}
    \item Y. Fang, S. Ergut  and P. Patras. ``SDGNet: A Handover-aware Spatiotemporal Graph Neural Network for Mobile Traffic Forecasting", IEEE Communications Letters, 2021.
\end{itemize}

\section{Thesis Organization}
We organize the rest of the thesis as follows:
\begin{itemize}
    \item \textbf{Chapter 2} presents an up-to-date background in regard to Machine Learning algorithms and related work on mobile traffic measurement collection and analysis utilizing these ML algorithms and other traditional algorithms. Specifically, we first review work on mobile traffic sampling and reconstruction. Then, we give a primer on Reinforcement Learning and its applications to sparse mobile crowdsensing. Finally, we summarize relevant work on Graph Neural Networks, the main types of graph convolution, and their applications on forecasting in spatiotemporal tasks.
    \item \textbf{Chapter 3} proposes Spider, a deep-learning mobile traffic measurement collection and reconstruction framework. We provide an in-depth explanation of the algorithm with visualization and pseudocode. We then perform experiments on a real-world mobile traffic dataset and evaluate the outcomes. 
    \item \textbf{Chapter 4}  introduces SDGNet, a graph  neural  network  model  for  long-term  mobile  traffic  forecasting. We explain the algorithm in details and perform experiments on a mobile traffic dataset collected by a major operator. We compare the performance with that of other graph models.
    \item \textbf{Chapter 5} concludes our thesis and provide several open research issues and future directions following from our projects, including (\textit{i}) end-to-end mobile traffic measurement collection and reconstruction,  (\textit{ii}) hybrid dynamic graphs for mobile traffic forecasting, and  (\textit{iii}) incorporating uncertainty in mobile traffic forecasting with Bayesian Graph Convolutional Networks.
\end{itemize}

\chapter{Background \& Related Work}
This chapter reviews relevant related work in Machine Learning and traditional algorithms that have applications to mobile traffic analysis. Specifically,
\begin{itemize}
    \item \textbf{Mobile Traffic Sampling and Reconstruction} focuses on the traditional algorithms (e.g., Adaptive sampling, Compressive Sensing) used in sampling and data reconstruction. 
    \item \textbf{Reinforcement Learning and Its Applications} elaborates on Reinforcement Learning algorithms and their applications to sparse mobile crowdsensing.
    
    \item \textbf{Graph Neural Networks and Their Applications} introduces the concept of Graph Neural Network and reviews recent work on forecasting in spatiotemporal tasks, including mobile traffic prediction.
\end{itemize}

\section{Mobile Traffic Sampling and Reconstruction}

Sampling  and  reconstruction  is  an  effective  way  to  process  vast  amounts  of data  in  a  scalable  and  timely  fashion,  where  only  a  subset  of  data  collection points  are  activated  and  complete  network  traffic  snapshots  are  subsequently reconstructed via interpolation. 

\subsection{Sampling} 
Sampling methods can be divided into uniform sampling and adaptive sampling. Adaptive sampling determines the sampling amount based on predefined criteria, and thus it is more effective and further reduce the sampling cost compared to random sampling.

Bash et al. \cite{bash2004approximately} perform random sampling on sensors in sensor networks for data collection, and the sampling rate is proportional to the diameter of the wireless network. Adaptive sampling methods are usually used in sensor management where the sensing system is capable of actively managing the sensor resources in real time. The sensor management problem is usually modelled as a partially observable Markov decision process (POMDP) \cite{castanon1997approximate}\cite{evans2001optimal}\cite{krishnamurthy2002algorithms}, and some works treat it as a multiarmed bandit (MAB) problem \cite{krishnamurthy2001hidden}\cite{krishnamurthy2003correction}. Willett et al. \cite{willett2004backcasting} propose an adaptive sampling approach to reduce communications and energy consumption, where the sampling rate is determined by estimating the environment through previous wireless sensors' communication. Unnikrishnan et al. \cite{unnikrishnan2013sampling} propose a sampling and reconstruction paradigm using sensors that move through space.
Because a moving sensor encounters such a spatial field along its path as a time-domain signal, they employ a time-domain anti-aliasing filter  prior to sampling the signal received at the sensor. This filtering process results in complete suppression of spatial aliasing in the direction of movement of the sensor. Salama et al. \cite{salama2017adaptive} design an adaptive sampling method for computer network traffic parameters, and the sample rate changes according to transmission rate or traffic behaviour. Wu et al. \cite{wu2019dynamic} propose an adaptive sampling and dynamic traffic prediction scheme for 5G HetNets, which predicts the IoT traffic load by compressed sensing and then use a linear predictor to adaptively adjust the sampling rate based on the prediction error. 

Although these adaptive sampling methods adjust the sampling frequency based on previously measured network activity, sampling locations are routinely selected at random, which \textit{(i)} overlooks important spatiotemporal correlations  specific to mobile traffic, leading to inaccurate re\-construction; and \textit{(ii)} such myopic view leaves room for sampling overhead reduction, without sacrificing interpolation~fidelity.

\subsection{Data reconstruction} 
Traditionally, K-Nearest Neighbors (KNN) and Compressive Sensing (CS) are widely applied to data reconstruction problems. KNN interpolates missing values by using a weighted average of the values of the k nearest neighbours. Wang et al. \cite{wang2017space} propose spatial KNN (KNN-S) and temporal KNN (KNN-T) for mobile traffic data reconstruction, which perform KNN on columns or rows. However, the KNN algorithm has significant drawbacks. Firstly, because the inferred values are based on their neighbours, the algorithm cannot work if the matrix is too sparse. Secondly, the inferred matrix can be too smooth and thus KNN ignores the effects of local fluctuations. 

CS infers the two-dimensional full sensing matrix $F \in \mathbb{R}^{n \times m}$ based on the low-rank property:
$$
\begin{array}{c}\min \operatorname{rank}(\hat{F}) \\ \text { s.t., } \hat{F} \circ \mathcal{S}=\mathcal{C}, \end{array}
$$
where $\circ$ represents element-wise multiplication, $\mathcal{S}$ is the two-dimensional binary selection matrix and $\mathcal{C}$ is the two-dimensional sparse matrix, in which non-zero elements are data that have been collected, and zero elements are unknown data. Matrices $F$ are required to satisfy properties such as the restricted isometry property (RIP), making this optimization problem nonconvex and NP-hard. According to singular value decomposition ($\hat{F} = \mathcal{L} \mathcal{R}^T$) and compressive sensing theory \cite{donoho2006compressed}, it is proved that minimizing the rank of $\hat{F}$ is equivalent to minimizing the sum of $\mathcal{L}$ and $\mathcal{R}$’s Frobenius norms under the condition that the rank of  $\hat{F}$ is less than the rank of $\mathcal{L} \mathcal{R}^T$ \cite{zhang2009spatio}:
$$
\begin{array}{c}\min \|\mathcal{L}\|_{F}^{2}+\|\mathcal{R}\|_{F}^{2} \\ \text { s.t.  } \mathcal{L} \mathcal{R}^{T} \circ S=C,\end{array}
$$
where $\mathcal{L} \in \mathbb{R}^{n \times r}$ and $\mathcal{R} \in \mathbb{R}^{r \times m}$. $r$ represents the rank. In practice, there are often some errors in real-life collected data, therefore the above optimization problem is usually converted as follows [48]:
$$
\min \lambda\left(\|\mathcal{L}\|_{F}^{2}+\|\mathcal{R}\|_{F}^{2}\right)+\left\|\mathcal{L} \mathcal{R}^{T} \circ S-C\right\|_{F}^{2},
$$
where $\lambda$ is used to make a trade-off between rank minimization and accuracy fitness.

Rana et al. \cite{earphone} propose an urban noise mapping system where compressive sensing is employed to extrapolate incomplete and random samples obtained via crowdsourcing. Zhu et al. \cite{zhu2012compressive} use compressive sensing on traffic estimation from the periodically collected data by probe vehicles. He and Shin \cite{he2018} propose a Bayesian compressive sensing to recover a crowdsourced signal map. Besides the reconstructed signal, the algorithm also outputs the confidence intervals that can be utilized for incentive map design in order to  incentivize crowdsourcing participants to provide data for better signal coverage and quality.
Wang et al. \cite{wang2017space} design Collective Spatiotemporal Compressive Sensing (Co-STCS), which sets temporal and spatial constraints on the reconstructed data. However, real-world traffic matrices rarely meet the conditions required by CS-like algorithms. Specifically, these algorithms 1) require measurement matrices satisfying the Restricted Isometry Property and low rank, which does not work with deterministic sampling constrained by the types and geographical distribution of probes; 2) require the mobile traffic distribution obeys a Gaussian distribution, which may not be satisfied in practice; 3) require a linear relationship between sparse traffic and actual traffic measurements, which is not the case, given the fact that they have highly non-linear relationships.

More recently, neural network models have been proposed to predict data traffic demand in large coverage areas. Tian \cite{tian2017data} et al. design a simple
four-layers neural network as an autoencoder for data reconstruction. S. Ma et al. \cite{ma:2019} propose an encoder-decoder model for service demand prediction from available incomplete measurements. C. Zhang et al. \cite{zipnet} design a Generative Adversarial neural Network (GAN) architecture tailored to mobile traffic super-resolution and infer traffic consumption with high geographic granularity based only on coarse aggregate measurements. Chai et al. \cite{chai2020deep} propose a model based on an encoder-decoder-style U-Net convolutional neural network for irregularly and regularly missing data reconstruction. The advantage of leveraging neural networks to learn spatiotemporal correlations of mobile traffic is salient. Aside from its excellent accuracy, it is also fast. In contrast, algorithms that aim to find the reconstructed matrix by minimizing the objective function (such as CS) take a long time to reach a convergence point.

Although this approaches attain accurate interpolations by harnessing the exceptional abstract feature extraction abilities of deep learning, they built on the premise that collection points are given. In practice, operators needs intelligent strategies to instrument dynamic measurement collection campaigns.

\subsection{Reconstruction Quality Estimation}
One challenge in data reconstruction is reconstruction quality estimation. In practice, we cannot get access to the complete data for evaluating the reconstructed data, therefore the reconstruction quality can only be estimated. In the following, we discuss several ways of estimating the reconstruction quality.

\paragraph*{Leave-One-Out Statistical-Analysis} 
Wang et al. \cite{wang2017space} leverage Leave-One-Out Statistical Analysis (LOO-SA) for quality estimation. Instead of directly predicting the inference error $e$, LOO-SA use statistical analysis to estimate the probability of the inference error being no larger than the predefined error threshold $\epsilon$ (i.e., $P(e \leqslant \epsilon)$). If the estimated probability  is larger than $\beta$ ($\beta$ being the predefined probability threshold), then the quality of reconstruction is considered to be satisfied. There are two steps in the LOO-SA method. First, LOO-SA uses the leave-one-out resampling method to obtain a set of LOO observations. Specifically,  for each collected value, LOO temporarily eliminates this value (i.e., sets it to zero in the sparse matrix) and then  re-infers the data.  After enumerating all the collected data in the current cycle, we finally get two vectors $\mathbf{y}$ and $\mathbf{\hat{y}}$: $\mathbf{y}$ stores the collected values and $\mathbf{\hat{y}}$ stores the corresponding re-inferred values.  Assume that we have already collected  data from $n$ cells, then:
$$
\mathbf{y}=\left\langle y_{1}, y_{2}, \ldots, y_{n}\right\rangle, \quad \mathbf{\hat{y}}=\left\langle \hat{y}_{1}, \hat{y}_{2}, \dots, \hat{y}_{n}\right\rangle,
$$
where $y_i$ is the $i$th ground truth data collected in the current cycle, and $\hat{y}_i$ is the corresponding re-inferred data by leaving $y_i$ out of the collected data. The observation $\mathbf{O}$ is defined as the absolute different of $y$ and $\hat{y}$:
$$
\begin{aligned} \boldsymbol{\mathbf{O}} &=\left\langle O _{1}, O_{2}, \ldots, O_{n}\right\rangle=\operatorname{abs}(\mathbf{\hat{y}}-\mathbf{y}) \\ &=\left\langle\left|\hat{y}_{1}-y_{1}\right|,\left|\hat{y}_{2}-y_{2}\right|, \ldots,\left|\hat{y}_{n}-y_{n}\right|\right\rangle . \end{aligned}
$$

Based on the $\mathbf{y}$ and $\mathbf{\hat{y}}$ obtained from LOO, we can conduct Bayesian analysis to estimate the probability distribution of the inference error $e$. $e$ is seen as an unknown parameter with a prior probability distribution $g(e)^3$. According to the \textit{Bayes’s Theorem}, the posterior probability distribution $g\left(e | \boldsymbol{O}\right)$ is:
$$
g\left(e | \boldsymbol{O}\right)=\frac{f\left(\boldsymbol{O} | \boldsymbol{e}\right) g\left(\boldsymbol{e}\right)}{\int_{-\infty}^{\infty} f\left(\boldsymbol{O} | \boldsymbol{e}\right) g\left(\boldsymbol{e}\right) d e} ,
$$
where $f\left(\boldsymbol{O} | \boldsymbol{e}\right)$ is the likelihood function that represents the conditional probability of observing $O$ given $e$. We can then estimate $P(e 	\leq \epsilon)$ given the posterior $g\left(e | \boldsymbol{O}\right)$:
$$
P\left(e \leq \epsilon\right) \approx \int_{-\infty}^{\epsilon} g\left(e | \boldsymbol{O}\right) d e.
$$

\begin{figure}[]
\begin{center}
  \includegraphics[width=0.65\linewidth]{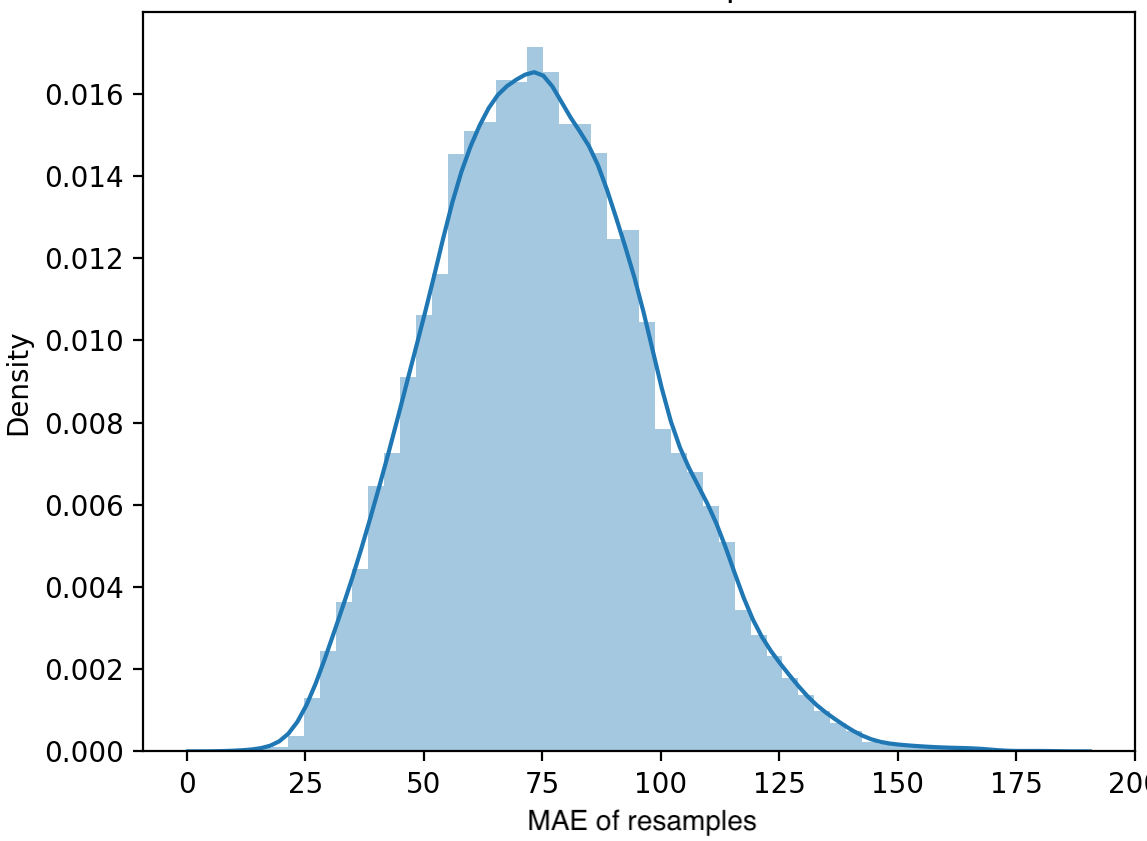}
  \caption{Normal distribution of bootstrap resamples}
  \label{fig:bs_dis}
 \end{center}
\end{figure}

For those inference errors that do not follow the normal distribution, one can use the Bootstrap to estimate the inference error distribution. The advantage of the Bootstrap is that no assumption on
the distribution of the observations needs to be made. The basic idea of the  Bootstrap is to generate
a large number of resamples from the observations with replacement, and
the size of each resample is equal to the original observations. Then, unknown statistics of the
population, e.g., the mean, can be inferred from the bootstrapping resamples \cite{wang2017space}. Specifically, after obtaining the observations $O$ from LOO, one would generate $m$ ($m$ is usually larger than one thousand) bootstrapping resamples by resampling $O$ and the size of each resample is the same as $O$. Then we can get a normal distribution of bootstrap resamples, and that distribution of bootstrap estimates is a data-driven estimation of the sampling distribution of the sample mean. Figure \ref{fig:bs_dis}  shows an example of the Mean Absolute Error (MAE) distribution of bootstrap resamples where $m=10000$ using a public mobile traffic dataset. Then we can calculate the confidence interval of this distribution, which  indicates the estimate probability $P(e \leqslant \epsilon)$.

\paragraph*{Concept drift} When making inferences, it is common to assume that the training and testing data obey the same distribution. If the distribution of the testing data changes (i.e., concept drift), the assumptions made by the regression model break, resulting in unknown and possibly large errors. Concept drift can be used to detect the generalization error of a model. There are two types of concept drift: real concept drift and virtual concept drift. The former refers to the change in the conditional probability $p(\mathbf{\hat{y}}|\mathbf{y})$ and the latter to the change in the distribution of the covariates $p(\mathbf{y})$. Tiittanen et al. \cite{tiittanen2019estimating} propose an efficient framework for estimating the generalization error of regression models due to virtual concept drift. A distance $d(\mathbf{y})$ is defined to measure how far a vector $\mathbf{y}$ is from the data which was used to train the regressor. Small values of $d(\mathbf{y})$ mean that the input is close to the training data and the regressor function should be reliable, while a large value of $d(\mathbf{y})$ means that the input has moved away from the training data, after which the regression estimate may be inaccurate. They first train different regression functions, say $f$ and $f^{\prime}$, on different subsets of the training data. Then $d(\mathbf{y})$ is the difference between the predictions of these two functions, e.g., $d(\mathbf{y}) = |f(\mathbf{y}) - f^{\prime}(\mathbf{y})|$. It is observed that this kind of distance measure has the suitable property that if some attributes are independent of the dependent variable $\mathbf{\hat{y}}$, then they will not affect the behaviour of the regression functions and, hence, the distance measure $d$ is not sensitive to them.

In summary, LOO-SA algorithm has a large time complexity, and requires configuring two hyperparameters (error threshold  $e$ and probability threshold $P(e \leqslant \epsilon)$). Concept drift can only be used for the generalization error of the model. In order to have a fast and simple solution, one can leverage the power of deep learning to approximate the function of reconstruction quality.

\section{Reinforcement Learning and Its Applications}
\subsection{Reinforcement Learning}
Reinforcement Learning (RL) is a commonly-used algorithm in Machine Learning where an agent learns by interacting with the environment. It has been widely used in various fields, such as chemistry, manufacturing, robotics, etc. Instead of explicitly being taught the behaviour, the agent learns its behaviour from the consequences of the taken actions in the past. We use a Markov Decision Process (MDP) \cite{mdp} to define an environment. There are five components in an MDP:
\begin{itemize}
    \item A set of states $S = \{ s_t \mid t=0, 1, ..., T \}$.
    \item A set of actions $A = \{ a_t \mid t=0, 1, ..., T \}$.
    \item $ P = \{ p_t \mid t=0, 1, ..., T \}$, the probability that action $a$ in state $s$ will lead to the next state $s^\prime$ at time $t$.
    \item $ R = \{ r_t \mid t=0, 1, ..., T \}$, the immediate reward received after the transition $(s, a)$ to $s^\prime$ at time $t$.
    \item Discount factor $\gamma$ that is used to compute the return (i.e., the accumulative reward) $
\mathcal{R}=\sum_{t=0}^{T} \gamma^{t} r_{t}
$. 
\end{itemize}

\begin{figure}[]
	\centering
\includegraphics[width=0.38\columnwidth]{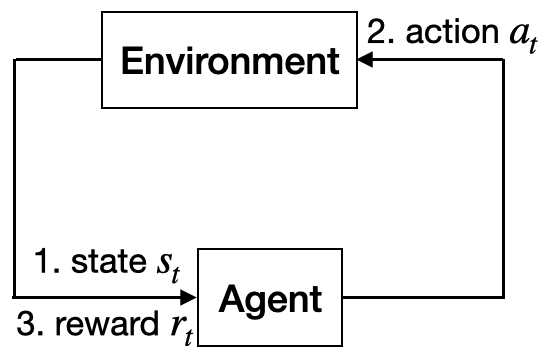}
	\caption{How an agent interacts with the environment in RL.
	}
	\label{rl_process}
\end{figure}

Figure \ref{rl_process} illustrates how the agent interacts with the environment. For every timestamp in an episode (i.e., from an initial state to a terminal state), the agent receives the current state $s_t$ and chooses an action $a_t$ from the set of available actions according to its policy. The environment receives the action, which returns a reward $r_t$. Then the environment transitions to a new state $s_{t+1}$. The agent continuously adjusts its behaviour based on the feedback from the environment. The goal of the agent is to find a policy that maximizes the return it receives over the whole episode. In order to find an optimal policy and obtain the maximal return, the agent must take the long-term consequences of its actions into consideration, even if sometimes the immediate reward associated with this action is negative. Additionally, the agent must balance exploration and exploitation of the actions. More exploitation might omit good actions and result in a suboptimal policy, while more exploration might take a long time, wasting computational memory. A good RL algorithm should find the optimal policy in the shortest possible time.

Based on the policy generation, we can divide RL into three categories: 1) Value-based: a value network assigns a value to a state by calculating the expected return of the state. The agent chooses the action that leads to that optimal state based on the values of a set of states; 2) Policy-based: a policy network is learned to output an optimal action given a state; 3) Actor-critic: a mix of the two. Based on the knowledge of the environment, RL can be divided into two categories: 1) Model-based:  the agent learns optimal policy indirectly by learning a model of the environment; 2) Model-free: the agent learns directly from the value function.

The Q-learning algorithm \cite{watkins1992q} is among the most well-known model-free value-based RL algorithms. A return function, Q, is introduced to map a state-action pair to the expected return (i.e., Q-value) for the agent to estimate and determine optimal actions in response to different states. The Q learning algorithm requires a Q-table which stores each state-action pair to its corresponding Q-value. However, Q learning is unstable, and this instability comes from the correlations present in the sequence of observations, the fact that small updates to Q may significantly change the policy and the data distribution, and the correlations between Q and the target values. 

\paragraph*{Deep RL} If the state space or action space is large, it is not feasible to remember all the possible state-action combinations and their corresponding rewards, and the RL learning becomes ineffective and overwhelming. This is why Deep Reinforcement Learning (DRL) is introduced. DRL leverages neural networks to approximate the  reward function, which can not only avoid such enumeration to save computational memory, but also generalize across the state spaces even if the state has never been seen before. To avoid storing a Q table that takes large memory,  Mnih et al. \cite{mnih2013playing} propose Deep Q Network (DQN), which maps input states to action-Q value pairs by a neural network, and they also propose an experience replay mechanism for training the neural network in order to alleviate the problems of correlated MDP transitions. Moreover, to improve the instability of Q learning, they introduce a target network that is a copy of the  Q network. This target network is frozen  to serve as a stable target for learning for a fixed number of timestamps. However, it is not feasible to train DQN if the action space is large.

\paragraph*{Monte Carlo Tree Search} 
It is not easy in RL to lean an optimal policy because we cannot enumerate all the possibilities at each episode and choose the combination with the highest reward. The agent inefficiently performs exploration of the action space by randomly selection, and it may lead to learning of some other suboptimal policy. To address this problem, Sliver et al. \cite{silver2017mastering} combine Monte Carlo tree search (MCTS) with RL, where a search tree is built to store action values and prior probabilities of actions. MCTS combines RL with  tree search by using  look-ahead  search to  decide  the  next  action  based  on  knowledge  of  the environment. Unlike value-based RL where a value network predicts state values, MCTS combines the actual returns from rollouts and predicted state values to approximate state values more accurately. It avoids the typical brute force tree searching and leverages statistical sampling by conducting Monte Carlo rollouts \cite{magnuson2015monte}. The efficiency of action sampling in MCTS results in superior policies and a faster convergence than other RL methods such as DQN.

\subsection{Sparse Mobile Crowdsensing}

\begin{figure}[]
	\centering
\includegraphics[width=0.73\textwidth]{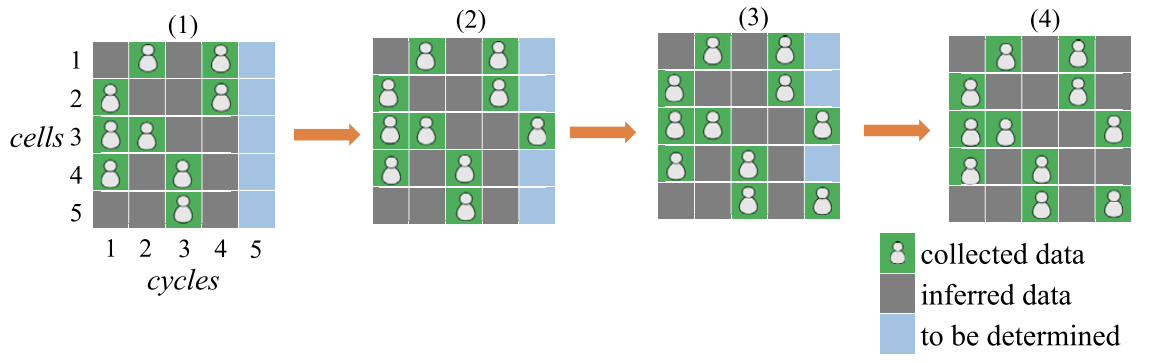}
\caption{An example of SMCS \cite{rl_mcs2} }
	\label{smcs}
\end{figure}

With the unprecedented growth of smartphones, urban sensing applications have been boosted by mobile crowdsensing (MCS). MCS is a paradigm that leverages the sensors in users' smartphones for data collection, especially for areas that are not covered by sensing infrastructure. In order to obtain high-quality sensed data, numerous users covering the whole area should be recruited, which sometimes is impossible. Early studies focus on minimizing the redundant number of allocated data collection tasks under a given quality requirement \cite{zhang20144w1h} \cite{guo2015mobile}. Such methods do not significantly reduce the number of total recruited users, thus still resulting in high sensing cost. To deal with this issue, a paradigm called Sparse Mobile Crowdsensing (SMCS) is proposed, which select urban subareas via a policy, and recruit users covering these subareas to upload sensing data using mobile devices, subsequently inferring the missing data \cite{sparsemcs}. Figure \ref{smcs} shows an example of SMCS. Assume the target area is divided into five cells, and there are four selection cycles previously. In the fifth cycle, the data at cell 3 and 5 is selected due to the policy, and the corresponding data is collected. Then the sparse data at the third step in the figure is used for data reconstruction at cell 1, 2 and 5. 

Wang et al. use Query by Committee (QBC) for cell selection and  Sparsity  Regularized  Matrix  Factorization (SRMF) for data inference \cite{rl_mcs}. Query by Committee (QBC) is a voting method of selective sampling, where disagreement amongst an ensemble of hypotheses is used to select data for labelling \cite{seung1992query}. In SMCS problem, QBC is used to select the salient cell to allocate the next sensing task, and `committee' in this context refers to a variety of data reconstruction algorithms, e.g., K-nearest neighbours and compressive sensing. This approach applies each algorithm in the committee to infer the complete data, and it allocates the next sensing task to the cell with the largest variance among the inferred values of different algorithms \cite{seung1992query}.

SMCS is later viewed as a Markov Decision Process, and trained in a manner of RL in  \cite{rl_mcs} \cite{rl_mcs2}. The aim of the RL policy is to select a minimum number of cells, so that the reconstruction quality based on collected values from these cells is under a predefined threshold. They model key factors in MDP as follows:
\begin{itemize}
    \item State: historical sparse snapshots that contains the collected measurements from previously chosen cells.
    \item Action: a cell that has not been chosen.
    \item Return: negative of the number of selected cells. The agent is trained to select the action with the highest return.
\end{itemize}
The agent chooses one cell each time and stops the MDP episode if the reconstruction error is less than a threshold. The design of the return function is not effective, because it is only related to the number of selected cells, and such feedback takes a long time to reflect on the agent's behaviour adjustment. They apply a CNN \cite{rl_mcs} or a LSTM \cite{rl_mcs2} for the agent's structure and train the agent using Deep Q-Network algorithm. The structure of the agent should be capable of extracting underlying spatiotemporal correlations from the state, however, CNN or LSTM fail to meet such a requirement. Additionally, we notice that in their evaluation, the variance of the testing dataset is small, which is unrealistic in practice. The mobile traffic data across a city should exhibit large variability.

\paragraph*{Lessons learned} From the previous studies on applying RL on SMCS, drawbacks and potential opportunities are now  identified. First, compared to infrastructure-level data that is collected from dedicated probes, such user-level data collection paradigm has given rise to concerns including data unavailability, privacy issues and inefficiency. Secondly, the proposed model in \cite{rl_mcs2} is evaluated on a mobile traffic dataset where the target area is partitioned into 10×10 cells, while does not match the scale of mobile network deployments that usually contain hundreds to thousands of cells. Since DQN is not capable of dealing with large action spaces, the proposed model fails to adapt to real-world city-scale cellular networks. Finding a policy that maps from states to probabilities of choosing an action in a large action space would require massive amounts of memory. Further, exploring the action space exhaustively until sufficient cells are selected would involve impractically large runtimes. 

Dulac-Arnold et al. \cite{dulac2015deep} provide a solution to large action space in RL. Instead of mapping from states to probabilities of choosing an action, the proposed policy directly outputs an action, thus decoupling the policy's complexity from the cardinality of the action space. This output action indicates a searching area in the action space, generating a set of potential actions that the agent explores. An approximate nearest neighbour search is used to find this set of potential actions. This proposed algorithm effectively allows both learning and acting in a tractable manner.

Going back to the previous SMCS challenge, we could potentially combine Monte Carlo Tree Search with this large action space solution to train an agent selecting from a large number of cells. Alternatively, thinking from another perspective, as SMCS involves mobile traffic reconstruction, we could leverage the reconstruction quality metrics to determine the action value, eliminating MCTS rollouts and speeding up the convergence. 

\newpage

\section{Graph Neural Networks and Its Applications}
\subsection{Graph Neural Networks}

\begin{figure}[]
	\centering
\includegraphics[width=0.65\textwidth]{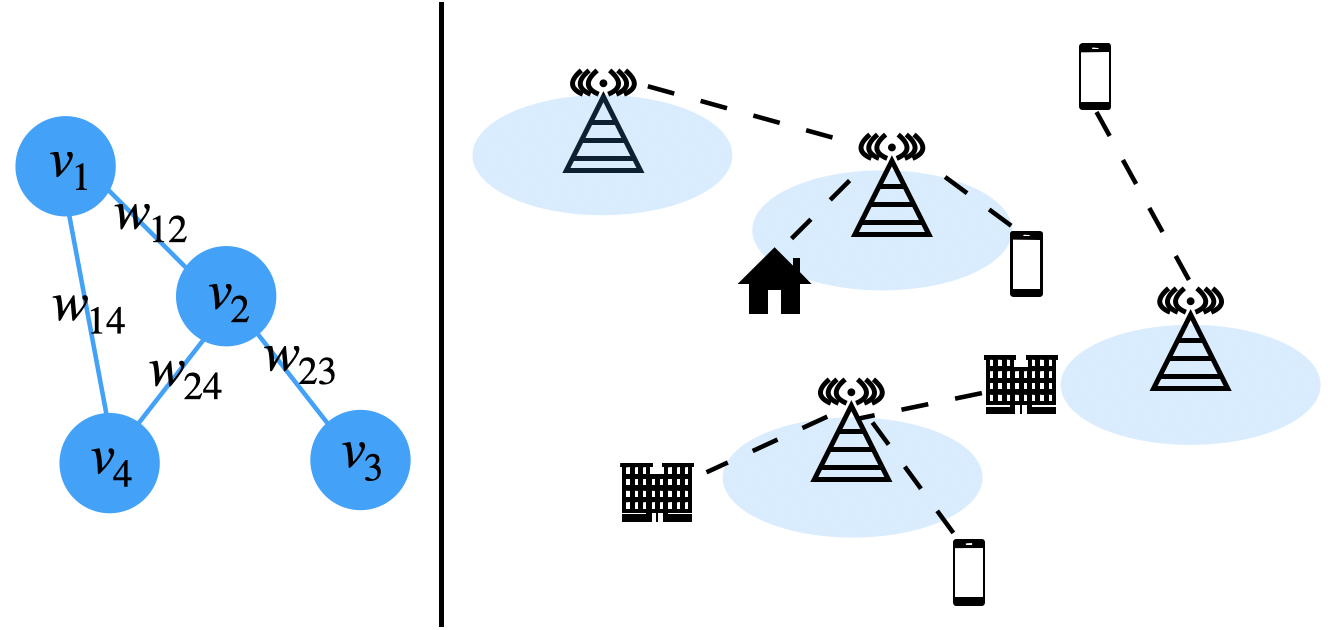}
\caption{Modelling the cellular network (on the right) as a graph (on the left). Intuitively, graphs can represent accurately the spatial correlations in the cellular network. The weight (i.e., dependency or relationship) between two nodes (i.e., base stations) is quantified.}
	\label{graph}
\end{figure}

In most machine learning tasks such as image classification and natural language processing, data is usually represented in Euclidean space (e.g., text, images, and video). In recent years, research on deep learning techniques for non-Euclidean data (e.g., citations, friendships, and interactions) has been gaining more and more attention \cite{graph_survey}. These data are represented as graphs with complex relationships and interdependency between objects \cite{graph_survey} which are overlooked in the Euclidean domains. Coates et al. \cite{998081}  utilized graph structure in network tomography, where a node represents a computer/cluster of computers/router and an edge represents a direct link between two nodes. The weight of an edge is the bandwidth of the corresponding connection. They use these to deal with large-scale network inference problems (e.g., link-level parameter estimation and sender-receiver path-level traffic intensity estimation). Assume a graph consists of $n$ nodes (i.e. vertices), it can be formulated as $G = (v, e, w)$, where $v \in \mathbb{R}^n$ represents the graph signal, $e \in \mathbb{R}^{n\times n}$ is a set of edges indicating the connectedness between vertices, and $w \in \mathbb{R}^{n\times n}$ denotes the weighted adjacency matrix of $G$. Figure \ref{graph} shows the benefits of modelling the cellular network as a graph. Base stations are represented as vertices, and the weight in the adjacency matrix indicates the dependency between two base stations. Such dependency is dynamic and non-linear. The graph signal can be modelled as the traffic consumption and the timestamp information. Compared to matrices, the merits of graph representation are obvious.

There are various approaches generalizing convolution operations on graphs. Message passing Neural Networks (MPNNs) \cite{gilmer2017neural} were initially used in chemistry. These are iterative decoding algorithms that aggregate the features of neighbourhood nodes into every node. Some algorithms generalize CNN to graph data. M. Niepert et al. \cite{spatial} propose non-spectral graph convolution by rearranging the vertices into certain grid formats and then processing these by normal convolution operations. J. Bruna et al. \cite{spectral} propose spectral graph convolutions to apply convolutions in spectral domains (details are explained in next subsection). Follow-up studies reduce the computational complexity. Kipf \cite{7kipf2016semi} et al. use a localized first-order approximation of spectral graph convolutions to simplify the spectral graph convolution. Defferrard et al. \cite{defferrard2016convolutional} define a convolutional neural network in the context of spectral graph theory. In contrast to CNN-type convolution, a  Diffusion Convolutional Recurrent Neural Network (DCRNN) \cite{li} models the traffic flow as a diffusion process on a directed graph (details are explained in next subsection). Velickovic et al. \cite{6velivckovic2017graph} propose Graph Attention Networks (GAT) that leverage masked self-attentional layers to learn graph features.

\subsection{Graph Convolution Primer} \label{gcn}
As spectral graph convolution and DCRNN are fundamental to Graph Neural Networks (GNNs) and is vital to our project, in this subsection, we explain these two types of graph convolution in details.

First, we define the spectral graph convolution operator $
* \mathcal{G}
$ as the multiplication of a graph signal $v \in \mathbb{R}^n $ and static adjacency matrix $A \in \mathbb{R}^{n \times n} $ with a kernel $g \in \mathbb{R}^n $, i.e.,
$$
v * \mathcal{G} g = g (L) v = g (U \Lambda U^{T}) v = U g (\Lambda) U^T v,
$$
where $L \in \mathbb{R}^{n \times n}$ is the normalized graph Laplacian, $\Lambda$ is the diagonal matrix of eigenvalues of $L$, and the graph Fourier basis $U \in \mathbb{R}^{n \times n}  $ is the matrix of eigenvectors of $L$. The transition is based on the equation $L = I_n - D^{-\frac{1}{2}} A D^{-\frac{1}{2}} = U \Lambda U^T$, where $I_n$ is an identity matrix and $D \in \mathbb{R}^{n \times n}$  is  the diagonal degree matrix with \(D_{i i}=\Sigma_{j} A_{i j}\).


\paragraph*{Chebyshev Polynomials Approximation} In order to localize the filter and reduce the algorithm complexity, the kernel $g$ can be restricted to a polynomial of $L$: \(g(L)=\sum_{k=0}^{K-1} \theta_{k} L^{k}\), where $\theta \in \mathbb{R}^K$ is a vector of polynomial coefficients. $K$ is the kernel size and indicates the area of the convolution in the neighbourhood. The  Chebyshev polynomial $\mathcal{T}_k (x)$ is used to approximate kernels as a truncated expansion of order $K-1$ \cite{hammond2011wavelets}:
$$
g * \mathcal{G} v=g(L) v \approx \sum_{k=0}^{K-1} \theta_{k} \mathcal{T}_{k}(\tilde{L}) v,
$$
where \(\tilde{L}=2 L / \lambda_{\max }-I_{n}\), $\lambda_{max}$ is the largest eigenvalue of $L$, and $I_n$ is an identity matrix.

\paragraph*{First-order Approximation}
Kipf and Welling propose a first-order approximation of spectral graph convolution \cite{7kipf2016semi} based on the assumption of \(\lambda_{\max } \approx 2\) and $K=2$. They argue that deep neural networks are capable of reconstructing spatial information in depth so that the explicit parameterization given by the polynomials can be avoided. Such convolution operation is parameter-economic and can be applied for large-scale graphs.  We can further combine the two parameters $\theta_0$ and $\theta_1$ by assuming \(\theta=\theta_{0}=-\theta_{1}\), which gives us:
$$
\begin{aligned} g *_\mathcal{G} v & \approx \theta_{0} v+\theta_{1}\left(\frac{2}{\lambda_{\max }} L-I_{n}\right) v \\ & \approx \theta_{0} v-\theta_{1}\left(D^{-\frac{1}{2}} A D^{-\frac{1}{2}}\right) v \\ &
= \theta (I_n + D^{-\frac{1}{2}} A D^{-\frac{1}{2}}) v .
\end{aligned}
$$

\paragraph*{DCRNN} the DCRNN operator $* \mathcal{D}$ over a graph signal $v \in \mathbb{R}^n $ and adjacency matrix $A \in \mathbb{R}^{n \times n} $ with a kernel $g \in \mathbb{R}^n $ is defined by:
$$
 v * \mathcal{D} g = \sum_{k=0}^{K} ( P_{f}^{k} v \sigma_{k, 1}+P_{b}^{k} v \sigma_{k, 2}),
$$
where $K$ is the number of finite steps in the diffusion process, $\sigma\in \mathbb{R}^{K \times 2} $ are the parameters of the filter and $P^{k}$ represents the power series of the transition matrix. For directed graphs, the diffusion process has a forward and backward direction, with forward transition matrix \(P_{f}=A / \operatorname{rowsum}(A)\) and backward transition matrix \(P_{b}=A^T / \operatorname{rowsum}(A^T)\).

\subsection{Forecasting in Spatiotemporal Tasks}
\begin{figure}[]
    \centering \includegraphics[width=0.65\textwidth]{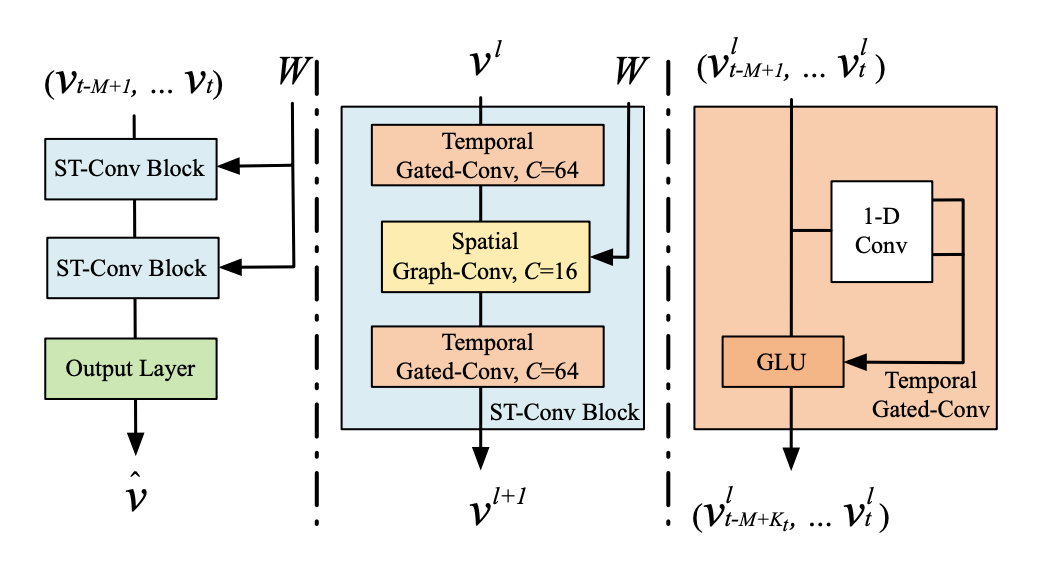}
    \caption{Spatio-Temporal Graph Convolutional Networks \cite{stgc}}
    \label{stgcn}
\end{figure}
Applying a Graph Neural Network (GNN) in spatiotemporal tasks is gaining wide interest. Y. Seo et al. \cite{seo} introduce a Graph Convolutional Recurrent Network (GCRN) to predict structured sequences of data. The proposed model combines graph convolution operations to identify spatial structures and RNN to capture dynamic patterns. The authors found that simultaneously exploiting graph spatial and dynamic information about the data can improve both prediction accuracy and training speed \cite{seo}. Y. Li et al. \cite{li} propose DCRNN for traffic prediction, which captures the spatial dependency using bidirectional random walks on the graph, and the temporal dependency using the encoder-decoder architecture with scheduled sampling. 
In contrast to these works, Spatio-Temporal Graph Convolutional Networks (STGCN) proposed by B. Yu et al. \cite{stgc} is established completely from convolutional structures, and such spatio-temporal architecture is fundamental to state-of-the-art spatiotemporal GNNs. As shown in Figure \ref{stgcn}, the framework consists of two spatio-temporal convolutional layers and a fully-connected output layer, with the residual connection and bottleneck strategy applied inside each block. The right part of the figure shows that the temporal convolutional layer contains a 1-D convolution followed by gated linear units (GLU) as a non-linearity. In Spatial Graph-Conv block, they use spectral graph convolutions proposed by Kipf et al. \cite{kipf2016semi}. This spatiotemporal structure is fundamental to the state-of-the-art spatiotemporal GNNs. Wu et al. proposed Graph WaveNet for road traffic prediction. Similar to STGCN architecture, Graph WaveNet consists of several Spatiotemporal blocks, with each having temporal layer and the spatial layer. They apply gating mechanism on dilated causal convolution for temporal correlation capturing. For spatial layer, they  develop an adaptive dependency matrix and learn the graph hidden  dependencies. The graph convolution layer in  Graph  WaveNet  applies  DCRNN  for  pre-defined  spatial dependencies  and  spectral  graph  convolution  for  adaptive dependency matrix. They also propose that Graph WaveNet can still be effective when the graph topology is unknown by only leveraging spectral graph convolution for adaptive dependency matrix in the graph convolution.

The aforementioned models deal with static graphs which have a constant adjacency matrix, and are not applicable for dynamic graphs where the dependencies between nodes constantly change. Diao et al. \cite{diao2019dynamic} propose a deep learning-based Laplacian matrix estimator to learn the Laplacian matrix at a specific time-of-day dynamically. This model learns the dynamics instead of capturing spatiotemporal correlations from dynamic graphs. 
 Malik et al. \cite{malik2021dynamic} propose  a tensor
algebra framework called  Tensor M-Product to learn representations
of dynamic graphs. They use a  Message Passing Neural
Network for convolution. EvolveGCN \cite{pareja2020evolvegcn} introduces dynamic graph convolution based on Kipf's spectral graph convolution. For the graph signal $x \in \mathbb{R}^{T \times N \times N}$ and the adjacency matrix $A \in \mathbb{R}^{T \times N \times N}$, they adopt spectral graph convolutions along the temporal dimension, acting as $T$ copies of spectral graph convolution with each  one  working  on  one  snapshot  of  the  sequence. The weights across these snapshots are temporal related using a LSTM, which effectively captures the temporal correlations of these snapshots.

Recent work has also explored attention mechanisms on graphs. Zhang et al. \cite{1zhang2019spatial} use a graph attention layer to extract the spatial dependencies and a LSTM network to extract temporal domain features. The works in  \cite{2wu2018graph}  \cite{3zheng2020gman} \cite{4park2019stgrat} harness transformer structures to capture spatial and/or temporal attentions. A transformer consists of an encoder and a decoder, where both the encoder and the decoder consist of multiple attention blocks to model the impact of the spatiotemporal factors. In these works, applying graph embedding is an essential step in order to convert the graph input into a vector. While graph embedding may lose the spatial information stored in graphs, some works \cite{5guo2019attention} \cite{6velivckovic2017graph} do not involve a graph embedding method, where the  attention mechanism is directly applied on the graph structure. Guo et al. \cite{5guo2019attention} use the spatial-temporal attention mechanism to compute spatial score and temporal score, and then combine these scores with the input before applying spectral graph convolutions and common standard convolutions. Additionally, they model three temporal properties of traffic flows, i.e., recent, daily-periodic and weekly-periodic dependencies. Tian et al. \cite{tian2021spatial} propose STAWnet. Similar to WaveNet, it integrates the DCRNN and attention mechanism into an end-to-end framework. By developing a self-adaptive node embedding, STAWnet can capture the hidden spatial relationship in the data without knowing the graph topology (i.e., adjacency matrix). This  model  refines  the  graph  topology automatically in the case that such dependency information is difficult to model or obtain. For the graph with a clearly defined adjacency matrix, STAWnet neglects such important predefined dependency information.

The performance of the GCN model heavily relies on the structure of the input graph \cite{7kipf2016semi}, and it is challenging to 
capture the complex spatial dependency of mobile traffic. In graphs, this dependency is reflected on the adjacency matrix, as $w(i, j)$ means how the traffic from $i$-th base station is related to the traffic from $j$-th base station. Intuitively, this adjacency is modelled as a function of distance between these two base station in \cite{stgc}\cite{li}\cite{fang2018mobile}, and the neighbouring base stations considered in the model were limited to those in location proximity. Such way of modelling the spatial dependency degrades the prediction performance, because as shown in \cite{shafiq2014geospatial}, spatial correlations in traffic also exist between distant base stations. This is partly due to long-distance travelling, or the daily commute of users between residential areas and workplaces, which are typically located separately. For example, users  connected  to  a  base  station in  a  valley  are  unlikely  to  be  handed  off  to  another  one on  the  other  side  of  an  edge,  but  may  be  handed  off  to a  base  station  across  a  stretch  of  water,  given  favourable signal propagation conditions. K. He et al. \cite{he2019graph} adopt the Dynamic Time Warping (DTW) algorithm to measure the traffic sequences similarity between two base stations as the adjacency matrix, because they state that base stations are closely related when they have similar traffic patterns. S. Zhao \cite{zhao2020cellular} consider the effect of handover on the spatial characteristics of the traffic, and build graphs for base stations based on their handover frequencies (i.e., the number of handovers over a period of time). Kalander et al. \cite{kalander2020spatio} use hybrid GCNs via three types of binary adjacency matrices: spatial proximity, functional similarity (i.e., similarity of average weekly traffic using Pearson Correlation Coefficient (PCC)) and recent trend similarity (i.e., constructing a separate graph for each timestamp based on the PCC similarity of the most recent traffic). Such hybrid graph construction requires high memory and becomes infeasible if the number of base stations is large.

\paragraph*{Lessons learned} 
Modelling the complex cellular network as matrices simply assumes the relationships  between  these  stations  remain static over time and ignore that there may be multiple sectors at  a  given  location  and  users  are  often  handed  off  between these, sometimes among different technologies (e.g., from 4G to  3G). Leveraging the graph structure allows these dependencies to be quantified, and such information is vital to forecasting.

Quantifying the dependency is a challenging task. The work in \cite{zhao2020cellular} looks into modelling graphs with handover information, which reflects user commute. For such dynamic graphs, we should also consider the spatiotemporal dependency in the dynamic adjacency matrix. Unfortunately, prior work uses the average adjacency matrix computed from a small set of randomly picked days and operates on static graph convolution. This overlooks dynamics of graph topology and leaves room for improvements.

\section{Summary}
This chapter discussed related work. We first introduced sampling and reconstruction approaches applied in the mobile traffic analysis, as well as how to estimate the reconstruction error. Then we explained basic concepts in Reinforcement Learning, including Q learning, deep reinforcement learning, and Monte Carlo Tree Search, and their applications to mobile traffic analysis, e.g., Spare Mobile Crowdsensing. Lastly, we took a look at Graph Neural Networks and popular graph convolution operations, followed by their applications in spatiotemporal forecasting tasks. Additionally, we summarize several limitations of the state-of-the-art and the opportunities that arise in the area of mobile traffic analysis. In the next chapter we will first focus on the sampling and reconstruction framework, where we propose a deep learning-based approach and show it outperforms a set of benchmarks.

\chapter[Deep Learning-driven Sparse Mobile Traffic Measurement ...]{Deep Learning-driven Sparse Mobile Traffic Measurement Collection and Reconstruction}


Precise traffic monitoring is costly due to dedicated equipment, substantial storage capability and transfer overhead. As discussed before, traditional sampling and reconstruction approaches are based on the assumption that the measurement collection points are fixed and randomly selected, which overlooks spatiotemporal  correlations  tailored  to  mobile  traffic. Therefore, it is vital to developing a solution to an intelligent selection policy and precise mobile traffic reconstruction, so that the cost of data collection is reduced while high reconstruction accuracy is retained.

This chapter describes our first contribution.
We propose \name, a deep learning-driven mobile traffic measurement collection and reconstruction framework for infrastructure level data. \name relies on a dedicated neural network that we train to selectively sample small subsets of target mobile coverage areas. It then employs a purpose-built mobile traffic reconstruction neural model, which exploits spatiotemporal correlations in historical data, to infer mobile traffic consumption across the entire deployment. Our framework reduces dramatically the cost of data collection while retaining high traffic consumption inference accuracy. In summary, we: 

\begin{enumerate}[leftmargin=*,label={(\arabic*)}]
    \item We introduce a policy network that takes as input sparse historical traffic snapshots and outputs cell selection matrices indicating at which locations to activate measurement collection, so as to minimize overhead while acquiring enough data to ensure high-quality traffic consumption interpolations at all non-sampled cells. We take a Deep Reinforcement Learning (DRL) approach to produce examples for training this policy network. Given the large action space, our DRL agent learns in a tractable manner by sampling small subsets of the action space based on the most likely action and its nearest neighbours. The highest valued action from such subsets is then selected, circumventing the need to evaluate the entire action space.
    \item We propose MTRnet, a deep neural network specifically tailored to mobile traffic reconstruction from sparse traffic measurements. This model outperforms existing traffic reconstructions methods both in terms of accuracy (up to 67\% lower mean absolute errors) and reconstruction speed (up to 835$\times$ runtime reduction).
    \item We evaluate our \name framework using a real-world mobile traffic dataset collected by a major European operator, showing that our solution learns to effectively reconstruct complete traffic matrices from sparse samples, even when applied to previously unseen traffic patterns such as those observed during holidays. On average, \name accurately reconstructs city-scale traffic snapshots with up to 48\% fewer samples than a range of  benchmarks considered. 
\end{enumerate}

\section{Problem Formulation}
Consider a mobile network coverage area that is geographically partitioned into a grid with $X\times Y$ squares (cells), where a square denotes an atomic region for mobile traffic collection. We denote by $F_t$ the traffic consumption snapshot across all cells, observed over an interval $[t-\Delta, t]$, i.e.,
\setlength{\belowdisplayskip}{3pt} \setlength{\belowdisplayshortskip}{0pt}
\setlength{\abovedisplayskip}{3pt} \setlength{\abovedisplayshortskip}{0pt}
\[
F_t = 
\begin{bmatrix}
d^t_{1,1} & d^t_{1,2} & \ldots & d^t_{1,Y}\\
d^t_{2,1} & d^t_{2,2} & \ldots & d^t_{2,Y}\\
\vdots & \vdots & \ddots & \vdots \\
d^t_{Y,1} & d^t_{Y,2} & \ldots & d^t_{X,Y}\\
\end{bmatrix},
\]
where $d^t_{i,j}$ is the volume of traffic at cell $(i, j)$,
and $\Delta$ is the time granularity configurable by a network administrator, with which traffic measurement equipment (probes) compute summaries of the volume of traffic observed at different locations.
A mobile network operator will need a sampling strategy that gives a binary selection matrix ${B}_t$ indicating which cells are to be selected for measurement collection at time $t$, i.e., $b^t_{i,j}= 1$, if cell $(i,j)$ is selected; $b^t_{i,j}=0$, otherwise; and subsequently use the measurements collected to infer the traffic consumption across the entire deployment. A selection matrix  produces a sparse measurement matrix $M_t$ of a network traffic snapshot $F_t$, i.e. 
$$
M_{t} = F_{t} \circ B_{t},
$$
where $\circ$ is the Hadamard product. 
Let $\mathbb{M}_t = [M_{t-T+1}, ..., M_t]$ denote the sparse measurement matrices collected over $T$ recent timestamps, and $\hat{F_t}$ the reconstruction of $F_t$ given $\mathbb{M}_t$. 

The problem we seek to solve is thus twofold: \textit{(i)} finding a policy $g(\cdot)$ that outputs cell selection matrices $\widetilde{B}_t=g(\mathbb{M}_{t})$, which contain minimum numbers of elements $b^t_{i,j}$ set to 1, so that the error of reconstructing $F_t$ using those measurements collected is below a predefined threshold $\epsilon$, by using \textit{(ii)} a data interpolation algorithm $f(\cdot)$. Formally,  
\begin{align}
&\widetilde{B}_t := \mathop{\arg\min}\limits_{B_t}\sum_{i, j} b^t_{i, j}, \label{eq:objective}\\
\text{s.t.} &  \text{ MAE}(\hat{F_t}, F_t) < \epsilon,\\
&\hat{F_t} = f(\mathbb{M}_{t}).\label{eq:reconstruct}
\end{align}
In the above MAE is the Mean Absolute Error, i.e.,
\begin{equation}
    \text{MAE}(\hat{F_t}, F_t)=\frac{1}{X\times Y}\sum_{i,j}|\hat{d}^t_{i,j}-d^t_{i,j}|.
\end{equation}

We solve this problem using a set of purpose-built deep neural network models, as we explain next.

\begin{figure}[]
	\centering
\includegraphics[width=0.8\columnwidth]{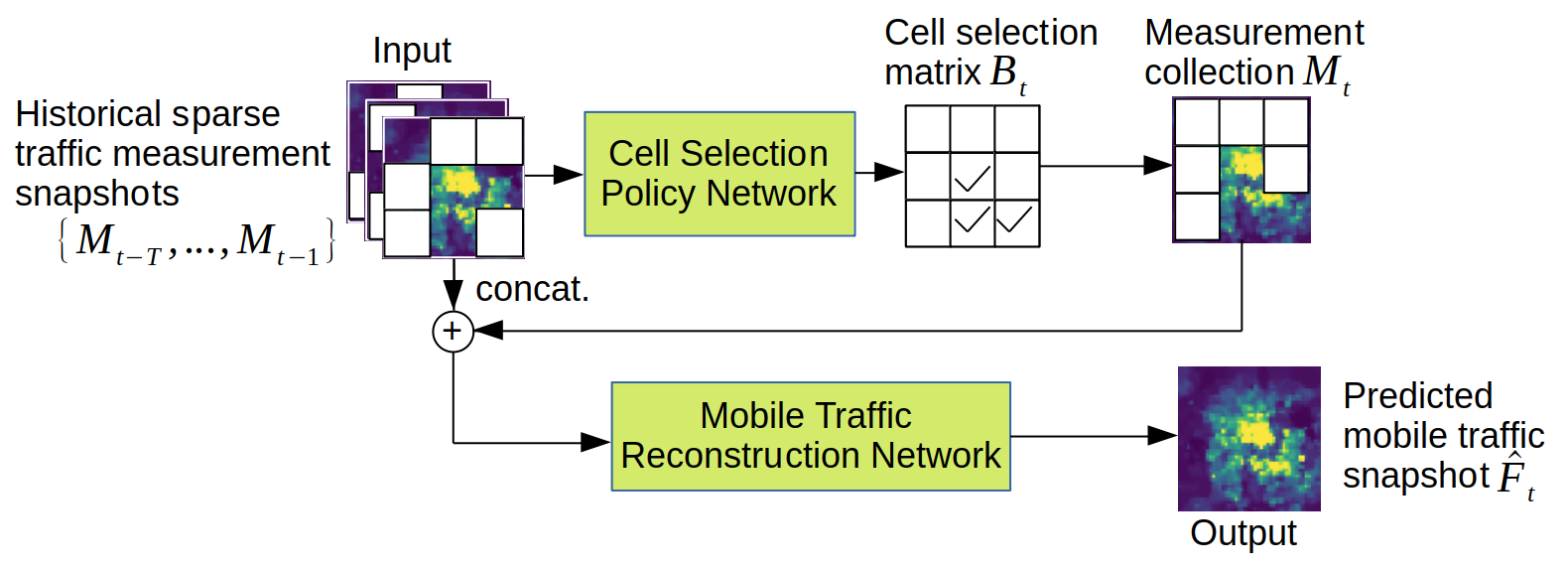}
	\caption{
	Overview of the proposed Spider framework: sparse mobile traffic snapshots used by a policy network to select optimal cells where to collect measurements; a dedicated reconstruction neural model outputs completed network traffic snapshot based on historical and current sparse measurements.
	}
	\label{illu}
\vspace*{-1em}
\end{figure}

\section{Our Framework: Spider}
To solve the problem defined by (\ref{eq:objective})--(\ref{eq:reconstruct}), we propose \name, an original mobile network traffic measurement collection and reconstruction framework that \textit{(i)} predicts a small number of optimal locations where traffic should be sampled, and \textit{(ii)}~learns to infer the volume of traffic at locations not selected, so as to reconstruct complete snapshots of the traffic consumption across an entire network deployment. We give a diagrammatic view of our approach in Figure \ref{illu}. We design original neural models that harness the unique characteristics of mobile traffic to solve both of these tasks. 

\subsection{RL for Cell Selection with Large Action Spaces}
\label{sec:agent}

To collect measurements with minimum sampling overhead, we first leverage Reinforcement Learning (RL) and train an agent that learns the likelihood of selecting a cell where to sample traffic, based on past experience. We subsequently use this agent to train a policy network that \textit{directly outputs optimal cell selection matrices}, given prior observations, and thus is suitable for real-time decision-making. 

We start by regarding cell selection as an episodic task, which can be modelled as a Markov Decision Process (MDP), $\mathcal{M} := (S, A, P, r, \gamma)$, where
\begin{itemize}
    \item $S$ is the set of states $s_i$, in our case a state representing a collection of sparse measurement matrices of the past $T$ network traffic snapshots, i.e., $ s_i = \mathbb{M}_t^i = [M_{t-T+1}, ..., M_t^{i{-1}}]$, with $M_t^{i{-1}}$ denoting the sparse measurement matrix at iteration $i -1$ of an episode $t$;
    \item $A$ is the set of possible actions, where an action $a_i$ corresponds to sampling one of all the previously unselected cells at timestamp $t$, from which the agent can choose;
    \item $P(s, a, s^\prime)$ is the probability that action $a$ in state $s$ will lead to next state $s^\prime$; 
    \item $r(s, a, s^\prime)$ is the reward the agent receives as a consequence of choosing action $a$ when in state $s$; we work with $r(s,a,s^\prime) = -\text{MAE}(f(s^\prime),F_t)$ to incentivize the agent to take actions that reduce the reconstruction error; while a range of methods $f(\cdot)$ can be employed for reconstruction, including compressive sensing or K-Nearest Neighbour-based interpolation, our \name framework evaluates how good an action is using a pre-trained Mobile Traffic Reconstruction neural Network (MTRNet), which we detail in Sec. \ref{sec:mtrnet}. 
    \item $\gamma \in [0,1]$ is the discount rate -- an agent's objective is to maximize a cumulative reward it receives (expected return), i.e. the sum of discounted rewards $r\gamma^{k-1}$ over $k$ iterations; we work with $\gamma=0$, meaning that at each iteration we seek to maximize the immediate return. The immediate return is inversely proportional to the reconstruction error and reflects directly the value of the action, so we are not concerned about the cumulative reward.
\end{itemize}
We consider a cell selection episode $t$ to be completed at an iteration $I$ when the inference error is less than a predefined reconstruction quality threshold $\epsilon$.

\begin{figure}
 \centering
    \subfloat[\centering ]{{\includegraphics[width=0.3744\textwidth]{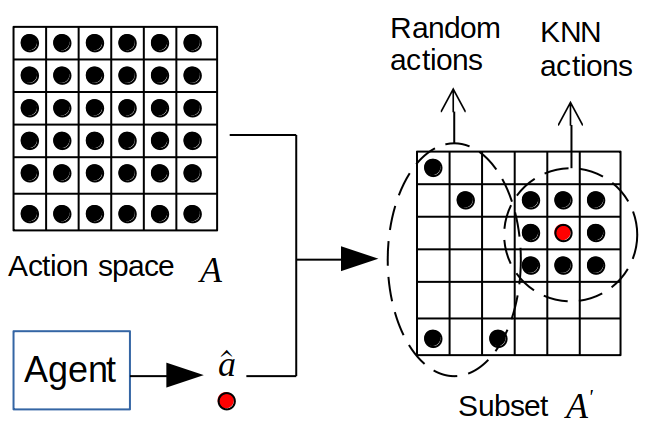} }}%
    \qquad
     \subfloat[\centering ]{{\includegraphics[width=0.2448\textwidth]{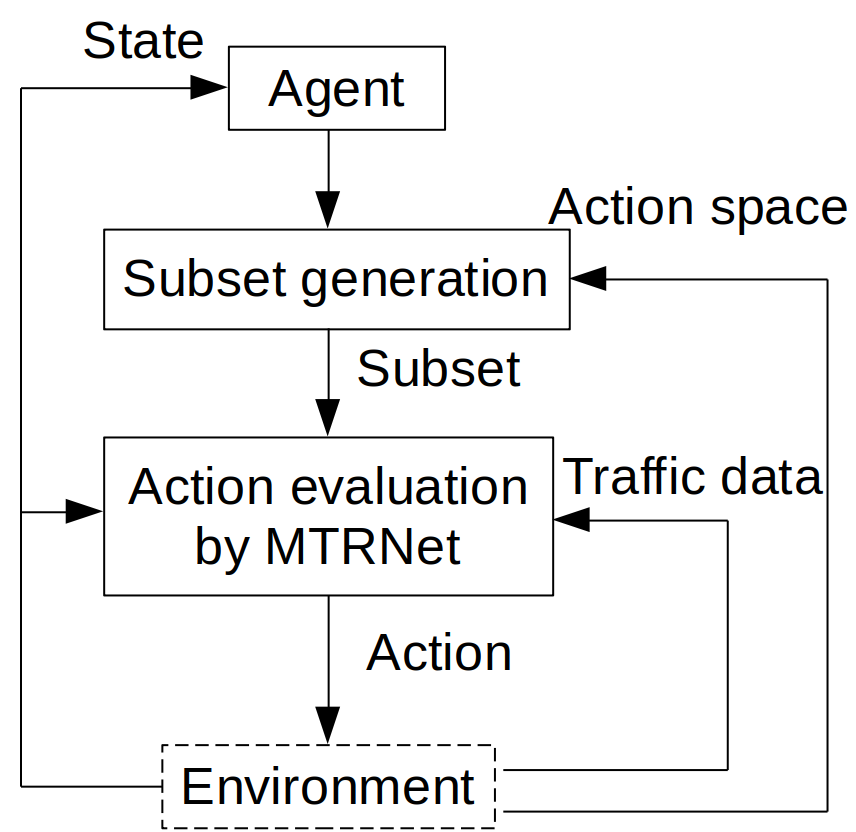} }}%
    \caption{(a.) Creating a subset from the large action space (b.) Reinforcement learning with MTRNet}%
     \label{subset_creation}%
 \end{figure}

\begin{figure}[]
	\centering
\includegraphics[width=0.6\textwidth]{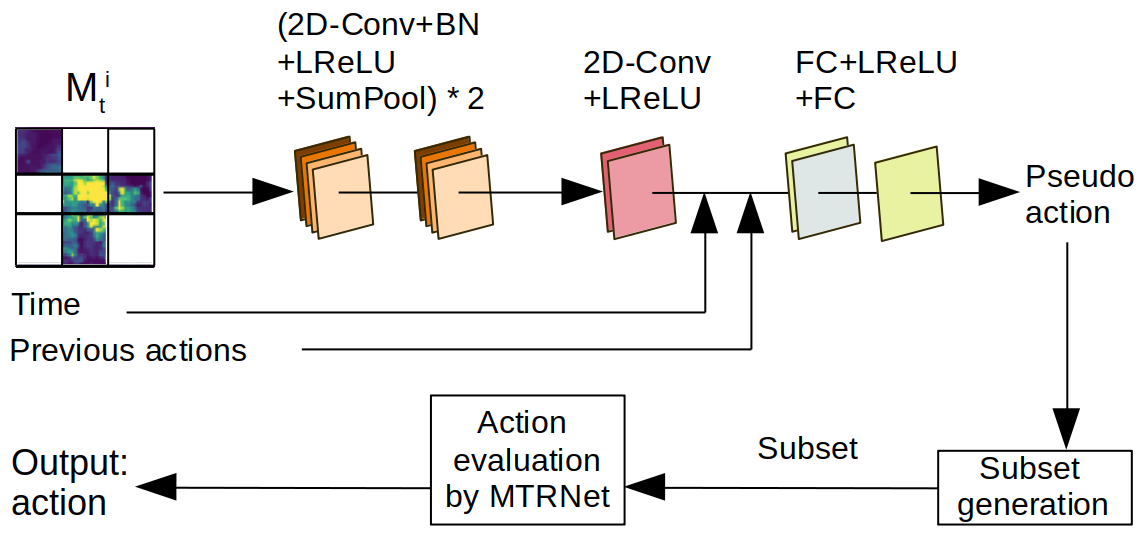}
\caption{
{Structure of the DRL agent. The pseudo action given by the neural network is used to generate a set of potential candidates that, when evaluated by MTRNet, allows the agent to select the action that minimizes the reconstruction error.}
}
	\label{policyn}
\end{figure}

{With traditional DRL,} finding a policy that maps from states to probabilities of choosing an action in a large action space such as city-scale mobile network deployments comprising thousands of cells
would require massive amounts of memory. Further, exploring the action space exhaustively, until sufficient  cells are selected to yield usable reconstructed traffic snapshots, would involve impractically large runtimes.
To address these issues, we design a DRL agent based on a neural network architecture to model the policy function, which we train using an efficient algorithm that only evaluates a subset of potentially good actions from the whole action space, among which the best action is selected following evaluation.
As such the agent will first output a pseudo action $\hat{a}$, which is used to generate a subset of $k$ potential actions by finding the K-Nearest Neighbors of $\hat{a}$ in $A$ based on $\ell_2$ distance, i.e.,
$$
{A^\prime_k =}\ \operatorname{knn}(\hat{a})=\underset{{a} \in \mathcal{A}}{\overset{k}{\arg  \min }}|| {a}-\hat{{a}}||_{2},
$$
because the spatial correlation between two cells is related to their distance. To explore previously unseen actions, the generated subset $A^\prime_{k}$ is expanded with $\eta$ randomly selected actions {${rand}_\eta(A^\prime_k)$ not already in $A^\prime_k$, where $\eta$ is given by} an exponentially decaying function, i.e.
$$
{A^\prime := A^\prime_{k - \eta} + rand_{\eta}(A^\prime_{k - \eta})},
$$
where $\eta = \lfloor k (0.1 + 0.9 / e^x) \rceil$
and $x$ is the number of training episodes completed. This approach reduces both memory requirements and execution time by constraining the search space, and favours exploration at the beginning, when 100\% of the actions are chosen randomly. We demonstrate this subset creation process in Figure \ref{subset_creation} (a). We illustrate the DRL agent's structure in Figure \ref{policyn} and present the training process in Algorithm \ref{alg:policy_training} and Figure \ref{subset_creation} (b).

The agent takes as input the sparse measurements snapshot $M_t^{i-1}$ built up to the current iteration $i$, the list of cells selected so far,
and the time $t$.
From the sparse measurements matrix, the neural network first extracts feature maps using 2D convolution layers, with a sum pooling layer applied between them to summarize these feature maps without diluting the active features. This enables faster learning of the traffic patterns observed within a geographical window. The resulting features are then fed to a 2D convolution layer to reduce the size of the aggregate feature map, such that each feature's magnitude is of the same scale as the time and  previous actions. Two fully connected layers then process the aggregate feature map, along with the time and the previous actions, to predict the next action. Based on this value, the agent generates a set of potential candidates to be evaluated by MTRNet (see Algorithm~\ref{alg:policy_training}, lines \ref{ln:evalstart}--\ref{ln:evalstop}), and the candidate yielding the smallest reconstruction error is chosen by the agent as next action (line \ref{ln:proto}). Finally, the neural network learns to improve its predictions by updating its weights based on gradients of the Mean Squared Error loss between the predicted (pseudo) action and the chosen candidate (lines \ref{ln:loss}--\ref{ln:gradient}). For any given snapshot, this process is repeated until the reconstruction error is less than a predefined reconstruction quality threshold $\epsilon$. 

\begin{algorithm}[t!]
\small
\SetAlgoLined
 Input: Pre-trained MTRNet $f()$, environment $E$, the number of epochs $n_e$, the number of snapshots in training data $n_s$, the number of historical snapshots used $T$\;
 Initialize the neural network $p()$\;
 \For{$epoch\gets1$ \KwTo $n_e$}{
    Initialize $E$\;
    \For{{$t \gets T+1 $ \KwTo $n_s$}}{
    
    \While{episode not finished}{
    Get current state $s_i = [M_{t-T}, ..., M_t^{i-1}]$, previously selected actions $\mathbb{A}$, and the ground truth snapshot $F_t$ from $E$ for the current iteration $i$\;
    Generate subset $A^{\prime}$ from $A$ and $p$($M_t^{i-1}$, $t$, $\mathbb{A}$) \;
    $e=\{\}$\;
    \For{$a$ in $A^{\prime}$}{ \label{ln:evalstart}
    Get $s_j$ by applying $a$ to $s_i$\;
    $e_j= MAE (f(s_j), F_t)$\;
    Push $e_j$ to $e$\;  
    }\label{ln:evalstop}
    $
{a^{\prime}}=\underset{a \in \mathcal{A^{\prime}}}{{\arg  \min }} (e) 
$  \; \label{ln:proto}
    Apply action ${a^{\prime}}$  to $E$\;
   {$Loss = \norm{p(M_t^{i-1}, t, \mathbb{A}) - a^{\prime})}_2^2 $} \; \label{ln:loss}
    Use $Loss$ to update $p()$ by gradient descent\; \label{ln:gradient}
    }
    }
    
    }
 \caption{Training the RL agent}
 \label{alg:policy_training}
\end{algorithm}

In choosing this threshold, we seek a trade-off between achieving low reconstruction errors and selecting a small number of cells. To find an appropriate quality threshold, we examine the MAE reduction as a function of (random) cell sampling rates and observe that MAE decreases only marginally beyond 35\% sampling rates (detailed explanation will be given later in Sec. \ref{choice}). Hence, we work with this value as the quality threshold.

When predicting the pseudo action, the agent only considers the current snapshot instead of the current state. This is because in choosing actions one by one, only one value in a matrix changes from zero to the measurement between two consecutive states, which would make the collected value less important in the input, wrt. historical data. Although the previous experience is not seen in the pseudo action prediction, MTRNet evaluates an action given the full state as the input, whereby temporal correlations are captured.

\begin{figure}[]
	\centering
\includegraphics[width=0.8\textwidth]{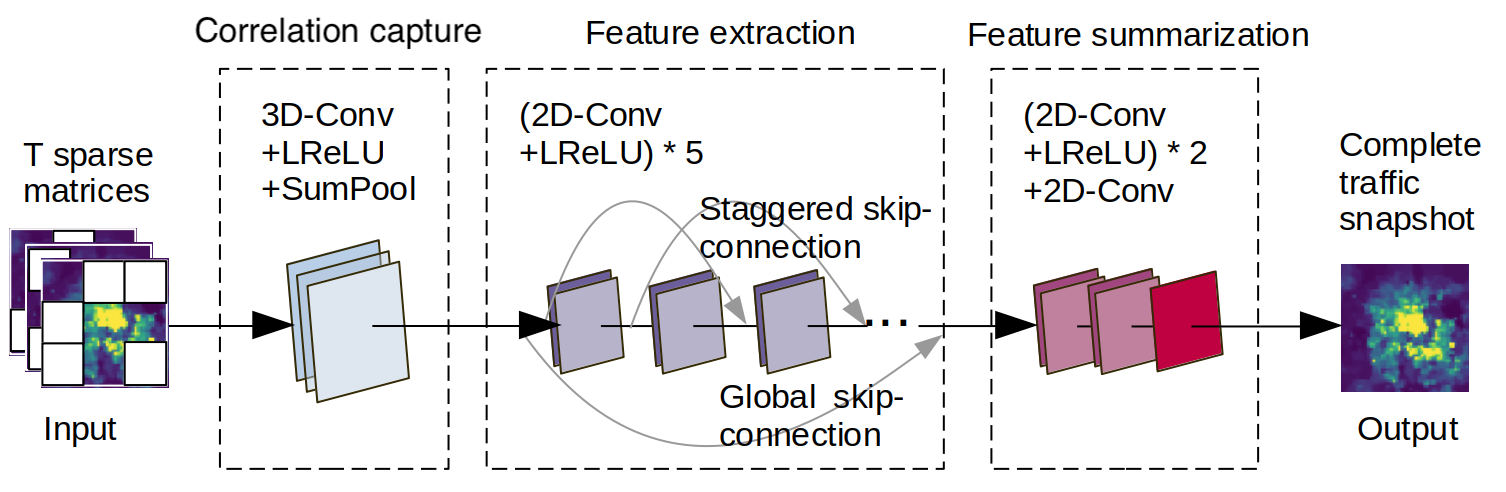}
\caption{MTRNet structure comprising correlation capture, feature extraction \& summarization functionality. Traffic snap\-shots reconstructed based on historical sparse \mbox{measurements.}}
	\label{MTRNet2}
	\vspace*{-1em}
\end{figure}

\subsection{Traffic Reconstruction from Sparse Measurements} \label{sec:mtrnet}
We perform traffic measurement reconstruction using a supervised learning approach. The proposed Mobile Traffic Reconstruction neural Network (MTRNet) takes as input the sparse measurement matrices for the current ($t$-th) and previous $T$ timestamps, i.e., $[M_{t-T}, ..., M_t] $, and outputs the traffic consumption of the full map $\hat{F}_{t}$ at the current timestamp.  

The model architecture is shown in Figure~\ref{MTRNet2}. We draw inspiration from ZipNet \cite{zipnet}, a mobile traffic super-resolution technique that infers fine-grained  traffic consumption from coarse measurements. The original ZipNet was modified in order to reduce the size of the model and improve its inference speed. Specifically, since in our setting the size of the  sparse measurement matrices is the same as the size of the output, we discard upscaling blocks. Further ablation study allows us to reduce the number of layers of the model (depth) while retaining high accuracy. 
The resulting MTRNet comprises three
key components: 
\begin{itemize}[leftmargin=*]
    \item \textbf{Correlation capture.} A 3D convolutional layer is used to capture spatiotemporal correlations between recent snapshots. The activation layer is a Leaky-ReLU (LReLU) that improves the model's non-linearity and its  robustness to network initialization. LReLU is defined by:
$$
\operatorname{LReLU}(x)=\left\{\begin{array}{ll}x, & x \geqslant 0 \\ \alpha x, & x<0\end{array}\right. ,
$$
where $\alpha$ is a configurable slope value. A SumPooling layer is applied after LReLU to reduce the size of the feature maps and speed up the training.
    \item \textbf{Feature extraction.} Several 2D Convolutional layers followed by a LReLU extract high level abstract features from the geographical configuration of the measured traffic, thereby enhancing the representability of the model. Staggered skip connections link every two blocks, and a global skip connection links the input and output of this component. These skip connections preserve different features extracted from previous layers in a hierarchical way, enabling feature reusability and stabilizing training and convergence. 
    \item \textbf{Feature summarization.} Three convolutional layers are used to summarize the features extracted in previous layers and output the final reconstruction result. The number of channels increases with each of these convolutional layers accurate reconstruction.
\end{itemize}

\subsection{Policy Network}
\begin{figure}[]
	\centering
\includegraphics[width=0.8\textwidth]{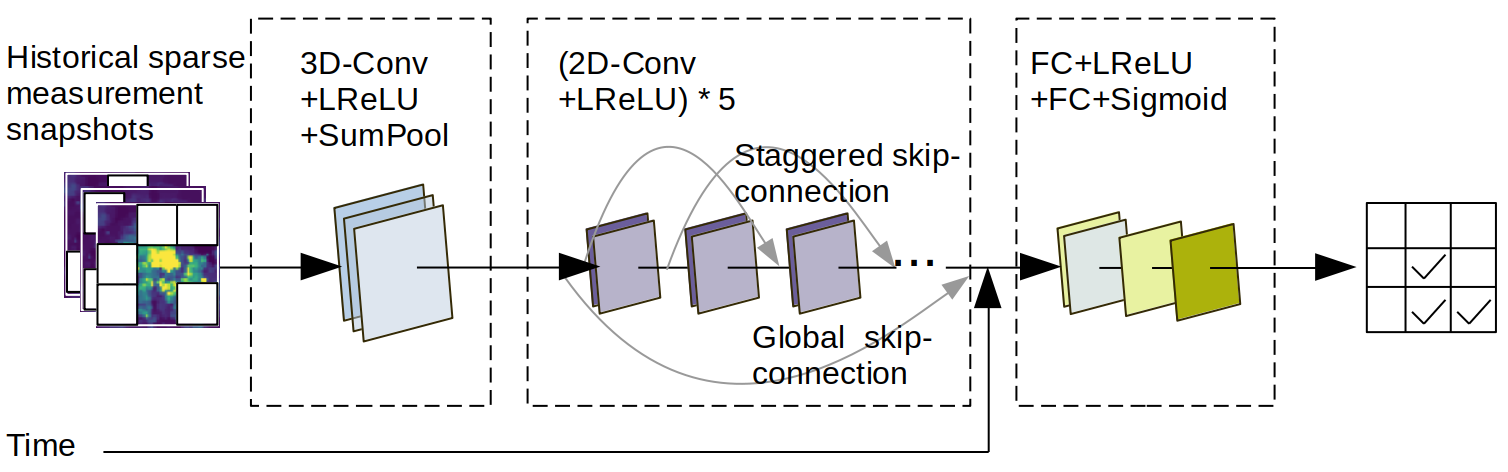}
\caption{Structure of policy network. It predicts all optimal cell locations to be sampled at once.}
	\label{policy}
	\vspace*{-1em}
\end{figure}

Since RL agents typically evaluate one action at a time, they are often impractical to deploy in settings where decisions have to be made within short deadlines. Therefore, we design a policy network that \textit{predicts all optimal cell locations to be sampled, $B_{t} = g(\mathbb{M}_t)$, at once}. This policy network ($g$)
takes historical sparse measurement snapshots and timestamp information as input and learns to predict binary selection matrices generated by the agent introduced in Sec. \ref{sec:agent} at the end of each episode.

This neural network effectively solves a multi-label classification task where positive labels represent cells to be selected, and negative labels indicate non-selected squares. The model architecture is similar to that of MTRNet (shown in Figure \ref{policy}), though here a final Sigmoid layer transforms the outputs into probability scores, and we assign positive labels to those selection matrix elements where the probability is above 0.5, and negative to all others. As the cell selection frequencies are likely to be skewed, the probabilities are normalized to the range $[0, 1]$ before the binarization process.

We use a Binary Cross-entropy (BCE) loss function to train this neural model, which is defined by $BCE =  -\frac{1}{N} \sum_{i=0}^{N} y_i \log(\hat{y_i}) + (1-y_i) \log(1-\hat{y_i})$, where $y_i$ is the value in the binary selection matrix, and $\hat{y_i}$ is the output probability score. $N$ denotes the total number of elements (cells).

\section{Performance Evaluation}

\begin{figure}[]
	\centering
\includegraphics[width=0.9\textwidth]{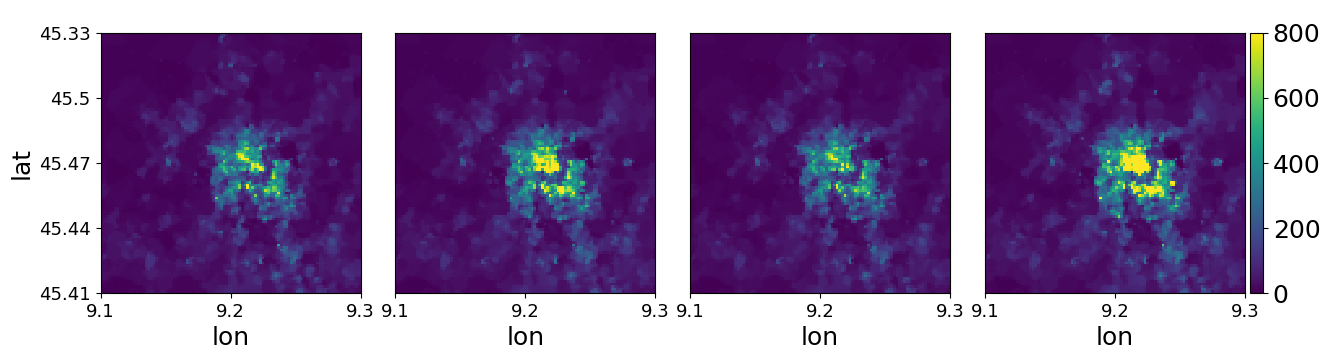}
\caption{Average traffic consumption of one snapshot in weekend, weekdays, off-peak, and peak (9am-6pm), respectively.}
	\label{traffic}
\end{figure}

\begin{figure}%
    \centering
    \subfloat[\centering ]{{\includegraphics[width=0.35\textwidth]{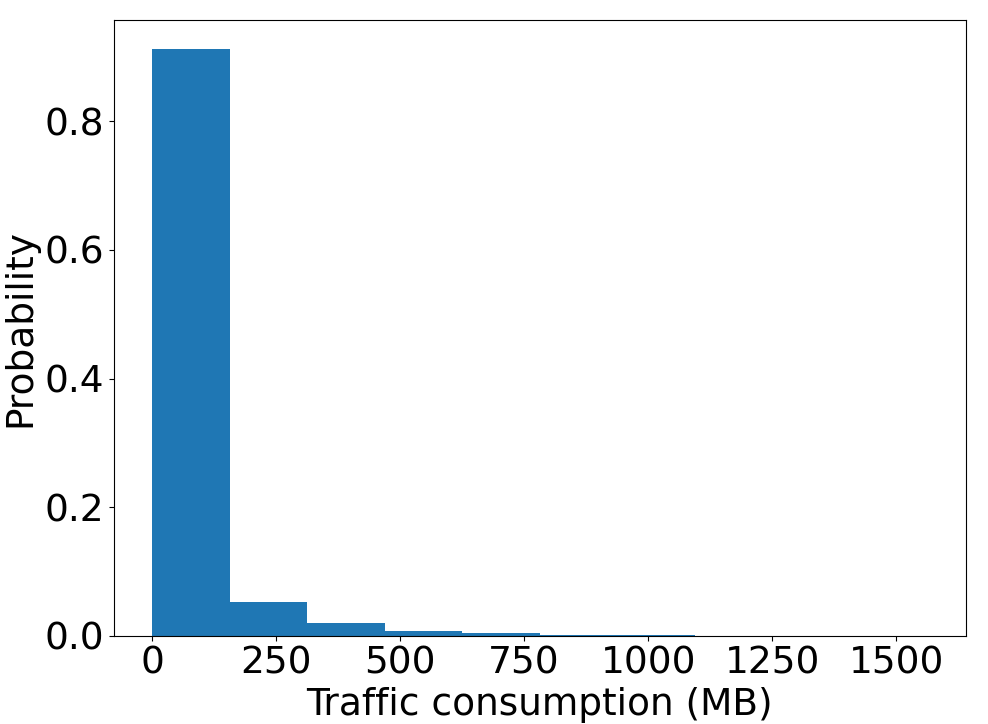} }}%
    \qquad
    \subfloat[\centering ]{{\includegraphics[width=0.35\textwidth]{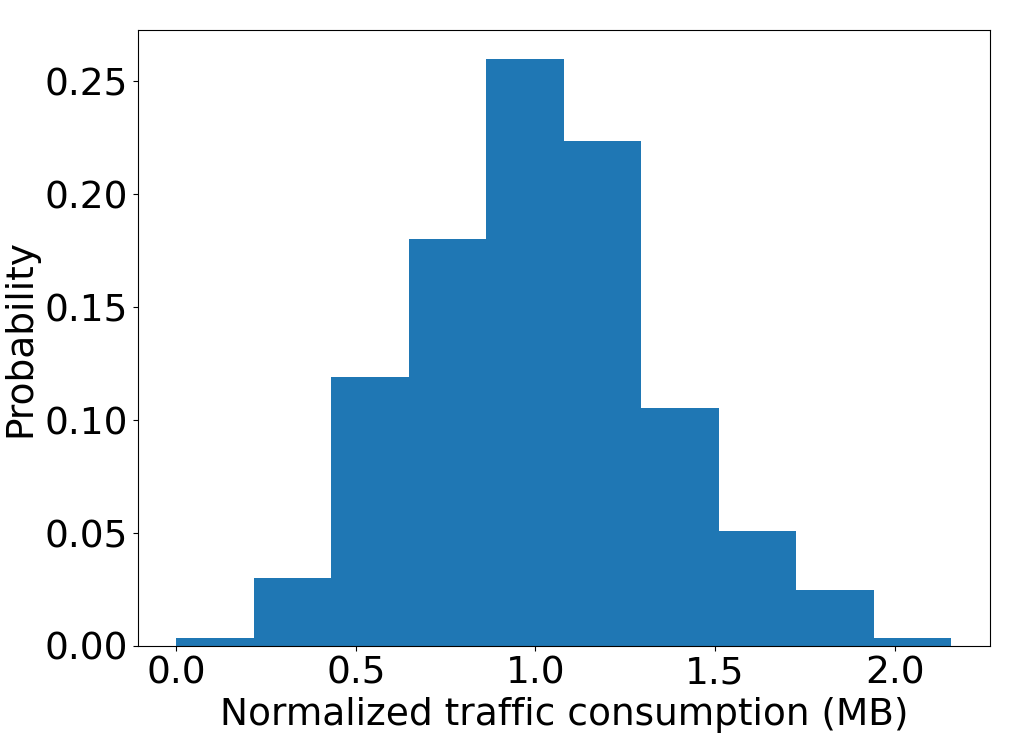} }}%
    \caption{Distribution of traffic consumption before (a.) and after (b.) normalization}%
    \label{dis}%
\end{figure}

\subsection{Dataset}
For evaluation, we adopt a real-world mobile traffic dataset collected by Telecom Italia in the city of Milan \cite{milano}. The city is divided into 100$\times$100 squares of 0.055 km$^2$. Mobile data traffic measurements were collected on aggregate at these locations every 10 minutes between 01/10/2013 and 01/01/2014 (3 months). We use 40 days worth of data to train the models, and 20 days for testing. Figure \ref{traffic} shows the heatmaps of average traffic consumption of one snapshot in weekend, weekdays, off-peak, and peak (9am-6pm), respectively, from left to right.

The distribution of traffic consumption is skewed, as shown in Figure \ref{dis} (a). We normalize the traffic consunmption by $x=\log(1 + x) / \Bar{x}$, where $x$ is the traffic consumption and $\Bar{x}$ is the mean of $\log(1 + x)$. This results in a normal distribution, as shown in Figure \ref{dis} (b).

\subsection{Implementation}
We use a compute cluster comprising 20+ nodes, each equipped
with 1-2 NVIDIA TITAN X GPUs (2280 cores) to train the neural models. We implement \name in Python using the PyTorch library, and train on 2 GPU nodes in parallel for 2 epochs, spending 9.6 hours on each GPU. The number of historical snapshots $T$ is 6. For all the models, we employ the Adam optimizer with learning rate \mbox{$\lambda = 10^{-4}$} and use 128 as the batch size. For MTRNet, the number of previous actions is 20 and the number of 2D Convolutional layers is 5.

\subsection{Choice of metrics}
We choose to use Normalized Mean Absolute Error (NMAE) and Mean Absolute Error (MAE) as the evaluation metrics. These metrics show the absolute difference between predictions and actual traffic values, and have a better interpretation of the model error compared to Root Mean Square Error (RMSE) and Mean Square Error (MSE). MAE is also used as loss function during model training.

\subsection{Results: measurement and collection}

We first examine \name's performance in terms of the average number of cells selected for measurement collection versus (1) a random selection strategy that chooses randomly which cells to sample, with their number being the average at any time of the day during different days of the week that yields reconstruction errors below the predefined threshold, as observed across the dataset used for training our agent; and (2) a strategy that chooses the most frequently selected cells based on past selection patterns. This frequency matrix is generated by averaging binary selection matrices obtained during training for the same time of the day and day of the week. To put things into perspective, we also compute the NMAE obtained with our MTRNet when each of these strategies are employed.  

We summarize the performances of
\name against the baselines considered in Table \ref{performance_analysis}, comparing NMAE and sampled cell count (rounded to the nearest integer) at peak and off-peak hours during a day, respectively during weekdays, weekends, and public holidays. Overall, \textbf{\name samples 48.0\% fewer cell than the random selection strategy and 32.4\% fewer than the selection approach based on historical data}, at a negligible cost in terms of additional reconstruction error introduced relative to the average volume of traffic (NMAE).

\begin{table}[]
\caption{Performance comparison between \name and cell selection baselines, in terms of number of cells sampled and resulting NMAE following MTRNet-based reconstruction.} \label{performance_analysis} 
\centering
\setlength{\tabcolsep}{5pt}
\begin{tabular}{|l|l|l|l|l||l|l|}
\hline
\multirow{2}{*}{Interval} & \multicolumn{2}{l|}{\begin{tabular}[c]{@{}l@{}}Random cell\\ selection\end{tabular}} & \multicolumn{2}{l||}{\begin{tabular}[c]{@{}l@{}}Historical data-\\ based selection\end{tabular}} & \multicolumn{2}{l|}{\begin{tabular}[c]{@{}l@{}}\textbf{\name}\end{tabular}} \\ \cline{2-7} 
                  & Count                                     & NMAE                                   & Count                                      & NMAE                                    & Count                                     & NMAE                                   \\ \hline \hline
Peak              & 4,737                                   & 0.06                                  & 4,185                                    & 0.04                                  & 2,643                                 & 0.08                                  \\ \hline
Off-peak          & 3,610                                   & 0.09                                 & 2,472                                    & 0.10                                  & 1,814                                   & 0.12                                 \\ \hline
Weekdays          & 4,116                                   & 0.08                                 & 3,142                                   & 0.07                                   & 2,378                                   & 0.09                                 \\ \hline
Weekend           & 3,838                                   & 0.08                                  & 3,033                                   & 0.07                                  & 2,237                                   & 0.09                                \\ \hline
Holiday           & 4,049                                   & 0.07                                  & 3,129                                   & 0.08                                  & 1,906                                   &0.11                               \\ \hline
Overall             & 4,039                                   & 0.08                                 & 3,122                                    & 0.08                                  & 2,109                                   & 0.10                                  \\ \hline
\end{tabular}
\vspace*{-1em}
\end{table}

We illustrate \name's behaviour and that of the benchmarks considered over an entire week (Monday to Sunday) using traffic from the testing set (red line) in Figure \ref{overall}. Observe that \name captures accurately the changes in traffic demand and samples cells for measurement collection accordingly. While the historical data-based approach is able to distinguish between different times of the day and days of the week, it clearly over-samples, thus incurring higher measurement collection overhead. This is even more noticeable when cells are sampled randomly and with the only goal of meeting the quality threshold. 

\begin{figure*}[ht]
	\centering
\includegraphics[width=1.2\textwidth]{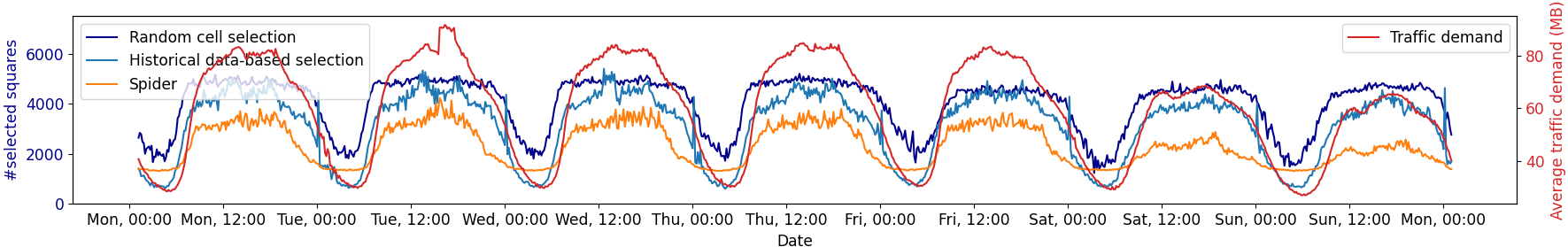}
\caption{Number of cells selected for measurement collection by Spider and  benchmarks, over one week (data representative for 16--22 Dec 2013). Average volume of traffic also plotted to highlight Spider's ability to adapt to changing patterns.}
	\label{overall}
\end{figure*}

\begin{figure}[]
	\centering
\includegraphics[width=0.8\textwidth]{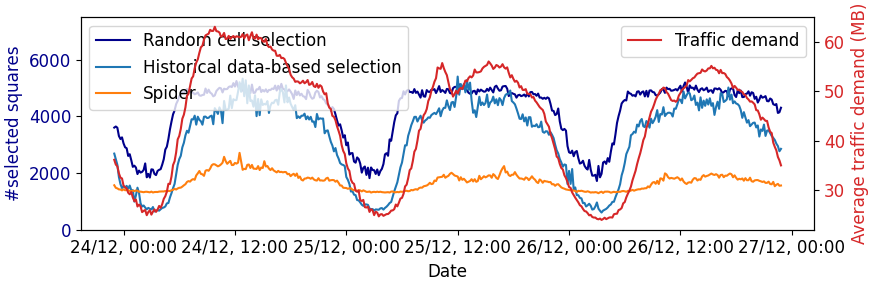}
\caption{Number of cells sampled by \name and the benchmarks considered, during Christmas holidays.}
	\label{res_holi}
	\vspace*{-1em}
\end{figure}

\begin{figure}[]
	\centering
\includegraphics[width=0.8\textwidth]{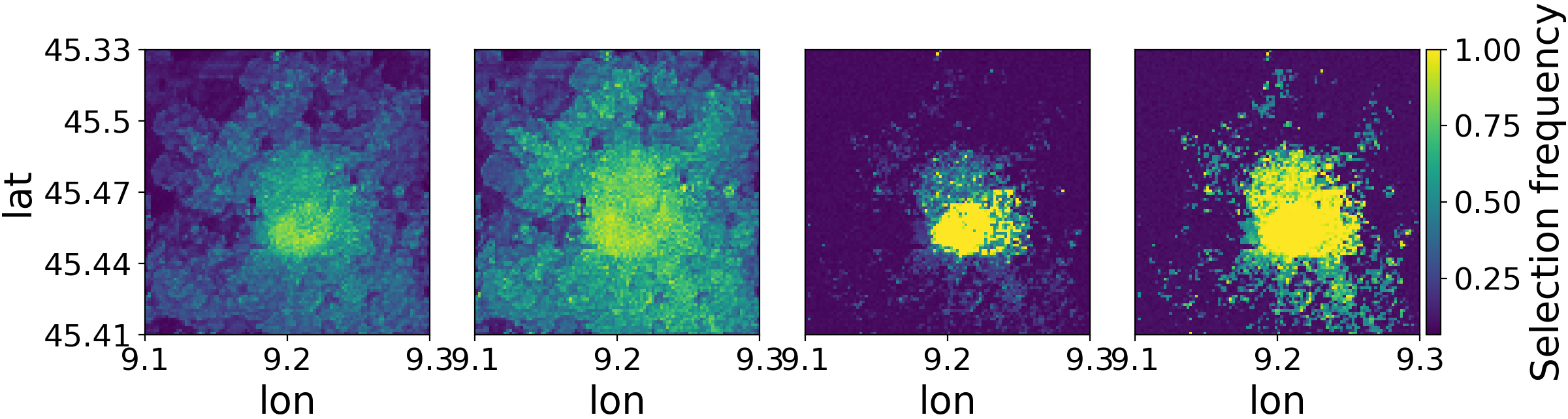}
\caption{Cell selection frequency. From left to right: historical data-based selection in off-peak, peak; \name off-peak, peak.}
	\label{peak}
\vspace*{-1em}
\end{figure}

\begin{figure}[]
	\centering
\includegraphics[width=0.65\textwidth]{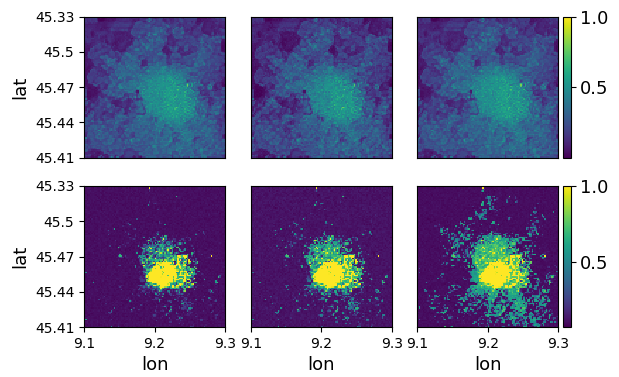}
\caption{Cell selection frequency. From top to bottom: historical data-based selection, Spider; From left to right: holiday, weekend, weekdays.}
	\label{week}
\end{figure}

The superior performance of \name is further emphasized in Figure \ref{res_holi} where our framework is applied for measurement collection during the Christmas holidays, when traffic demand decreases below typical daily averages. Spider is able to adapt to previously unseen traffic patterns and distinguishes between holidays taking place on weekdays, and normal weekdays. Precisely, it selects 19.8\% fewer cells on this occasion. In contrast, random and historical data-based selection approaches fail to adapt to such circumstances.

We delve deeper into which cells are selected by Spider, showing in Figure \ref{peak} (right) their selection frequency at peak (7AM--7PM) and off-peak (7PM--7AM) times during weekdays, juxtaposed with the behaviour of the historical data-based approach (left). Observe that \name focuses more on the city centre where the traffic demand is relatively larger, expanding sampling coverage at peak time. Figure \ref{week} shows the selection frequency at holiday, weekend and weekdays. Consistently, Spider increases the sampling coverage on weekdays and samples less in the weekend. It changes the selection pattern and samples more on holidays given the unusual smaller traffic demand, even though the holiday covers both weekend and weekdays.

Finally, we note that directly predicting a selection matrix by \name's policy network takes 7 milliseconds on average, whereas predicting the next best action by the RL agent would have taken 2 seconds. Thus, \textbf{our approach is 284 times faster and suitable for operational settings.}

\subsection{Results: traffic reconstruction}

Next we examine the performance of our MTRNet in terms of reconstruction quality, versus that of KNN, compressive sensing (CS) \cite{Jiang:2020} and spatio-temporal compressive sensing (STCS) \cite{roughan2011spatio} alternatives. With KNN, traffic at unsampled locations is approximated by the $K$ nearest sensed values with a weight of their distance. CS exploits and iteratively solves $\ell_1$ minimization problems to reconstruct a sparse snapshot. STCS incorporates global spatio-temporal  properties into CS.

\begin{figure}[]
	\centering
\includegraphics[width=0.7\textwidth]{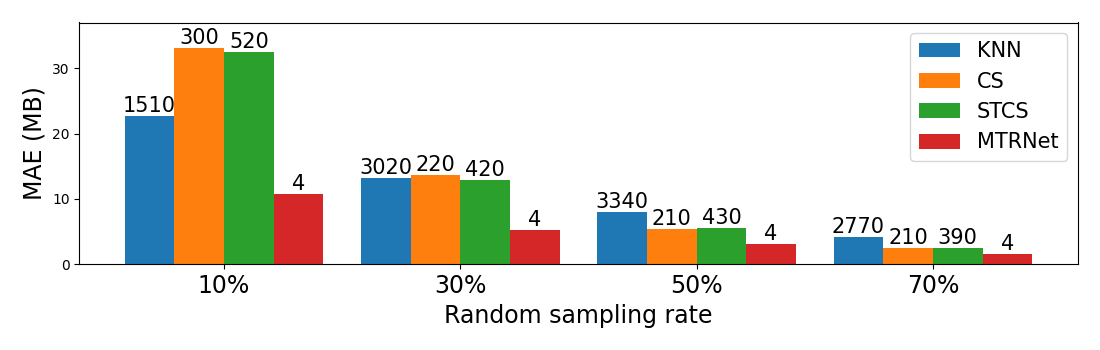}
\caption{Comparison between MTRNet and alternatives for traffic reconstruction in terms of MAE wrt. ground truth. Inference time in millisecond (ms) is shown above the bars.}
	\label{inference_comp}
\end{figure}

\begin{figure}[]
	\centering
\includegraphics[width=0.8\textwidth]{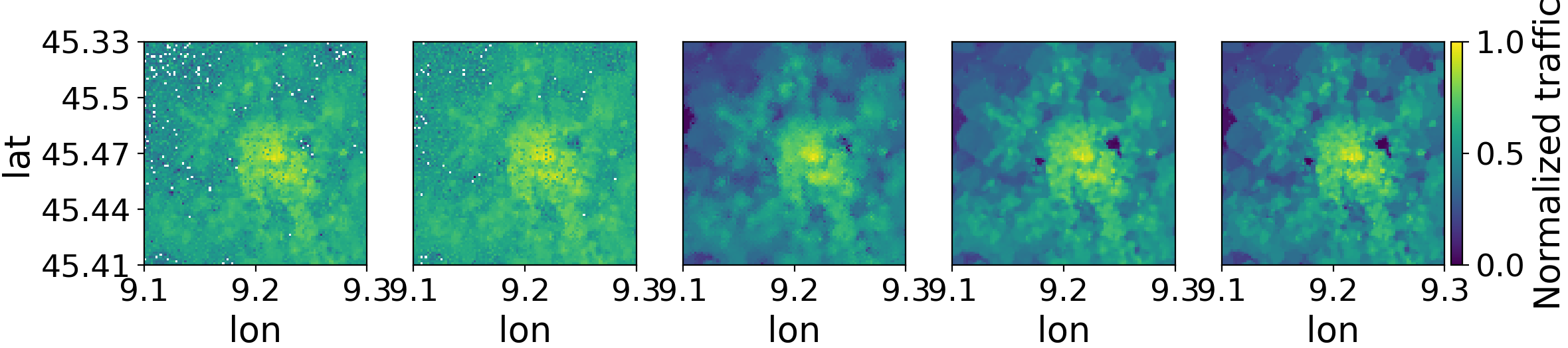}
\caption{Example normalized traffic volume of snapshot predicted by CS, STCS, KNN and MTRNet, and respectively the ground truth (from left to right)}
	\label{predicted}
\vspace*{-1em}
\end{figure}

Figure \ref{inference_comp} shows the average MAE attained by each approach 
when the sampling rate is varied between 10--70\%. MTRNet clearly outperforms the benchmarks considered, achieving the smallest MAE. The performance gap is wider when the random sampling rate is small. Specifically, \textbf{MTRNet reduces the MAE attained with KNN, CS and STCS by 52.4\%, 67.43\% and respectively 66.8\%} when the sampling rate is 10\%. Further,  \textbf{MTRNet reduces the inference time per snapshot by up to 835$\times$}.

Examining the predictions made by all algorithms when traffic is sampled randomly with 35\% rate, Figure~\ref{predicted} demonstrates that MTRNet achieves remarkable fidelity in reconstructing traffic snapshots. In contrast, the baselines deviate considerably from the ground truth, CS and STCS over-estimating the traffic volume, while KNN being unable to accurately reconstruct traffic in areas with lower consumption. 

\section{Discussion}

\subsection{Choice of threshold} \label{choice}
 The reconstruction quality threshold in the DRL process (explained in Sec. \ref{sec:agent}) is a predefined hyperparameter. It is a trade-off between the number of selected cells and mean MAE. Larger threshold results in less selected cells and larger mean MAE, and vice verse. It affects the performance of the whole framework, therefore choosing a good value where we could obtain a relatively small error and a relatively small number of cells is critical. 

\begin{figure}[]
	\centering
\includegraphics[width=0.5\textwidth]{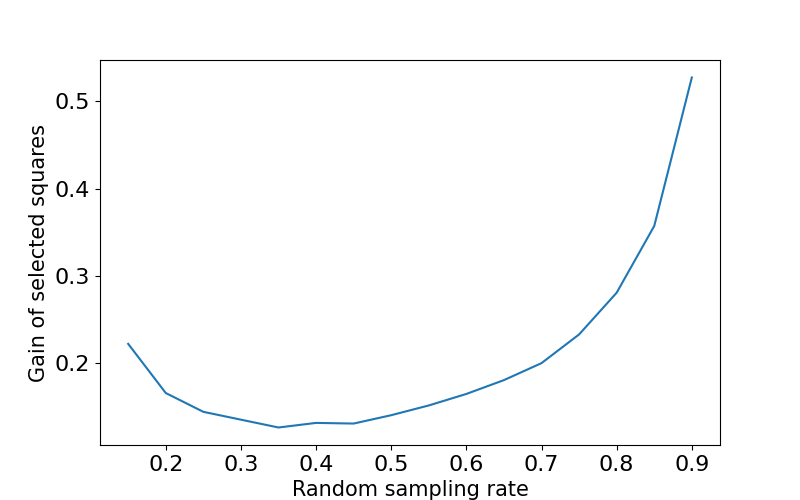}
\caption{Gain of selected squares with different random sampling rates}
	\label{gain_thre}
\vspace*{-1em}
\end{figure}

Figure \ref{gain_thre} (a) shows gain of selected cell with different random sampling rates. For instance, the point $(0.15, 0.18)$ means 18\% less cells are selected when the sampling rate changes from 10\% to 15\%. The gain is decreasing when increasing the sampling rate until it is 35\%,  which means at this point we do not gain much if we continue increasing the sampling rate. Therefore, the mean MAE value when the random sampling rate is 35\% could be a good choice for the threshold. On the other hand, if the time and collection budget is enough, the mobile operator could choose a smaller threshold to obtain a smaller inference error.

\subsection{Complexity comparison of MTRNet}
\begin{figure}[]
	\centering
\includegraphics[width=0.63\textwidth]{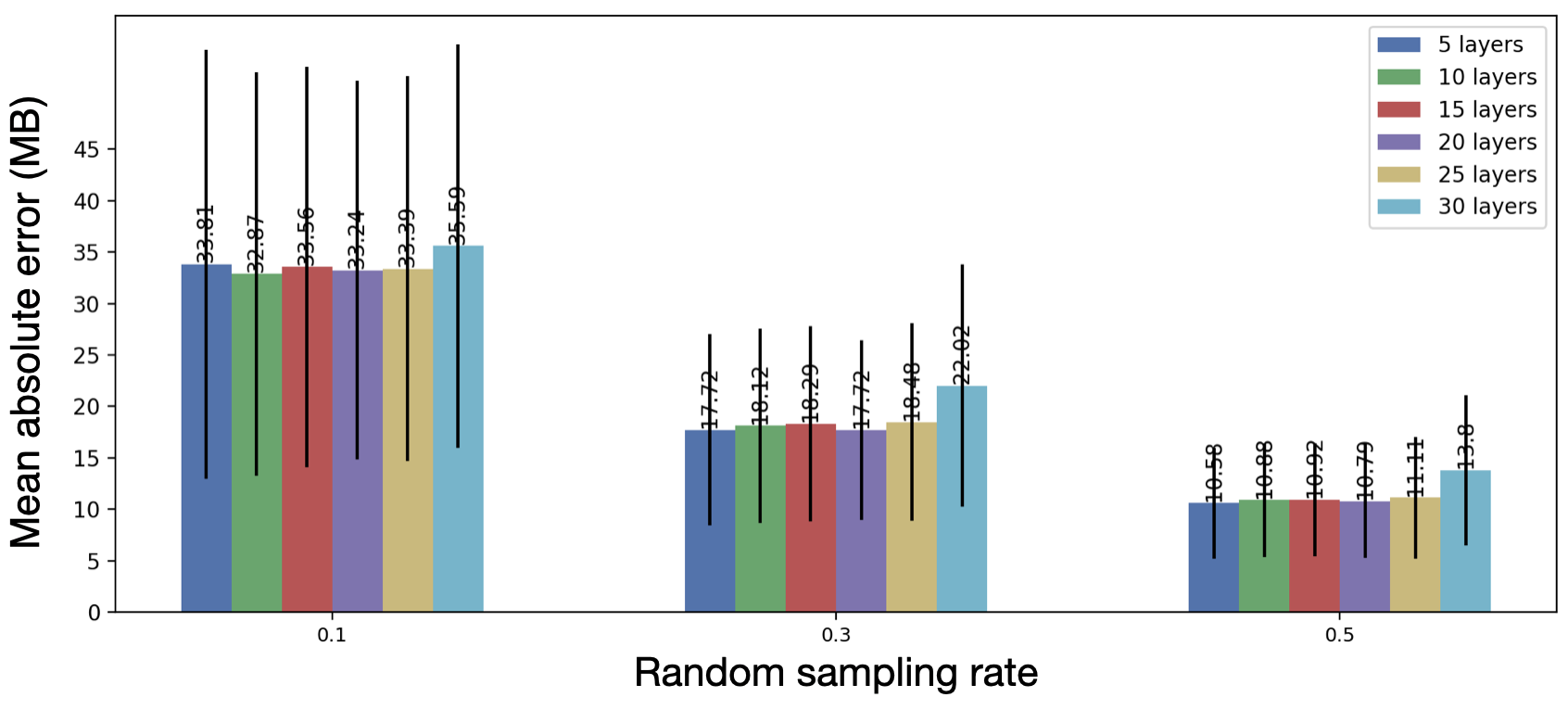}
\caption{Performance complexity of MTRNet with different number of 2D Convolutional layers
}
	\label{mtr_per}
\vspace*{-1em}
\end{figure}

\begin{figure}[]
	\centering
\includegraphics[width=0.6\textwidth]{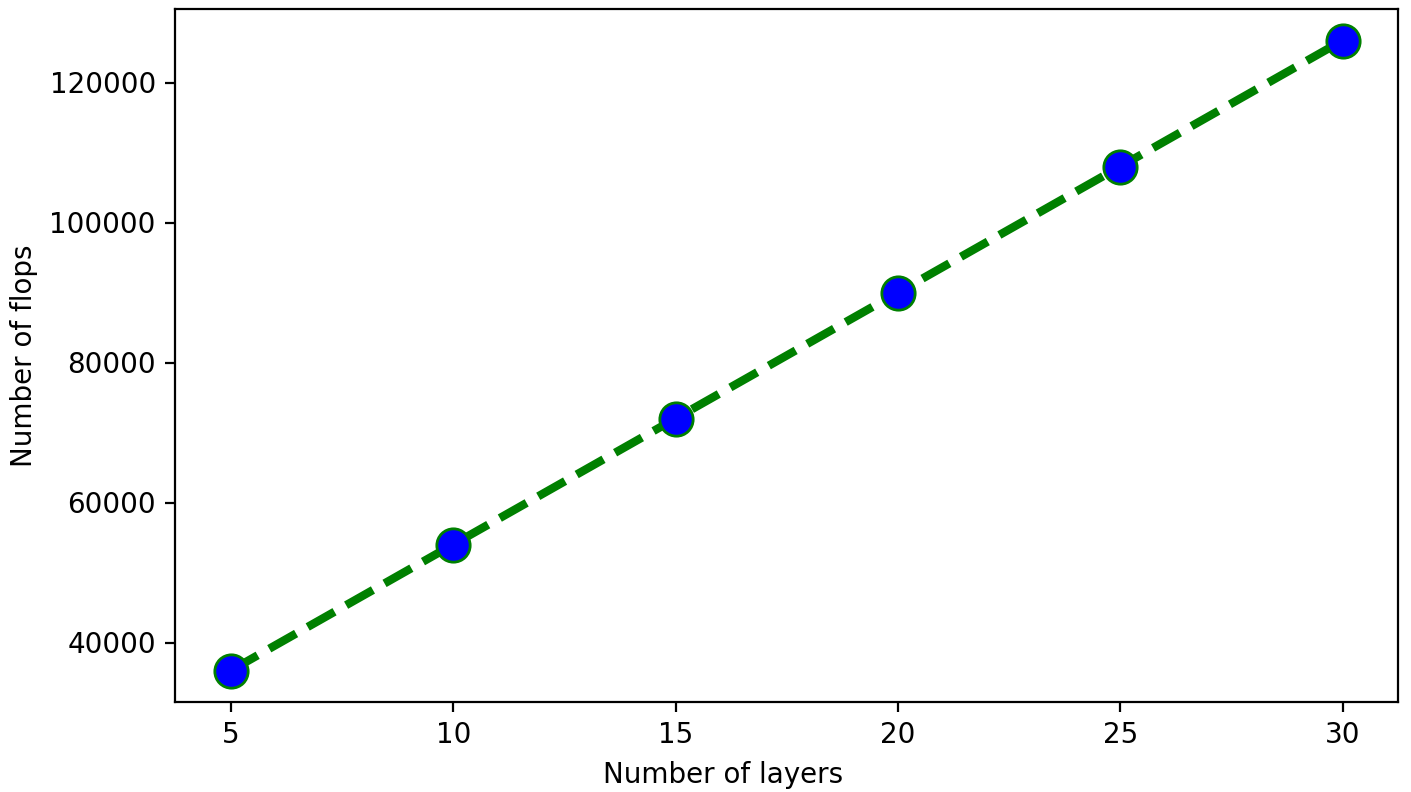}
\caption{Time complexity of MTRNet with different number of 2D Convolutional layers}
	\label{mtr_time}
\vspace*{-1em}
\end{figure}

Recall that in the Feature Extraction block of MTRNet, there are several 2D Convolutional layers. Choosing a good value of the number of these layers is critical to maximizing the model's performance with low time complexity. We train models with different number of layers and compare their performances and their time complexities given random sampled testing data. For performance comparison, we calculate the average MAE and the standard derivation between ground truth and inferred snapshots. As shown in Figure \ref{mtr_per}, we can see that the error does not differ much from 5 layers to 25 layers, and it only becomes worse when increasing it to 30. For a  time complexity comparison, we measure the average number of flops for one inference. In Figure \ref{mtr_time}, the number of flops increases linearly with respect to the number of layers, and the model with 5 layers achieves the smallest number of flops. Based on the results in these two plots, we choose the number of layers to be 5.

\subsection{Collecting and processing data in the real world}
There are practical factors in the real world that may affect the mobile traffic inference accuracy. Firstly, the mobile operator needs to partition the target area into grids, which highly depends on the distribution of the probes and the terrain. If the probes are sparsely distributed, it would be difficult to determine their precise coverage area. Secondly, our approach is based on the assumption that data from every cell could be available as needed. However, in reality, sometimes the probes might stop working, they are not activated due as they can slow down traffic, etc. and the data may not available. In that case, the agent is not able to select the corresponding action. If some locations are not available for collection, then measurement collection cost is smaller but the prediction error grows. We conducted an experiment where 10\% random locations are not available for collection at every timestamp. In that configuration, Spider selects 1965.44 cells with NAME of 0.1091, which is 6.82\% fewer cells and 4.2\% larger error.

Additionally, latency is critical in real-time system. There are hardware constraints on computing capabilities, especially in the data collection stage. Data collection includes capture, preprocessing from selected probes, transferring all the data to a central site, and storage for the current and previous readings. Therefore, it requires high computing power and large storage capabilities in order to run in real-time.

\section{Summary}
In this project, we introduced \name, a mobile traffic measurement collection and reconstruction framework that harnesses deep (reinforcement) learning to reduce measurement collection overhead and infer mobile traffic consumption at city-scale with high accuracy. We overcame the challenges of handling large action spaces in the cell sampling process and designed a purpose-built neural model that exploits spatiotemporal patterns characteristic to mobile traffic, to reconstruct network-wide traffic snapshots. We experimented with a real-world dataset and demonstrated \name reduces cell sampling overhead by up to 48\% while significantly outperforming existing measurement interpolation methods.

Here, we modelled the cellular network via Euclidean representations (e.g. matrices). One of the drawbacks which is discussed in Sec. 3.4.3 is that our approach assumes the probes are evenly distributed across the target area, which is usually not the case for urban deployments. In the next Chapter, we will discuss the disadvantages of Euclidean representations and propose a non-Euclidean graph-based structure for long-term mobile traffic forecasting.

\chapter[A Handover-aware Spatiotemporal Graph Neural Network ... ]{A Handover-aware Spatiotemporal Graph Neural Network for Mobile Traffic Forecasting}
Real-time mobile traffic prediction is becoming critical to mobile network resource management. For example, based on the prediction result, mobile operators can disable 
specific probes for measurement collection to save energy, thus promoting greener cellular networks. On the other hand, if the result suggests there will be a surge in some area, the mobile operator can assign more resources to specific base stations and plan in advance.

Previous forecasting solutions build on Euclidean distance-based representations of RAN layouts, which are mapped onto grids, before employing Long Short Term Memory (LSTM) structures \cite{zhang2019new}, Convolutional Neural Networks (CNNs) \cite{zhang2018citywide}, or a combination of these \cite{zhang2018long}. Such approaches ignore \emph{(i)}  that users' traffic consumption might be handed over to other sectors or to different technologies (e.g. from 4G to 3G), and \emph{(ii)} the fact that relationships between these stations change over time.

To be able to capture inter-sector/-base station dependencies at different timestamps and produce more accurate forecasts, in this work we use graphs to represent cellular networks, where vertices correspond to antenna sectors, and weights quantify the dependency between. As discussed in the related work, recent studies on Graph Convolutional Network (GNN) 
model such dependency as the distance between two locations  but such location proximity of base stations may not reflect strong dependency due to terrain constraints, while distant base stations may exhibit strong dependencies because of user daily commute patterns. For example, the fact that people go to work/school in the morning and leave in the evening everyday results in a regular traffic handover pattern, and this should be taken into consideration when modelling the cellular network. Therefore we harness fine-grained handover frequency information, which reflects user mobility to some extent and better captures traffic dynamics. 

In this chapter, we propose \textbf{\model (Spatiotemporal Dynamic Graph Network)}, a handover-aware spatiotemporal graph neural network for mobile traffic forecasting. We model the cellular network as a directed and weighted dynamic graph, which captures spatiotemporal correlations from both traffic consumption and handover frequency. In particular, vertices keep information about the volume of traffic at a particular sector and time, while the weights are handover frequencies between vertices over fixed observation windows. We leverage Gated Linear Units (GLU) to extract temporal features, and unlike RNN-based models, output traffic predictions at all the base stations in a deployment simultaneously, while the structure is faster to train. We propose a novel approach to dynamic graph convolution, combining dynamic graph spectral convolution and diffusion graph convolution, thereby capturing spatial correlations both short- and long-term. We apply a real-world mobile traffic dataset provided by a major mobile operator in Turkey. This dataset is different from the dataset in Chapter 3, as it has the critical handover information to be captured by the graph weights. We validate SDGNet with this mobile traffic dataset, demonstrating that our model achieves traffic forecasting with up to 75.2\%, 59.4\% and 56.0\% higher short-, mid- and long-term accuracy, compared with a range of benchmarks.


\section{Problem Formulation}
\label{problem_formulation}
Consider a mobile network deployment comprising $N$ base stations. We denote by  $\mathbb{F}_t=\{F_{t-T+1}, ..., F_t\} \in \mathbb{R}^{T \times N}$ a sequence of mobile traffic consumption measurements over $T$ timestamps up to the current time $t$, where $F_t$ is the traffic snapshot across all base stations, observed over an interval $[t-\Delta, t]$, i.e., $F_t = \{f^t_1, ..., f^t_n \}$, where $f^t_i$ is the volume of traffic at the $i$-th base station, and $\Delta$ is the temporal granularity of traffic observations configurable by a network administrator.

The mobile traffic network is represented as a directed, weighted and dynamic graph. At the $t$-th time step, we define a graph $G_t = (v_t, A_t)$, where $v_t \in \mathbb{R}^{N \times C}$ is the graph signal and $C$ is the number of features (i.e., traffic consumption and time information), and $A_t \in \mathbb{R}^{N \times N}$ is the adjacency matrix where an element $a^t_{i, j}$ represents the handover frequency between base stations $i$ and $j$ observed at that time. 

Our objective is to predict the most likely mobile traffic consumption in the next $H$ time steps, given the past $T$ observations, i.e.,
\begin{align*}
& \hat{F}_{t+1}, .., \hat{F}_{t+H}   = \\
& \mathop{\arg \max }\limits_{v_{t+1}, \ldots, v_{t+H}} \log P \left(F_{t+1}, ..., F_{t+H} \mid G_{t-T+1}, \ldots, G_{t}\right).
\end{align*}

\begin{figure}[t]
	\centering
\includegraphics[width=0.65\textwidth]{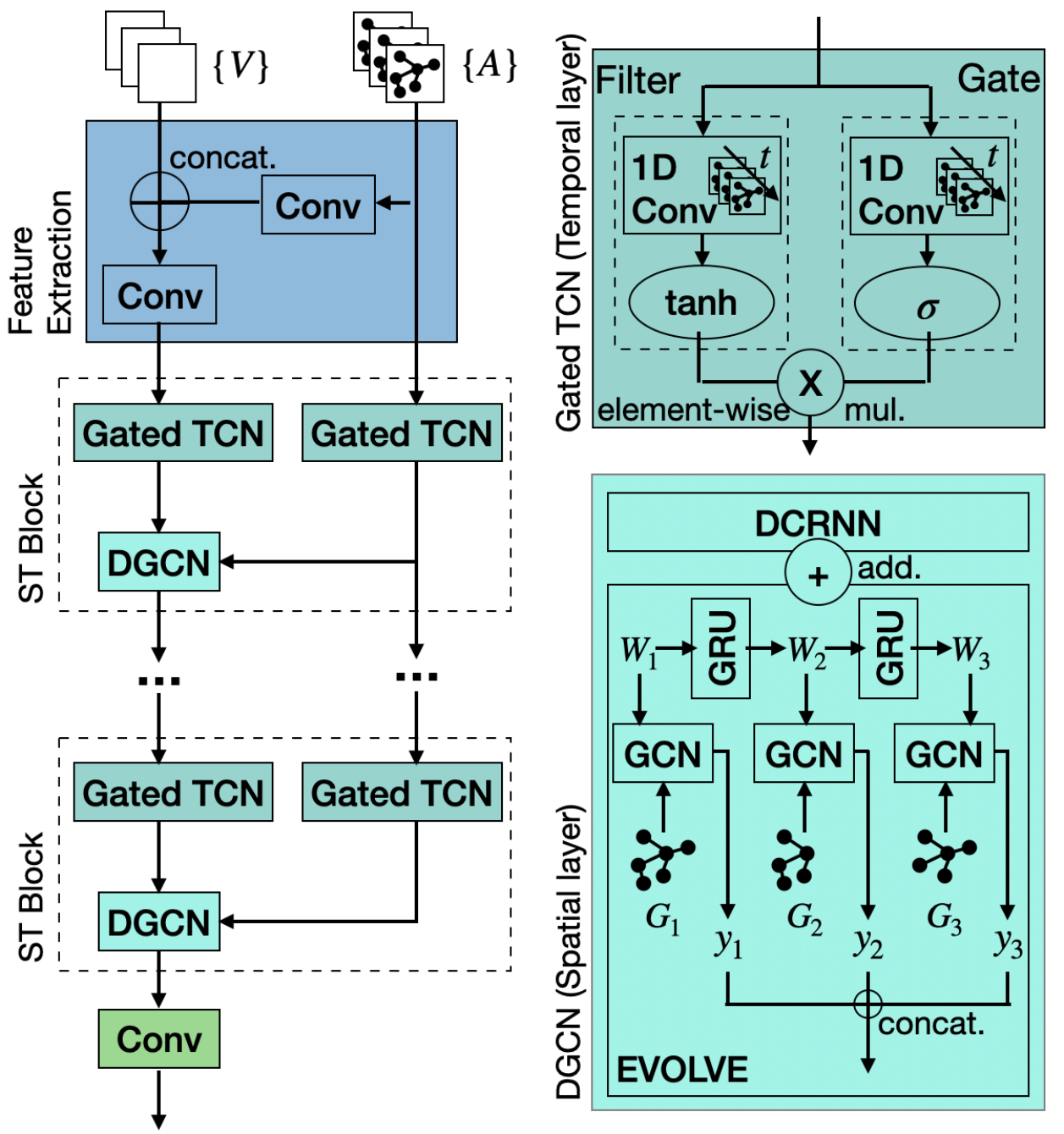}
\caption{Proposed SDGNet structure consisting of a feature extraction block, several ST blocks, and a readout 2D Conv layer (left). ST blocks consist of 2 temporal and 1 spatial layer (the structure of each shown on the right).}
	\label{structure}
\end{figure}

\section{SDGNet}
Inspired by Graph WaveNet \cite{wu2019graph}, we propose SDGNet, a deep neural network that solves the mobile traffic forecasting problem posed in Sec. \ref{problem_formulation}. SDGNet captures spatiotemporal correlations among traffic consumption at different locations and dynamic adjacency matrices modelled from handover data. The proposed model takes as input $T$ past graphs (i.e., $G_{t-T+1}, ..., G_t$); therefore, the input consists of two parts: graph signals $\mathbf{V} \in \mathbb{R}^{T \times N \times C}$ and dynamic adjacency matrix $\mathbf{A} \in \mathbb{R}^{T \times N \times N}$. The input is fed into a feature extraction block followed by several spatiotemporal (ST) blocks, as shown in Figure \ref{structure}. Each ST block comprises two temporal layers that handle graph signals and adjacency matrices, and a spatial layer for dynamic graph convolution. In what follows, we explain in detail the inner workings of these different modules. 

\paragraph*{Spatiotemporal feature extraction} 
The first block generates feature maps for the next module by capturing spatiotemporal correlations from graph signals $\mathbf{V} \in \mathbb{R}^{T \times N \times C}$ and dynamic adjacency matrix $\mathbf{A} \in \mathbb{R}^{T \times N \times N}$. The feature dimension of $\mathbf{A}$ is first reduced by a Conv layer before concatenating with $\mathbf{V}$, so that features pertaining to $\mathbf{A}$ do not dominate in the concatenated matrix. We then pass the results through another Conv layer to extract feature maps that will be handled by the subsequent spatiotemporal block. 

\begin{figure}[]
	\centering
\includegraphics[width=0.49\textwidth]{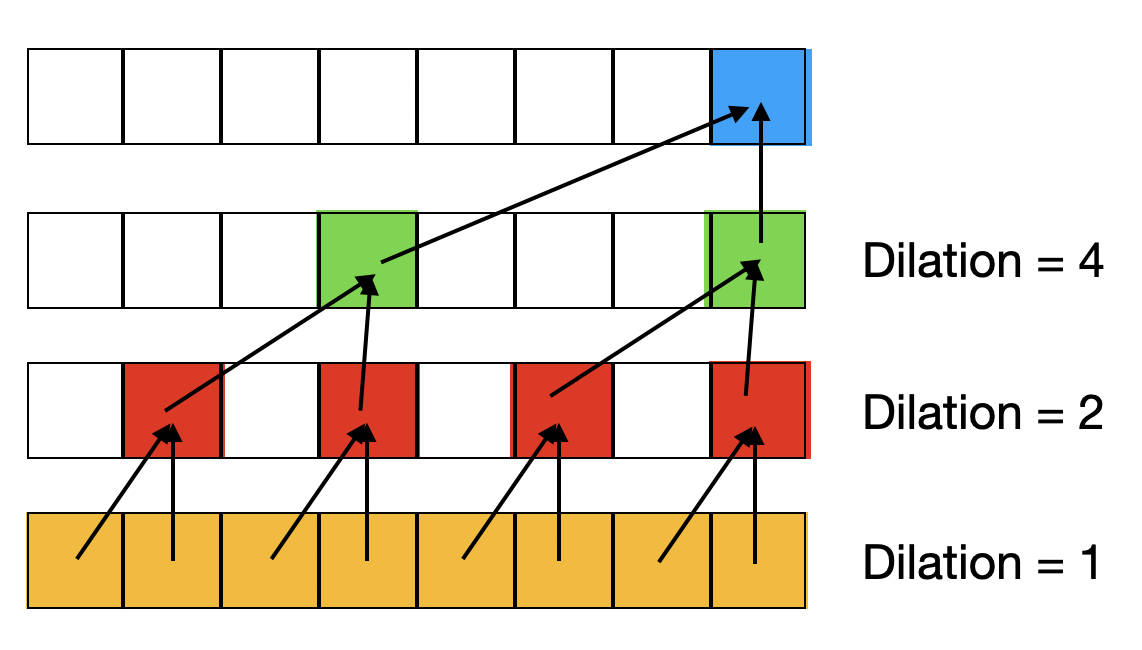}
\caption{Illustration of dilated causal convolution}
	\label{tcn}
\end{figure}

\begin{figure}[]
	\centering
\includegraphics[width=0.2\textwidth]{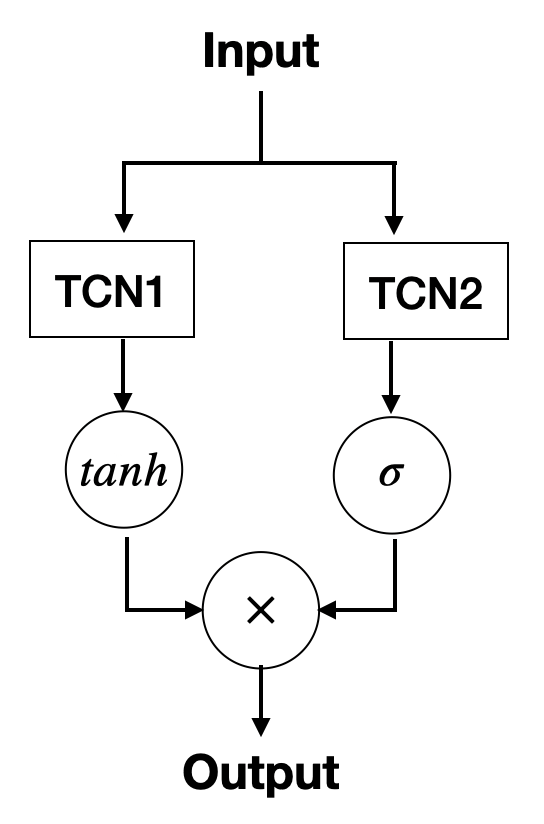}
\caption{Illustration of gating mechanism}
	\label{gated}
\end{figure}

\paragraph*{Gated TCN} Each ST block encompasses two gated temporal convolution networks (TCNs) and a Dynamic Graph Convolution Network (DGCN). We adopt gated 1-D dilated causal convolution \cite{wu2019graph} as the temporal convolution layer to capture complex temporal dependencies.  Dilated causal convolution can handle long-term sequences in a non-recursive manner and converges faster compared to traditional RNN-based models. The structure works by sliding over inputs and skipping elements with a given step, which increases with the layer depth. Given an input \(\mathbf{x} \in \mathbf{R}^{T}\) and filter \(\mathbf{g} \in \mathbf{R}^{K}\),
the dilated causal convolution operation $\star$ is represented as:
$$
\mathbf{x} \star \mathbf{g}(t)=\sum_{s=0}^{K-1} \mathbf{g}(s) \mathbf{x}(t-d \times s) ,
$$
where $d$ is the dilation factor determining the length of the skipping step. Several dilated causal convolution layers are stacked with increasing receptive fields. Shown in Figure \ref{tcn}, as the dilation step increases, the features are further extracted and summarized over the temporal dimension. 

A gating mechanism (shown in Figure \ref{gated}) is applied to control the information flow through layers \cite{dauphin2017language}, leading to the following output:
$$
\mathbb{H}(\mathbf{x}) = z\left(\mathbf{x} \star \mathbf{g}_1(t)+\mathbf{b}\right) \odot \sigma\left(\mathbf{x} \star \mathbf{g}_2(t)+\mathbf{c}\right),
$$
where $\mathbf{b}$ and $\mathbf{c}$ are model weights, $\odot$ is the element-wise multiplication, $z(\cdot)$ is an activation function, and $\sigma(\cdot)$ is the sigmoid function which controls the  information passed to the next layer.
We apply Gated TCNs on both inputs $\mathbf{V}$ and $\mathbf{A}$, to learn their temporal dependencies while reducing the temporal dimension of the propagated output. 

\paragraph*{Dynamic Graph Convolution Network} 
To obtain accurate forecasts both short- and long-term, we combine spectral graph convolution and DCRNN into the dynamic graph convolution network.
Intuitively, we can adopt spectral graph convolution by applying the multiplication of $\mathbf{x} \in \mathbb{R}^{T \times N \times C}$ and the adjacency matrix $\mathbf{A} \in \mathbb{R}^{T \times N \times N}$ along the first dimension. This acts as $T$ copies of spectral graph convolution, each one working on one snapshot of the sequence. One drawback is that these $T$ copies share one weight, and have no temporal correlations. To circumvent this issue, we adopt EvolveGCN \cite{pareja2020evolvegcn}, where we assign a weight to each snapshot, and these weights are temporally related. Mathematically, for every snapshot $\mathbf{x}_t$ and its corresponding adjacency matrix $\mathbf{A}_t$,
\begin{align*}
 \mathbf{x}_{t} &=\mathbf{x_t} * \mathcal{G} \mathbf{w}_{t}  =\sigma\left(\tilde{\mathbf{A}}_{t} \mathbf{x}_{t} \mathbf{w}_{t}\right),\\ 
\mathbf{w}_{t}&=\operatorname{GRU}(\mathbf{w}_{t-1}),
\end{align*}
where \(\tilde{\mathbf{A_t}}=\mathbf{I}_{\mathbf{n}}+\mathbf{D}^{-\mathbf{1} / \mathbf{2}} \mathbf{A_t} \mathbf{D}^{-\mathbf{1} / \mathbf{2}}\) and $\mathbf{w_t}$ is the weight of $t$-th snapshot. Each GCN operation has a weight, which is generated from the weight in the last snapshot using a Gated recurrent unit (GRU). The GRU is used to capture long-term temporal correlations. Compared to the Long Short-Term Memory (LSTM), it has fewer gates and therefore is faster to train and uses less memory. We initialize $\mathbf{w}_1$ at the beginning. For the following recurrent steps, we use the last output as both hidden state and the input to the GRU. Finally, we concatenate the output from every snapshot as the final output of EvolveGCN. We denote the EvolveGCN operator as  $* \mathcal{E} $.


We express the DGCN operation in matrix form:
$$
\begin{aligned}
\mathbb{Y}(\mathbf{x}, \mathbf{A}) = & \mathbf{x} * \mathcal{E} \mathbf{w^e} + \mathbf{x} * \mathcal{D} \mathbf{w^d}  \\
= &
\|_{t=1}^{T} \mathbf{\Tilde{A}_t} \mathbf{x} \mathbf{w_t^e} + \sum_{k=0}^{K} ( \mathbf{P}_{f}^{k} \mathbf{x} \mathbf{w^d_{k, 1}}+ \mathbf{P}_{b}^{k} \mathbf{x} \mathbf{w^d_{k, 2}}),
\end{aligned}
$$
where $\|$ denotes concatenation, \(\tilde{\mathbf{A}}=\mathbf{I}_{\mathbf{n}}+\mathbf{D}^{-\mathbf{1} / \mathbf{2}} \mathbf{A} \mathbf{D}^{-\mathbf{1} / \mathbf{2}}\), 
$\mathbf{w^e} \in \mathbb{R}^{T \times C \times C^{\prime}}$, $\mathbf{w^d} \in \mathbb{R}^{C \times C^{\prime}}$. $C^{\prime}$ is the dimension of the hidden states. 

Finally, the formula of the $l$-th ST block given the input graph signal $v^l \in \mathbb{R}^{T \times N \times C^{\prime}}$  and input adjacency matrix $A^l \in \mathbb{R}^{T \times N \times N}$, is given by:
$$
v^{l+1} = \mathbb{Y}(\mathbb{H}(v^{l}), \mathbb{H}(A^{l})); \quad
A^{l+1} = \mathbb{H}(A^{l}).
$$


\paragraph*{Training Loss}
We seek to minimize the Mean Square Error (MSE) when training the overall neural model, i.e.
$$
\mathbb{L}(\hat{F}_{t+1}, .., \hat{F}_{t+H}) = \frac{1}{H\times N}\sum_{i=1}^{H}\sum_{n=1}^{N} (\hat{f}^i_n-f^i_n)^2,
$$
where $H$ is the number of prediction steps and $N$ is the number of base stations in the deployment.

\section{Performance Evaluation}

\subsection{Dataset}
\begin{figure}[]
	\centering
\includegraphics[width=0.65\textwidth]{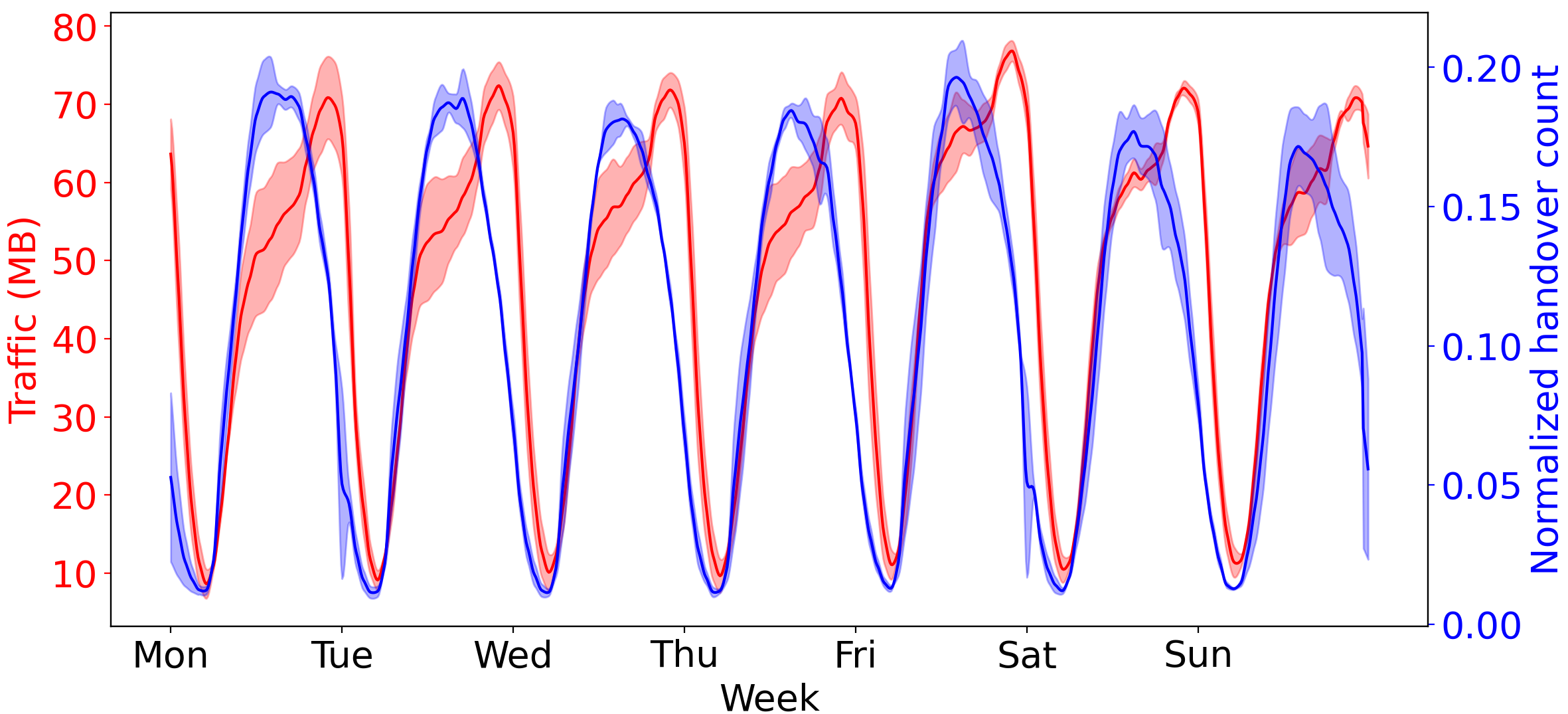}
\caption{Average and std deviation of the weekly traffic consumption and handover frequency, over the entire deployment. 
}
	\label{th}
\end{figure}

We evaluate the proposed model using real-world mobile traffic data collected by Turkcell between 5--29 March 2020 in a medium-size city. Figure \ref{th} shows the weekly average and standard deviation of the volume of traffic and handover frequency across the entire deployment. To ensure sufficient data samples for training, we augment the dataset using bicubic interpolation, enhancing the  granularity from 1 hour to 10 minutes. We split the data into training, validation and testing using the 6:2:2 ratio. We confine consideration to 100 base stations that exhibit the overall highest traffic consumption. 

To speed up training, we normalize the traffic consumption data by $\log(1 + x) / \Bar{x}$, where $x$ is a traffic consumption measurement and $\Bar{x}$ is the mean of $\log(1 + x)$ across all data points. We further normalize the adjacency matrices by a Softmax function. We add absolute time information as additional features of the graph signals, namely day of the week, hour of the day, and minute of the hour, which we map to the $[-1, 1]$ range. We use historical two-hour windows as inputs to model input, i.e. 12 snapshots.

\subsection{Implementation}
We implement \model using the PyTorch library and employing 8 ST blocks. We train using the Adam optimizer with learning rate \mbox{$\lambda = 10^{-4}$} and a batch size of 128.

\subsection{Results}

\begin{table}[]
\centering
\caption{Model configurations}
\begin{tabular}{lll}
\hline
\multicolumn{1}{c}{\multirow{2}{*}{Model}}                       & \multicolumn{2}{c}{Configuration}                                                                                                                                                  \\ \cline{2-3} 
\multicolumn{1}{c}{} & {Spatial operation}                                                          & Temporal  operation \\ \hline
STGCN-HO                                   & GCN                                                                                            & 1D CNN                                                                            \\ \hline
DCRNN                                      & DCRNN                                                                                          & Gated TCN                                                                         \\ \hline
WaveNet                                    & \begin{tabular}[c]{@{}l@{}}GCN with self-adaptive \\ adjacency matrix +DCRNN\end{tabular}      & Gated TCN                                                                         \\ \hline
STAWnet                                    & \begin{tabular}[c]{@{}l@{}}GCN+DCRNN, both with \\ self-adaptive adjacency matrix\end{tabular} & Gated TCN                                                                         \\ \hline
SDGNet                                     & EvolveGCN+DCRNN                                                                                & Gated TCN                                                                         \\ \hline
\end{tabular}
\label{config}
\end{table}

\begin{table}[]
\centering
\caption{Forecasting performance comparison}
\begin{tabular}{lllllll}
\hline
         & \multicolumn{2}{l}{30 min}    & \multicolumn{2}{l}{60 min}    & \multicolumn{2}{l}{120 min}   \\ \cline{2-7} 
Model    & MAE           & RMSE          & MAE           & RMSE          & MAE           & RMSE          \\ \hline
STGCN-HO & 3.14          & 4.78          & 3.23          & 4.85          & 4.29          & 6.12          \\
DCRNN    & 1.03          & 1.70          & 2.17          & 3.53          & 3.81          & 5.92          \\
WaveNet  & 0.83          & 1.38          & 1.99          & 3.16          & 3.23          & 5.03          \\
STAWnet  & 1.16          & 1.74          & 1.85          & 3.11          & 3.00          & 4.84          \\
\textbf{\model}     & \textbf{0.78} & \textbf{1.28} & \textbf{1.31} & \textbf{2.42} & \textbf{2.75} & \textbf{4.49} \\ \hline
\end{tabular}
\label{overall}
\end{table}

We measure the prediction accuracy of our SDGNet over 3 steps (30 min), 6 steps (60 min), and respectively 12 step (120 min), in terms of mean absolute area (MAE) and root mean squared error (RMSE), which we compare against that of the following graph neural network benchmarks (their configuration is shown in Table \ref{config}):
\begin{itemize}
    \item STGCN-HO \cite{zhao2020cellular} -- also models dependencies through handover frequency, but uses the average handover frequency computed over four random days, without capturing any handover dynamics.
    \item DCRNN \cite{li} -- as introduced before, DCRNN models the traffic flow as a diffusion process on a directed graph.
    \item WaveNet \cite{wu2019graph} -- introduces an adaptive adjacency matrix, which is initialized with the adjacency matrix and learned in the training process.
    \item STAWnet \cite{tian2021spatial} -- introduces self-learned node embedding by a dynamic attention mechanism; no prior knowledge of the graph is needed; the adjacency matrix is not used in attention learning.
\end{itemize}

The results are summarized in Table \ref{overall}. Observe that our solution outperforms the existing models, reducing the MAE of the best performing benchmark by 6.0\% (DCRNN), 7.2\% (WaveNet) and 8.3\% (STAWnet), when making 3-step, 6-step, and 12-step predictions respectively. Short-term, our SDGNet provides 4$\times$ smaller errors than STGCN-HO (which uses fixed adjacency matrices), while over 2 hours our model offers up to 36\% lower MAE. 

\begin{figure*}[t]
	\centering
\includegraphics[width=1.1\textwidth]{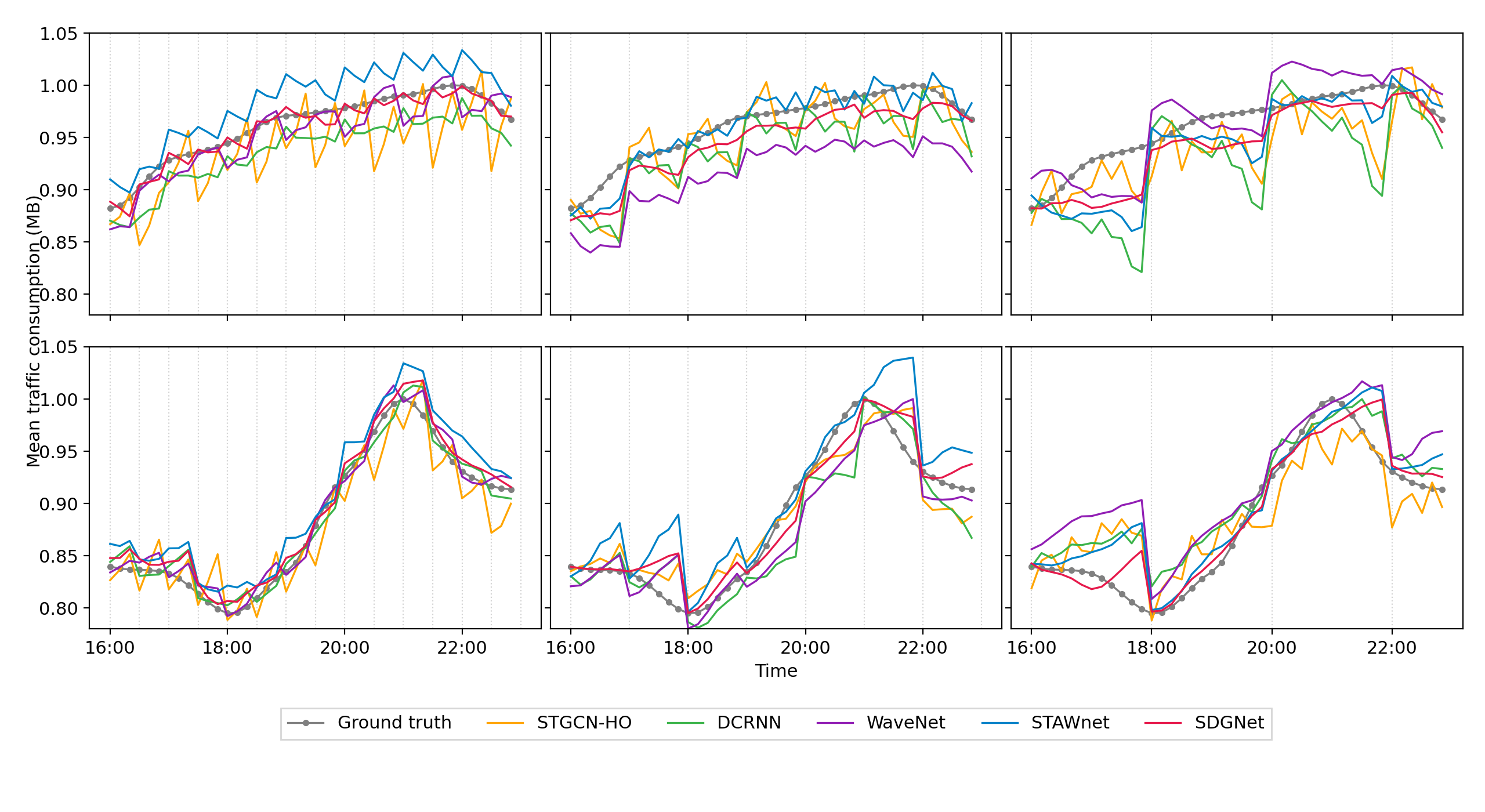}
\caption{Normalized traffic predictions between 16:00 and 23:00, averaged over 5 days, when forecasting window is 3 steps (30 min, left), 6 steps (60 min, middle), and 12 steps (120 right), at a heavily-loaded (top) and lightly-loaded (bottom) base~station.}
	\label{steps}
\end{figure*}

We delve deeper into the behaviour of SDGNet and that of the benchmarks considered, and plot the predicted traffic consumption using these models between times of the day with low demand (16:00 hrs) and high demand (23:00 hrs), zooming in on a heavily-loaded and lightly-loaded base station and examining different forecasting windows. The results are illustrated in Figure \ref{steps}, where dashed line stands indicate the beginning of a prediction window as we apply a sliding window with different numbers of predicted steps -- 3 (left), 6 (middle), 12 (right). 
Observe that DCRNN always underestimates the traffic volume at the  heavily-loaded base station, while WaveNet tends to overestimate the traffic consumption at both base stations. STAWnet performs worse when making 6-step predictions at the heavily-loaded base station. STGCN-HO fails to capture the traffic patterns correctly, which results in significant deviations from the ground truth. In contrast, our \model forecasts the traffic consumption accurately irrespective of the load regime or forecasting window length, following closely the ground truth.

\begin{figure}[t]
	\centering
\includegraphics[width=0.62\textwidth]{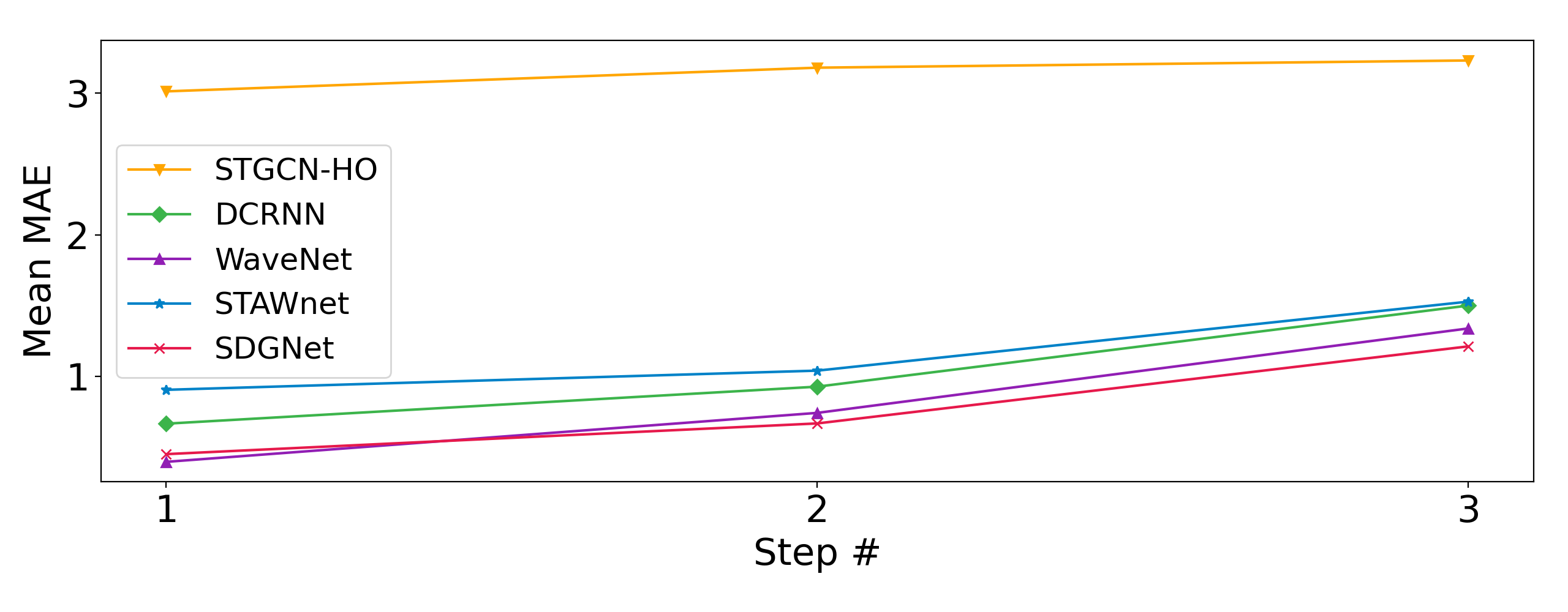}
\caption{Average MAE of each step in 3-step prediction}
	\label{3step_error}
\end{figure}

\begin{figure}[]
	\centering
\includegraphics[width=0.62\textwidth]{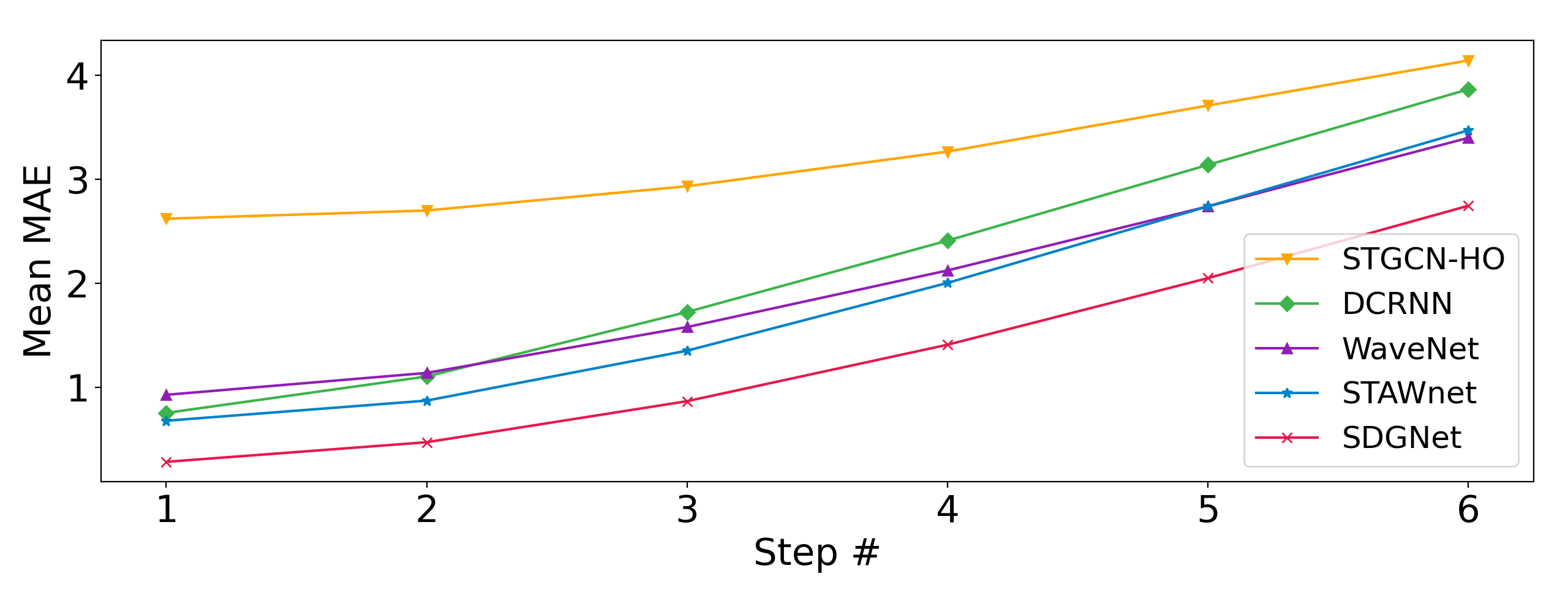}
\caption{Average MAE of each step in 6-step prediction}
	\label{6step_error}
\end{figure}

\textbf{Short- and mid-term performance:}  We take a closer look at how the different approaches compare when making short- and mid-term predictions. To this end, we plot the average MAE at each step of 3-step (30min) and 6-step 
(60min) forecasting windows, shown in Figure \ref{3step_error} and Figure \ref{6step_error}, respectively. STGCN-HO performs poorly in both the short- and mid-term. In the short-term, SDGNet and WaveNet have similar performance, but in mid-term they have a distinct accuracy difference, which means WaveNet is only suitable for short-term prediction. Our SDGNet achieves the smallest error at every step in mid-term prediction.

\begin{figure}[]
	\centering
\includegraphics[width=0.62\textwidth]{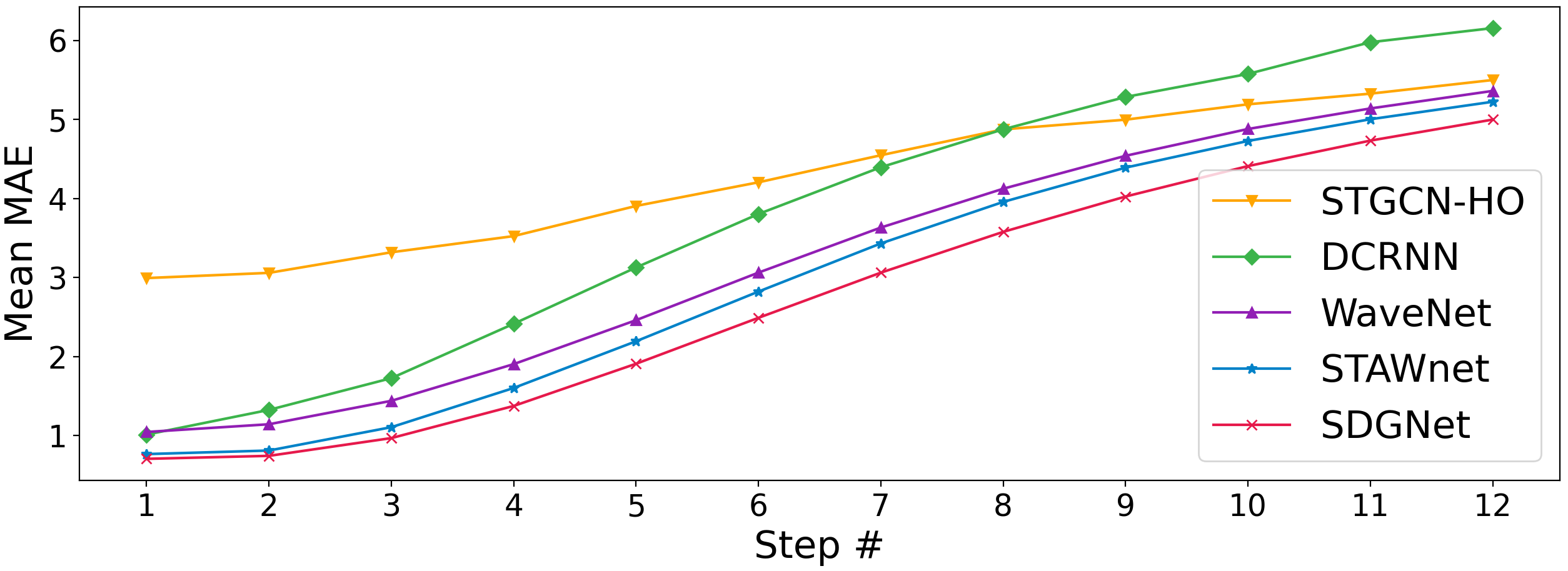}
\caption{Average MAE of each step in 12-step prediction}
	\label{12step_error}
\end{figure}

\textbf{Long-term performance:} Then we take a look at long-term predictions in Figure \ref{12step_error} where we plot the average MAE at each prediction step of 12-step (120 min) forecasting windows. STGCN-HO performs modestly at the beginning but approaches the performance of  most models in the final steps, while DCRNN predicts reasonably well at the beginning but much worse in the final steps. This indicates that spectral graph convolution is helpful for long-term prediction, while DCRNN structures benefit short-term forecasting. This is also seen in the behaviour of the other three models, where spectral convolution-based and DCRNN are combined. Our SDGNet yields the smallest error at every step, which confirms the benefit of capturing handover dynamics when making predictions.

\begin{figure}[]
	\centering
\includegraphics[width=0.73\textwidth]{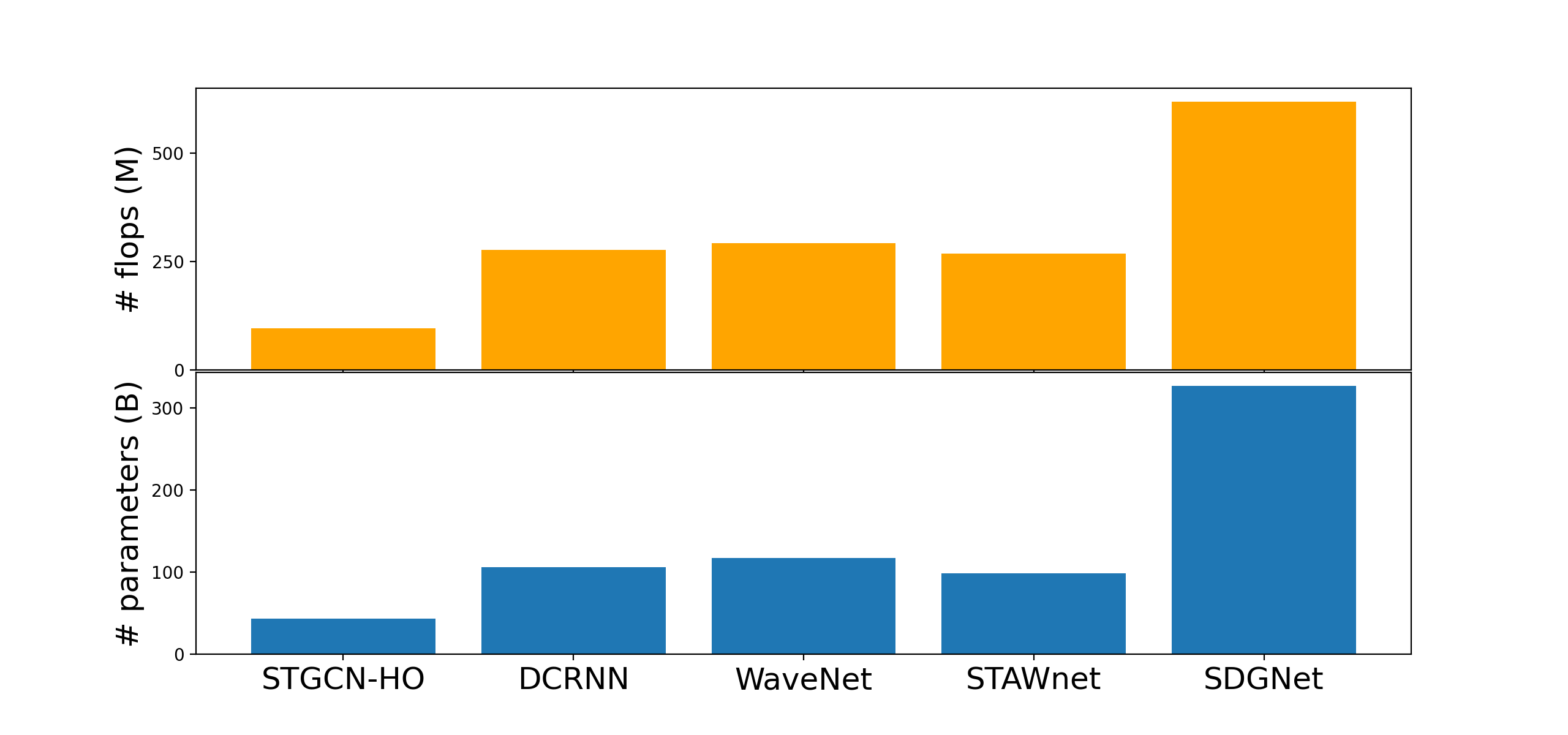}
\caption{Time complexity and model complexity of SDGNet and different benchmarks.}
	\label{flops}
		\vspace*{-1em}
\end{figure}

\textbf{Time complexity and model complexity:} Figure \ref{flops} shows the number of FLOPs and the number of parameters per inference instance of SDGNet and the benchmark models considered. Because SDGNet works with adjacency matrices of larger dimensionality and encompasses several GRUs that process the data sequentially, the accuracy gains it achieves come at the price of higher inference times and larger model size. However, given that off-the-shelf GPUs routinely handle over 20 TFLOPs per second, the cost is affordable.

\textbf{Real-world applications:}
Our neural network provides a significant MAE reduction compared to other benchmarks and brings benefits if deployed in the real-world, including intelligent management of resources, reduced operational expenditure, and smaller network carbon footprint. Traffic forecasts can enable mobile operators to switch off parts of their infrastructure or measurement equipment in order to save both energy and cost. On the other hand, anticipating traffic surges allows provisioning the right resources in advance, as well as maintaining user quality of experience by performing network capacity planning ahead time.

\section{Summary}
In this chapter, we proposed SDGNet, a handover-aware dynamic graph neural network on mobile traffic prediction. We
addressed the challenges of modelling the graph precisely of dynamic traffic patterns, and designed a model that capture the spatiotemporal correlations from both mobile traffic consumption and handover frequency. We experimented our model on real-world mobile traffic data in Turkey, and our model outperformed other benchmark models in terms of the accuracy, and it especially excels in long-term prediction.

This chapter not only provided a novel way of modelling the cellular network through non-Euclidean representations, but also focused on long-term prediction, while the previous chapter focused on the prediction of the uncollected data at the current timestamp. We used different datasets in these two chapters which encompass different connections between base stations (distance vs. handover).

\chapter{Conclusion and Future Work }

\section{Conclusion}
This thesis addresses several challenges in mobile traffic measurement collection and analysis. We first introduced widely-used Machine Learning algorithms, and surveyed research in different domains, such as mobile traffic sampling and reconstruction, sparse mobile crowdsensing, and forecasting in spatiotemporal tasks. We subsequently presented Spider, a mobile traffic measurement collection and reconstruction framework which significantly reduces the cost of measurement collection and infers traffic consumption with high fidelity from sparse information. Secondly, we design SDGNet, a handover-aware graph  neural  network  model  for  mobile  traffic  forecasting.   We  model  the  cellular network  as  a  graph,  and  leverage  handover  frequency  to  capture  the  dependencies  between base stations across time.  Overall, our work harnesses the powerful hierarchical feature extraction abilities of deep learning, both in the spatial and temporal domains, and puts forward new solutions for precise city-scale mobile traffic analysis and forecasting, that can support greener and more intelligent cellular networks.

\section{Future Work}
Several potential future work directions follow from our work, as we discuss next.

\subsection{End-to-end traffic measurement collection \& reconstruction} 

Potentially, our proposed framework can be further improved from two perspectives.

The first aspect to consider is making the training process more efficient. One drawback of Spider is that the training of this framework is not end-to-end. We train MTRNet first, and then use the trained MTRNet as the value network, to train a policy network in a RL manner. Next, we train the final RL agent by the dataset generated by the policy network. The training process involves three stages, which is not ideal. However, if we train the MTRNet and policy network simutanously in the RL process, both of these networks face difficulties in converging. This is because the update of the policy network's parameters relies on the output of MTRNet, and vice-versa. This situation makes the training unstable.

Similarly in Q Learning, the update of a Q network's parameters relies on the Q network itself, which leads to instability and slow training. Deep Q-Networks \cite{mnih2013playing} provide a solution to this problem by introducing a duplicated target network whose parameters only get updated every few iterations.
Q network updates based on the target network result in faster convergence. Double Q-Leaning \cite{hasselt2010double} has been proposed to maintain two Q-value functions (can be neural networks). Each one receives updates from the other for the next state. This algorithm improves the poor performance caused by large overestimation of action values. Inspired by these two contributions, the potential of end-to-end mobile traffic measurement collection and reconstruction framework remains to be further explored.

\begin{figure}[]
    \centering \includegraphics[width=0.58\textwidth]{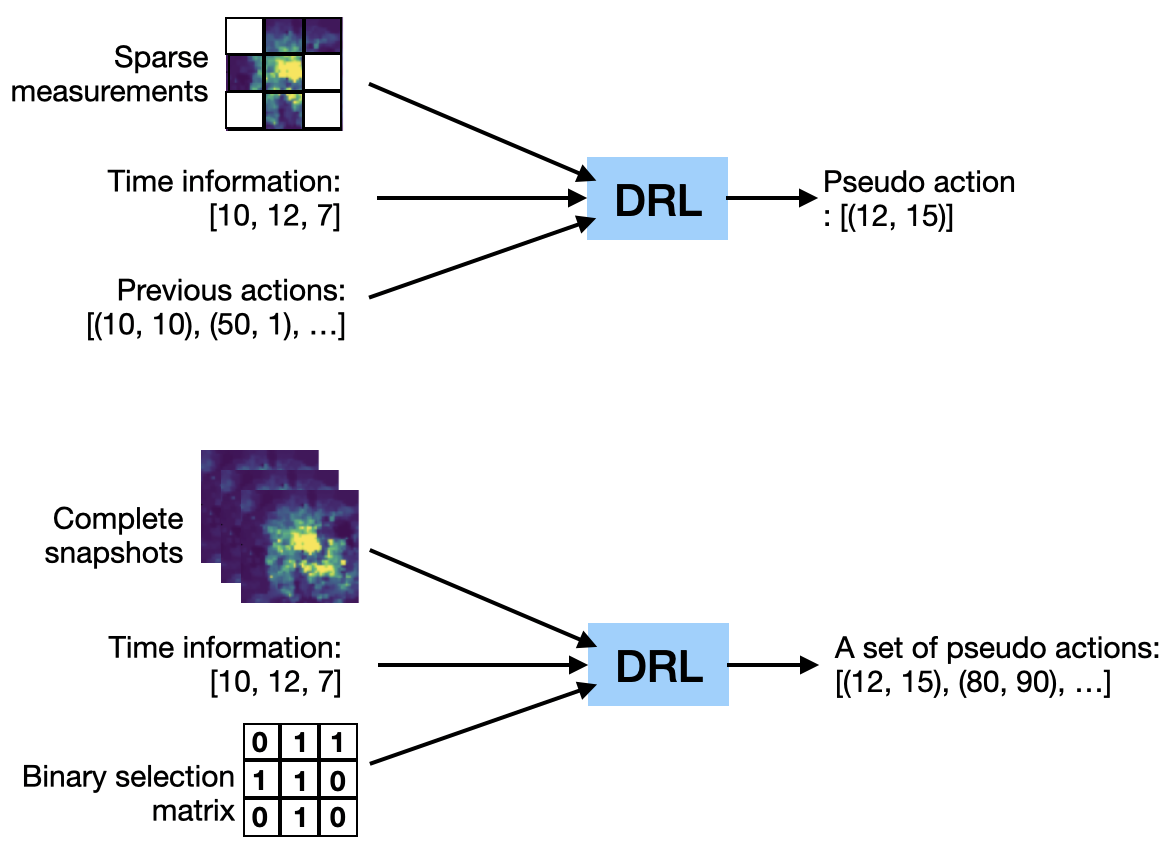}
    \caption{Current inputs (top) and suggested inputs (bottom) to DRL}
    \label{input_drl}
\end{figure}

The second perspective is to improve the accuracy by revising the input to the agent. The input to a model is vital because it contains all the information from which spatiotemporal features are  extracted. Recall that the DRL agent takes the sparse measurements, previous actions, and timestamp information (i.e.,  hour of the day, day of the week and week of the month) as input. When designing the input, we need to make sure that two consecutive inputs in the RL process cannot be similar in order to avoid similar output actions. For example, assume we use a sparse measurement $x_1$ as input and the next action $y_1$ as output of the agent, we apply the ground truth value of $y_1$ to $x_1$ and obtain the new sparse measurement matrix $x_2$ for the next input. Now $x_1$ and $x_2$ are similar, in our case they only have a single different value out of 10,000 values (recall that the target area is divided into 100*100 cells in the dataset that we use). So the output $y_1$ and $y_2$ are also similar, which means the agent chooses from nearby cells. This is an unwanted behavior, because the agent is supposed to explore the entire action space in the deployment area. To address this issue, our input contains the sparse measurement at current timestamp and a vector of previous actions. The drawbacks of our current input are: 1) the vector has order, and different orders result in different output actions, but the output action should not depend on its previous selection order; 2) as we use sparse measurement at current timestamp, there is an information loss since the environment has complete traffic consumption information. The agent should have better knowledge of the environment and can implicitly learn sparse measurement through complete snapshots and selection matrices. Therefore the agent might be improved if the inputs are complete snapshots, selection matrix, and timestamp information. Facing the same problem, in this way only one value in the binary selection matrix changes for consecutive inputs. To solve it, we can output a set of actions each time, so that there are a number of values changed and it also speeds up the training. After we revise on the input, we should also revise on the structure. With better structure and input with more information, the agent may find a more optimal policy. Fig. \ref{input_drl} shows our current inputs (top) and suggested inputs (bottom).
 
\subsection{Hybrid dynamic graphs for mobile traffic forecasting} 
\begin{figure}[]
    \centering \includegraphics[width=0.56\textwidth]{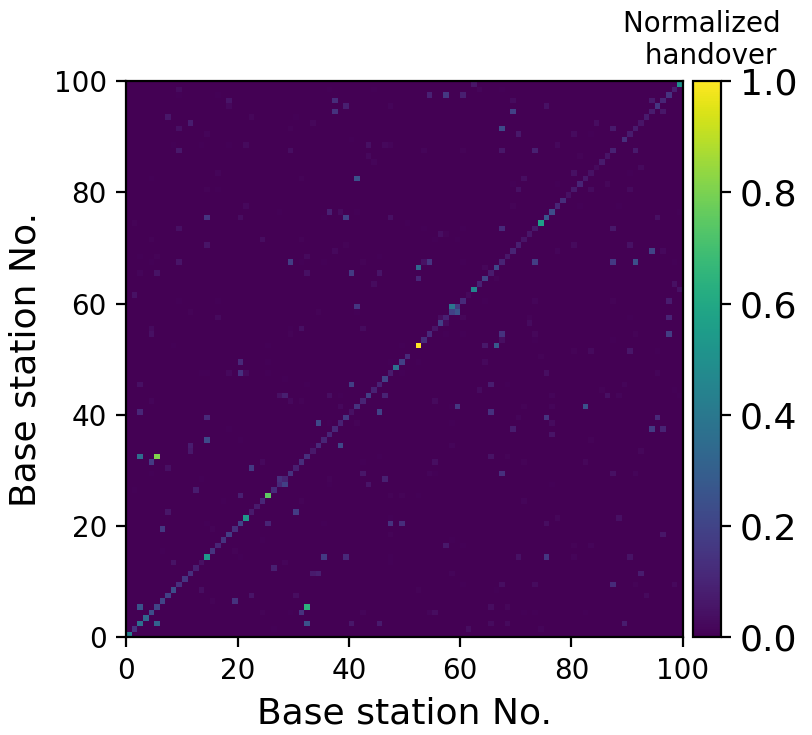}
    \caption{Handover visualization. We normalize the handover frequency by dividing by its maximum value. The y axis represents the outcoming handover, and the x axis represents the incoming handover. For example, the point (20, 40) represents the handover handed from No.40 to No.20 base station. 
    }
    \label{handover}
\end{figure}

Our work showed that modeling cellular network as a graph with handover representing the dependency between base stations can result in precise forecasting, especially long-term. One potential drawback of handover-aware graphs is that, if handovers only happen among a few base stations, the dynamic adjacency matrices do not bring much information about the graph topology. We visualize the normalized adjacency matrix in Fig. \ref{handover}. We observe that the handovers are sparsely scattered, and for some base stations, self-handovers (e.g., 3G/4G technology switch) account for the largest portion. Therefore, we should also consider other factors that contribute to cellular network modeling. Establishing a hybrid graph with a set of weights representing different factors can model the cellular network more accurately, leading to better forecasting results.

As discussed in the related work, Kalander et al. \cite{kalander2020spatio}
 use hybrid GCNs of three types:  spatial proximity, functional similarity and recent trend similarity. This paper can offer a viable inspiration of establishing the hybrid graph. On the other hand, such hybrid graph construction requires high memory and becomes infeasible if the number of base stations is large. This is another question remaining to address in future research.

Practical problems also arise when deploying the forecasting algorithm in a real system. Recall that this takes previous snapshots as input and predicts the future snapshots based on extracted spatiotemporal correlations. However, if a mobile operator wants to use this in real-time, the actual traffic measurements at every timestamp become (partly) unavailable, which means they have to use predicted snapshots as the input. This may lead to concept drift as we discussed before, and the prediction error accumulates over the time. A solution to eliminating error accumulation is critical. Zhang et al. \cite{zhang2018long} propose an Ouroboros Training Scheme (OTS) that could solve the problem. The word `Ouroboros' means a snake eating its own tail. They fine-tune the deep neural network on actual traffic measurements combined with predictions made over these, in order to minimise the difference between its output and the ground truth. The result shows that continuously training the model with the output results in a smaller error compared to those trained with only ground truth data. This approach effectively reduces the error accumulation in long-term predictions and is suitable for adoption in the real-time system.

\subsection{Incorporating uncertainty in mobile traffic forecasting with Bayesian Graph Convolutıonal Networks}

Incorporating uncertainty is one of the prerequisites of reliable forecasting for applications such as health monitoring \cite{catbas2013predictive} and mechanical systems \cite{mocko2005incorporating}. This problem can be challenging due to high  variability caused by unexpected activities  (e.g., holidays, sporting  events), as  extreme  event  predictions  depend  on  some factors  not limited to weather and  city  population growth, and these factors all  contribute  to  the forecasting uncertainty. Especially under the Covid pandemic, the public health measures implemented impact directly on the mobile traffic consumption, as people start work from home. To the best of our knowledge, there are no studies that looked into this with a view to mobile traffic forecasting. 

Traditionally, a probabilistic formulation is usually served as a indicator of uncertainty estimation. However, such models are hard to tune, scale, and add exogenous variables to. Recently,  Bayesian  neural  networks  (BNNs)  have  gained  increasing  attention  as  a  principled  framework  to provide uncertainty estimation for deep neural networks. There are three types of the  prediction  uncertainty in BNNs, namely: model  uncertainty, inherent  noise,  and model  mis-specification.  Model  uncertainty is used in the situations where model parameters are overlooked and can be avoided as more data being collected. Inherent noise captures the uncertainty in the data creation process and cannot be avoided. Mis-specification is used in the situations where the testing data comes from a different distribution than the training data, which happens frequently in time-series problems. BNN application on incorporating uncertainty in mobile traffic forecasting is a potential topic of further study.

\bibliographystyle{IEEEtran}
\bibliography{citation}

\begin{thebibliography}{10}
\providecommand{\url}[1]{#1}
\csname url@samestyle\endcsname
\providecommand{\newblock}{\relax}
\providecommand{\bibinfo}[2]{#2}
\providecommand{\BIBentrySTDinterwordspacing}{\spaceskip=0pt\relax}
\providecommand{\BIBentryALTinterwordstretchfactor}{4}
\providecommand{\BIBentryALTinterwordspacing}{\spaceskip=\fontdimen2\font plus
\BIBentryALTinterwordstretchfactor\fontdimen3\font minus
  \fontdimen4\font\relax}
\providecommand{\BIBforeignlanguage}[2]{{%
\expandafter\ifx\csname l@#1\endcsname\relax
\typeout{** WARNING: IEEEtran.bst: No hyphenation pattern has been}%
\typeout{** loaded for the language `#1'. Using the pattern for}%
\typeout{** the default language instead.}%
\else
\language=\csname l@#1\endcsname
\fi
#2}}
\providecommand{\BIBdecl}{\relax}
\BIBdecl

\bibitem{ericsson}
``{Ericsson Mobility Report},'' June 2021.

\bibitem{zhang2018long}
C.~Zhang and P.~Patras, ``Long-term mobile traffic forecasting using deep
  spatio-temporal neural networks,'' in \emph{Proceedings of the Eighteenth ACM
  International Symposium on Mobile Ad Hoc Networking and Computing}, 2018, pp.
  231--240.

\bibitem{naboulsi2015large}
D.~Naboulsi \emph{et~al.}, ``Large-scale mobile traffic analysis: a survey,''
  \emph{IEEE Communications Surveys \& Tutorials}, vol.~18, no.~1, pp.
  124--161, 2015.

\bibitem{cheng:2017}
X.~{Cheng} \emph{et~al.}, ``Mobile big data: The fuel for data-driven
  wireless,'' \emph{IEEE Internet of Things Journal}, vol.~4, no.~5, pp.
  1489--1516, 2017.

\bibitem{goodfellow2016deep}
I.~Goodfellow, Y.~Bengio, and A.~Courville, \emph{Deep learning}.\hskip 1em
  plus 0.5em minus 0.4em\relax MIT press, 2016.

\bibitem{zhang2019deep}
C.~Zhang, P.~Patras, and H.~Haddadi, ``Deep learning in mobile and wireless
  networking: A survey,'' \emph{IEEE Communications surveys \& tutorials},
  vol.~21, no.~3, pp. 2224--2287, 2019.

\bibitem{bash2004approximately}
B.~A. Bash, J.~W. Byers, and J.~Considine, ``Approximately uniform random
  sampling in sensor networks,'' in \emph{Proceeedings of the 1st international
  workshop on Data management for sensor networks: in conjunction with VLDB
  2004}, 2004, pp. 32--39.

\bibitem{willett2004backcasting}
R.~Willett, A.~Martin, and R.~Nowak, ``Backcasting: adaptive sampling for
  sensor networks,'' in \emph{Proceedings of the 3rd international symposium on
  Information processing in sensor networks}, 2004, pp. 124--133.

\bibitem{unnikrishnan2013sampling}
J.~Unnikrishnan and M.~Vetterli, ``Sampling and reconstruction of spatial
  fields using mobile sensors,'' \emph{IEEE Transactions on Signal Processing},
  vol.~61, no.~9, pp. 2328--2340, 2013.

\bibitem{salama2017adaptive}
A.~Salama, R.~Saatchi, and D.~Burke, ``Adaptive sampling technique for computer
  network traffic parameters using a combination of fuzzy system and regression
  model,'' in \emph{2017 Fourth International Conference on Mathematics and
  Computers in Sciences and in Industry (MCSI)}.\hskip 1em plus 0.5em minus
  0.4em\relax IEEE, 2017, pp. 206--211.

\bibitem{wu2019dynamic}
S.~Wu, W.~Mao, C.~Liu, and T.~Tang, ``Dynamic traffic prediction with adaptive
  sampling for 5g hetnet iot applications,'' \emph{Wireless Communications and
  Mobile Computing}, vol. 2019, 2019.

\bibitem{castanon1997approximate}
D.~A. Castanon, ``Approximate dynamic programming for sensor management,'' in
  \emph{Proceedings of the 36th IEEE Conference on Decision and Control},
  vol.~2.\hskip 1em plus 0.5em minus 0.4em\relax IEEE, 1997, pp. 1202--1207.

\bibitem{evans2001optimal}
J.~Evans and V.~Krishnamurthy, ``Optimal sensor scheduling for hidden markov
  model state estimation,'' \emph{International Journal of Control}, vol.~74,
  no.~18, pp. 1737--1742, 2001.

\bibitem{krishnamurthy2002algorithms}
V.~Krishnamurthy, ``Algorithms for optimal scheduling and management of hidden
  markov model sensors,'' \emph{IEEE Transactions on Signal Processing},
  vol.~50, no.~6, pp. 1382--1397, 2002.

\bibitem{krishnamurthy2001hidden}
V.~Krishnamurthy and R.~J. Evans, ``Hidden markov model multiarm bandits: a
  methodology for beam scheduling in multitarget tracking,'' \emph{IEEE
  Transactions on Signal Processing}, vol.~49, no.~12, pp. 2893--2908, 2001.

\bibitem{krishnamurthy2003correction}
------, ``Correction to" hidden markov model multiarm bandits: a methodology
  for beam scheduling in multitarget tracking",'' \emph{IEEE Transactions on
  Signal Processing}, vol.~51, no.~6, pp. 1662--1663, 2003.

\bibitem{wang2017space}
L.~Wang, D.~Zhang, D.~Yang, A.~Pathak, C.~Chen, X.~Han, H.~Xiong, and Y.~Wang,
  ``Space-ta: Cost-effective task allocation exploiting intradata and interdata
  correlations in sparse crowdsensing,'' \emph{ACM Transactions on Intelligent
  Systems and Technology (TIST)}, vol.~9, no.~2, pp. 1--28, 2017.

\bibitem{donoho2006compressed}
D.~L. Donoho, ``Compressed sensing,'' \emph{IEEE Transactions on information
  theory}, vol.~52, no.~4, pp. 1289--1306, 2006.

\bibitem{zhang2009spatio}
Y.~Zhang, M.~Roughan, W.~Willinger, and L.~Qiu, ``Spatio-temporal compressive
  sensing and internet traffic matrices,'' in \emph{Proceedings of the ACM
  SIGCOMM 2009 conference on Data communication}, 2009, pp. 267--278.

\bibitem{earphone}
R.~K. Rana \emph{et~al.}, ``Ear-phone: an end-to-end participatory urban noise
  mapping system,'' in \emph{IPSN}, 2010.

\bibitem{zhu2012compressive}
Y.~Zhu, Z.~Li, H.~Zhu, M.~Li, and Q.~Zhang, ``A compressive sensing approach to
  urban traffic estimation with probe vehicles,'' \emph{IEEE Transactions on
  Mobile Computing}, vol.~12, no.~11, pp. 2289--2302, 2012.

\bibitem{he2018}
S.~He and K.~G. Shin, ``Steering crowdsourced signal map construction via
  {B}ayesian compressive sensing,'' in \emph{INFOCOM}, 2018.

\bibitem{tian2017data}
L.~Tian, G.~Li, and C.~Wang, ``A data reconstruction algorithm based on neural
  network for compressed sensing,'' in \emph{2017 Fifth International
  Conference on Advanced Cloud and Big Data (CBD)}.\hskip 1em plus 0.5em minus
  0.4em\relax IEEE, 2017, pp. 291--295.

\bibitem{ma:2019}
S.~{Ma}, S.~{Guo}, K.~{Wang}, and M.~{Guo}, ``Service demand prediction with
  incomplete historical data,'' in \emph{Proc. IEEE ICDCS}, 2019.

\bibitem{zipnet}
C.~Zhang \emph{et~al.}, ``Zip{N}et-{GAN}: Inferring fine-grained mobile traffic
  patterns via a generative adversarial neural network,'' in \emph{CoNEXT},
  2017.

\bibitem{chai2020deep}
X.~Chai, G.~Tang, S.~Wang, K.~Lin, and R.~Peng, ``Deep learning for irregularly
  and regularly missing 3-d data reconstruction,'' \emph{IEEE Transactions on
  Geoscience and Remote Sensing}, 2020.

\bibitem{tiittanen2019estimating}
H.~Tiittanen, E.~Oikarinen, A.~Henelius, and K.~Puolam{\"a}ki, ``Estimating
  regression errors without ground truth values,'' \emph{arXiv preprint
  arXiv:1910.04069}, 2019.

\bibitem{mdp}
\BIBentryALTinterwordspacing
R.~BELLMAN, ``A markovian decision process,'' \emph{Journal of Mathematics and
  Mechanics}, vol.~6, no.~5, pp. 679--684, 1957. [Online]. Available:
  \url{http://www.jstor.org/stable/24900506}
\BIBentrySTDinterwordspacing

\bibitem{watkins1992q}
C.~J. Watkins and P.~Dayan, ``Q-learning,'' \emph{Machine learning}, vol.~8,
  no. 3-4, pp. 279--292, 1992.

\bibitem{mnih2013playing}
V.~Mnih, K.~Kavukcuoglu, D.~Silver, A.~Graves, I.~Antonoglou, D.~Wierstra, and
  M.~Riedmiller, ``Playing atari with deep reinforcement learning,''
  \emph{arXiv preprint arXiv:1312.5602}, 2013.

\bibitem{silver2017mastering}
D.~Silver, J.~Schrittwieser, K.~Simonyan, I.~Antonoglou, A.~Huang, A.~Guez,
  T.~Hubert, L.~Baker, M.~Lai, A.~Bolton \emph{et~al.}, ``Mastering the game of
  go without human knowledge,'' \emph{nature}, vol. 550, no. 7676, pp.
  354--359, 2017.

\bibitem{magnuson2015monte}
M.~Magnuson, ``Monte carlo tree search and its applications,'' \emph{Scholarly
  Horizons: University of Minnesota, Morris Undergraduate Journal}, vol.~2,
  no.~2, p.~4, 2015.

\bibitem{zhang20144w1h}
D.~Zhang, L.~Wang, H.~Xiong, and B.~Guo, ``4w1h in mobile crowd sensing,''
  \emph{IEEE Communications Magazine}, vol.~52, no.~8, pp. 42--48, 2014.

\bibitem{guo2015mobile}
B.~Guo, Z.~Wang, Z.~Yu, Y.~Wang, N.~Y. Yen, R.~Huang, and X.~Zhou, ``Mobile
  crowd sensing and computing: The review of an emerging human-powered sensing
  paradigm,'' \emph{ACM computing surveys (CSUR)}, vol.~48, no.~1, pp. 1--31,
  2015.

\bibitem{sparsemcs}
L.~Wang \emph{et~al.}, ``Sparse mobile crowdsensing: challenges and
  opportunities,'' \emph{IEEE Comm. Mag.}, vol.~54, no.~7, pp. 161--167, 2016.

\bibitem{rl_mcs2}
W.~Liu \emph{et~al.}, ``Multi-dimensional urban sensing in sparse mobile
  crowdsensing,'' \emph{IEEE Access}, vol.~7, pp. 82\,066--82\,079, 2019.

\bibitem{rl_mcs}
L.~Wang \emph{et~al.}, ``Cell selection with deep reinforcement learning in
  sparse mobile crowdsensing,'' in \emph{ICDCS}, 2018.

\bibitem{seung1992query}
H.~S. Seung, M.~Opper, and H.~Sompolinsky, ``Query by committee,'' in
  \emph{Proceedings of the fifth annual workshop on Computational learning
  theory}, 1992, pp. 287--294.

\bibitem{dulac2015deep}
G.~Dulac-Arnold, R.~Evans, H.~van Hasselt, P.~Sunehag, T.~Lillicrap, J.~Hunt,
  T.~Mann, T.~Weber, T.~Degris, and B.~Coppin, ``Deep reinforcement learning in
  large discrete action spaces,'' \emph{arXiv preprint arXiv:1512.07679}, 2015.

\bibitem{graph_survey}
Z.~Wu, S.~Pan, F.~Chen, G.~Long, C.~Zhang, and P.~S. Yu, ``A comprehensive
  survey on graph neural networks,'' \emph{arXiv preprint arXiv:1901.00596},
  2019.

\bibitem{998081}
A.~Coates, A.~Hero~III, R.~Nowak, and B.~Yu, ``Internet tomography,''
  \emph{IEEE Signal Processing Magazine}, vol.~19, no.~3, pp. 47--65, 2002.

\bibitem{gilmer2017neural}
J.~Gilmer, S.~S. Schoenholz, P.~F. Riley, O.~Vinyals, and G.~E. Dahl, ``Neural
  message passing for quantum chemistry,'' in \emph{International conference on
  machine learning}.\hskip 1em plus 0.5em minus 0.4em\relax PMLR, 2017, pp.
  1263--1272.

\bibitem{spatial}
M.~Niepert, M.~Ahmed, and K.~Kutzkov, ``Learning convolutional neural networks
  for graphs,'' in \emph{International conference on machine learning}, 2016,
  pp. 2014--2023.

\bibitem{spectral}
J.~Bruna, W.~Zaremba, A.~Szlam, and Y.~LeCun, ``Spectral networks and locally
  connected networks on graphs,'' \emph{arXiv preprint arXiv:1312.6203}, 2013.

\bibitem{7kipf2016semi}
T.~N. Kipf and M.~Welling, ``Semi-supervised classification with graph
  convolutional networks,'' \emph{arXiv preprint arXiv:1609.02907}, 2016.

\bibitem{defferrard2016convolutional}
M.~Defferrard, X.~Bresson, and P.~Vandergheynst, ``Convolutional neural
  networks on graphs with fast localized spectral filtering,'' \emph{Advances
  in neural information processing systems}, vol.~29, pp. 3844--3852, 2016.

\bibitem{li}
Y.~Li, R.~Yu, C.~Shahabi, and Y.~Liu, ``Diffusion convolutional recurrent
  neural network: Data-driven traffic forecasting,'' \emph{arXiv preprint
  arXiv:1707.01926}, 2017.

\bibitem{6velivckovic2017graph}
P.~Veli{\v{c}}kovi{\'c}, G.~Cucurull, A.~Casanova, A.~Romero, P.~Lio, and
  Y.~Bengio, ``Graph attention networks,'' \emph{arXiv preprint
  arXiv:1710.10903}, 2017.

\bibitem{hammond2011wavelets}
D.~K. Hammond, P.~Vandergheynst, and R.~Gribonval, ``Wavelets on graphs via
  spectral graph theory,'' \emph{Applied and Computational Harmonic Analysis},
  vol.~30, no.~2, pp. 129--150, 2011.

\bibitem{stgc}
B.~Yu, H.~Yin, and Z.~Zhu, ``Spatio-temporal graph convolutional networks: A
  deep learning framework for traffic forecasting,'' \emph{arXiv preprint
  arXiv:1709.04875}, 2017.

\bibitem{seo}
Y.~Seo, M.~Defferrard, P.~Vandergheynst, and X.~Bresson, ``Structured sequence
  modeling with graph convolutional recurrent networks,'' in
  \emph{International Conference on Neural Information Processing}.\hskip 1em
  plus 0.5em minus 0.4em\relax Springer, 2018, pp. 362--373.

\bibitem{kipf2016semi}
T.~N. Kipf and M.~Welling, ``Semi-supervised classification with graph
  convolutional networks,'' \emph{arXiv preprint arXiv:1609.02907}, 2016.

\bibitem{diao2019dynamic}
Z.~Diao, X.~Wang, D.~Zhang, Y.~Liu, K.~Xie, and S.~He, ``Dynamic
  spatial-temporal graph convolutional neural networks for traffic
  forecasting,'' in \emph{Proceedings of the AAAI conference on artificial
  intelligence}, vol.~33, no.~01, 2019, pp. 890--897.

\bibitem{malik2021dynamic}
O.~A. Malik, S.~Ubaru, L.~Horesh, M.~E. Kilmer, and H.~Avron, ``Dynamic graph
  convolutional networks using the tensor m-product,'' in \emph{Proceedings of
  the 2021 SIAM International Conference on Data Mining (SDM)}.\hskip 1em plus
  0.5em minus 0.4em\relax SIAM, 2021, pp. 729--737.

\bibitem{pareja2020evolvegcn}
A.~Pareja, G.~Domeniconi, J.~Chen, T.~Ma, T.~Suzumura, H.~Kanezashi, T.~Kaler,
  T.~Schardl, and C.~Leiserson, ``Evolvegcn: Evolving graph convolutional
  networks for dynamic graphs,'' in \emph{Proceedings of the AAAI Conference on
  Artificial Intelligence}, vol.~34, no.~04, 2020, pp. 5363--5370.

\bibitem{1zhang2019spatial}
C.~Zhang, J.~James, and Y.~Liu, ``Spatial-temporal graph attention networks: A
  deep learning approach for traffic forecasting,'' \emph{IEEE Access}, vol.~7,
  pp. 166\,246--166\,256, 2019.

\bibitem{2wu2018graph}
T.~Wu, F.~Chen, and Y.~Wan, ``Graph attention lstm network: A new model for
  traffic flow forecasting,'' in \emph{2018 5th International Conference on
  Information Science and Control Engineering (ICISCE)}.\hskip 1em plus 0.5em
  minus 0.4em\relax IEEE, 2018, pp. 241--245.

\bibitem{3zheng2020gman}
C.~Zheng, X.~Fan, C.~Wang, and J.~Qi, ``Gman: A graph multi-attention network
  for traffic prediction,'' in \emph{Proceedings of the AAAI Conference on
  Artificial Intelligence}, vol.~34, no.~01, 2020, pp. 1234--1241.

\bibitem{4park2019stgrat}
C.~Park, C.~Lee, H.~Bahng, K.~Kim, S.~Jin, S.~Ko, J.~Choo \emph{et~al.},
  ``Stgrat: A spatio-temporal graph attention network for traffic
  forecasting,'' \emph{arXiv preprint arXiv:1911.13181}, 2019.

\bibitem{5guo2019attention}
S.~Guo, Y.~Lin, N.~Feng, C.~Song, and H.~Wan, ``Attention based
  spatial-temporal graph convolutional networks for traffic flow forecasting,''
  in \emph{Proceedings of the AAAI Conference on Artificial Intelligence},
  vol.~33, 2019, pp. 922--929.

\bibitem{tian2021spatial}
C.~Tian and W.~K. Chan, ``Spatial-temporal attention wavenet: A deep learning
  framework for traffic prediction considering spatial-temporal dependencies,''
  \emph{IET Intelligent Transport Systems}, vol.~15, no.~4, pp. 549--561, 2021.

\bibitem{fang2018mobile}
L.~Fang, X.~Cheng, H.~Wang, and L.~Yang, ``Mobile demand forecasting via deep
  graph-sequence spatiotemporal modeling in cellular networks,'' \emph{IEEE
  Internet of Things Journal}, vol.~5, no.~4, pp. 3091--3101, 2018.

\bibitem{shafiq2014geospatial}
M.~Z. Shafiq, L.~Ji, A.~X. Liu, J.~Pang, and J.~Wang, ``Geospatial and temporal
  dynamics of application usage in cellular data networks,'' \emph{IEEE
  Transactions on mobile computing}, vol.~14, no.~7, pp. 1369--1381, 2014.

\bibitem{he2019graph}
K.~He, Y.~Huang, X.~Chen, Z.~Zhou, and S.~Yu, ``Graph attention
  spatial-temporal network for deep learning based mobile traffic prediction,''
  in \emph{2019 IEEE Global Communications Conference (GLOBECOM)}.\hskip 1em
  plus 0.5em minus 0.4em\relax IEEE, 2019, pp. 1--6.

\bibitem{zhao2020cellular}
S.~Zhao, X.~Jiang, G.~Jacobson, R.~Jana, W.-L. Hsu, R.~Rustamov, M.~Talasila,
  S.~A. Aftab, Y.~Chen, and C.~Borcea, ``Cellular network traffic prediction
  incorporating handover: A graph convolutional approach,'' in \emph{2020 17th
  Annual IEEE International Conference on Sensing, Communication, and
  Networking (SECON)}.\hskip 1em plus 0.5em minus 0.4em\relax IEEE, 2020, pp.
  1--9.

\bibitem{kalander2020spatio}
M.~Kalander, M.~Zhou, C.~Zhang, H.~Yi, and L.~Pan, ``Spatio-temporal hybrid
  graph convolutional network for traffic forecasting in telecommunication
  networks,'' \emph{arXiv preprint arXiv:2009.09849}, 2020.

\bibitem{milano}
G.~Barlacchi \emph{et~al.}, ``{A multi-source dataset of urban life in the city
  of Milan and the Province of Trentino},'' \emph{Scientific data}, vol.~2,
  no.~1, 2015.

\bibitem{Jiang:2020}
D.~{Jiang} \emph{et~al.}, ``A compressive sensing-based approach to end-to-end
  network traffic reconstruction,'' \emph{IEEE Transactions on Network Science
  and Engineering}, vol.~7, no.~1, pp. 507--519, 2020.

\bibitem{roughan2011spatio}
M.~Roughan \emph{et~al.}, ``Spatio-temporal compressive sensing and internet
  traffic matrices,'' \emph{IEEE/ACM Trans. Netw.}, vol.~20, no.~3, 2011.

\bibitem{zhang2019new}
K.~Zhang, G.~Chuai, W.~Gao, X.~Liu, S.~Maimaiti, and Z.~Si, ``A new method for
  traffic forecasting in urban wireless communication network,'' \emph{EURASIP
  Journal on Wireless Communications and Networking}, vol. 2019, no.~1, pp.
  1--12, 2019.

\bibitem{zhang2018citywide}
C.~Zhang, H.~Zhang, D.~Yuan, and M.~Zhang, ``Citywide cellular traffic
  prediction based on densely connected convolutional neural networks,''
  \emph{IEEE Communications Letters}, vol.~22, no.~8, pp. 1656--1659, 2018.

\bibitem{wu2019graph}
Z.~Wu, S.~Pan, G.~Long, J.~Jiang, and C.~Zhang, ``Graph wavenet for deep
  spatial-temporal graph modeling,'' \emph{arXiv preprint arXiv:1906.00121},
  2019.

\bibitem{dauphin2017language}
Y.~N. Dauphin, A.~Fan, M.~Auli, and D.~Grangier, ``Language modeling with gated
  convolutional networks,'' in \emph{International conference on machine
  learning}.\hskip 1em plus 0.5em minus 0.4em\relax PMLR, 2017, pp. 933--941.

\bibitem{hasselt2010double}
H.~Hasselt, ``Double q-learning,'' \emph{Advances in neural information
  processing systems}, vol.~23, pp. 2613--2621, 2010.

\bibitem{catbas2013predictive}
N.~Catbas, H.~B. Gokce, and D.~M. Frangopol, ``Predictive analysis by
  incorporating uncertainty through a family of models calibrated with
  structural health-monitoring data,'' \emph{Journal of Engineering Mechanics},
  vol. 139, no.~6, pp. 712--723, 2013.

\bibitem{mocko2005incorporating}
G.~M. Mocko and R.~Paasch, ``Incorporating uncertainty in diagnostic analysis
  of mechanical systems,'' \emph{J. Mech. Des.}, vol. 127, no.~2, pp. 315--325,
  2005.

\end{thebibliography}


\end{document}